\documentclass[]{dukedissertation2023}

\usepackage[T1]{fontenc}
\usepackage{titlesec}
\usepackage{amsmath, amssymb, amsfonts, amsthm}
\usepackage{bm}
\usepackage{graphicx}

\usepackage[square,numbers]{natbib}

\usepackage[title,titletoc]{appendix}
\usepackage{apptools}
\usepackage{color}
\usepackage{bm}
\usepackage{mathabx} 
\usepackage{multirow}
\usepackage{setspace}

\usepackage{mathpazo} 
\usepackage{indentfirst}
\usepackage{enumitem}
\usepackage{xfrac}
\usepackage{mathtools}
\usepackage{caption}
\usepackage{subcaption}
\usepackage{tabularx}
\usepackage{booktabs}
\usepackage{siunitx}

\usepackage{algorithm}
\usepackage{algorithmic}

\usepackage{pdflscape}
\usepackage{listings}
\usepackage{xcolor}
\usepackage{pgfplots}
\usepackage{makecell}
\usepackage[super]{nth}
\usepackage{xr}

\setlist[itemize]{noitemsep,nolistsep}
\setlist[enumerate]{noitemsep,nolistsep}
\newtheorem{theorem}{Theorem}

\newtheorem{proposition}[theorem]{Proposition}

\newtheorem{assumption}[theorem]{Assumption}

\graphicspath{{./Pictures/}{./Chapter1}} 

\author{Jianyi Zhang}
\title{Advancing Deep Learning through Probability Engineering: A Pragmatic Paradigm for Modern AI}
\supervisor{Yiran Chen}
\department{Electrical $\&$ Computer Engineering} 

\date{2025} 

\copyrighttext{All rights reserved except the rights granted by the \href{http://creativecommons.org/licenses/by-nc/3.0/us/}{Creative Commons Attribution-Noncommercial Licence}}

\defensedate{March 26, 2025}
\member{Hai "Helen" Li}
\member{Jian Pei}
\member{Rong Ge}
\member{Maria Gorlatova}

\usepackage[pdftex, pdfusetitle, plainpages=false, letterpaper, bookmarks, bookmarksnumbered, colorlinks, linkcolor=black, citecolor=black, filecolor=black, urlcolor=black,linktoc=none]{hyperref}
\usepackage[capitalise]{cleveref}
\creflabelformat{equation}{#2#1#3}
\makeatletter
\newcommand{\RN}[1]{%
	\textup{\lowercase\expandafter{\it \romannumeral#1}}%
}
\newcommand\getfontsizeofheading{The current font size is: \fontname\font \f@size pt}
\newcommand\getfontsizeforfontsizeclass[1]{{\string #1 is printed as #1  \fontname\font \f@size pt\par}}
\newcommand\printfontsizeforfontsizeclass[1]{{#1 \string #1 is printed in  \fontname\font \f@size pt\par}}
\makeatother

\titleformat{\section}[block]{\fontencoding{T1}\fontfamily{phv}\fontseries{b}%
  \fontshape{it}\fontsize{14pt}{11}\selectfont}{\thesection}{.2em}{\normalbaselines}
 \titleformat{\subsection}[block]{\fontencoding{T1}\fontfamily{phv}\fontseries{b}%
  \fontshape{n}\fontsize{13pt}{11}\selectfont}{\thesubsection}{.2em}{\normalbaselines}
 \titleformat{\subsubsection}[block]{\fontencoding{T1}\fontfamily{ppl}\fontseries{b}%
  \fontshape{n}\fontsize{11pt}{11}\selectfont}{\thesubsubsection}{.2em}{\normalbaselines}

\titlespacing\section{0pt}{14pt plus 0pt minus 0pt}{6pt plus 0pt minus 0pt}
\titlespacing\subsection{0pt}{14pt plus 0pt minus 0pt}{6pt plus 0pt minus 0pt}
\titlespacing\subsubsection{0pt}{14pt plus 0pt minus 0pt}{6pt plus 0pt minus 0pt}
\titlespacing\paragraph{0pt}{6pt plus 0pt minus 0pt}{6pt plus 0pt minus 0pt}
\titlespacing\subparagraph{0pt}{6pt plus 0pt minus 0pt}{6pt plus 0pt minus 0pt}

\newcolumntype{L}[1]{>{\raggedright\let\newline\\\arraybackslash\hspace{0pt}}b{#1}}
\newcolumntype{C}[1]{>{\centering\let\newline\\\arraybackslash\hspace{0pt}}b{#1}}
\newcolumntype{R}[1]{>{\raggedleft\let\newline\\\arraybackslash\hspace{0pt}}b{#1}}

\usepackage{subfiles}

\begin{document}

\maketitle

\abstract
Recent years have witnessed the rapid progression of deep learning, pushing us closer to the realization of AGI (Artificial General Intelligence). Probabilistic modeling is critical to many of these advancements, which provides a foundational framework for capturing data distributions. However, as the scale and complexity of AI applications grow, traditional probabilistic modeling faces escalating challenges—high-dimensional parameter spaces, heterogeneous data sources, and evolving real-world requirements often render classical approaches insufficiently flexible.

This paper proposes a novel concept, “Probability Engineering,” which treats the already-learned probability distributions within deep learning as engineering artifacts. Rather than merely fitting or inferring distributions, we actively modify and reinforce them to better address the diverse and evolving demands of modern AI. Specifically, Probability Engineering introduces novel techniques and constraints to refine existing probability distributions, improving their robustness, efficiency, adaptability, or trustworthiness.

We showcase this paradigm through a series of applications spanning Bayesian deep learning, Edge AI (including federated learning and knowledge distillation), and Generative AI (such as text-to-image generation with diffusion models and high-quality text generation with large language models). These case studies demonstrate how probability distributions—once treated as static objects—can be engineered to meet the diverse and evolving requirements of large-scale, data-intensive, and trustworthy AI systems. By systematically expanding and strengthening the role of probabilistic modeling, Probability Engineering paves the way for more robust, adaptive, efficient, and trustworthy deep learning solutions in today’s fast-growing AI era.
\dedication{To my family and all those who love me}

\tableofcontents 

\listoftables	

\listoffigures	


\acknowledgements

Pursuing a PhD has been both a challenging and deeply rewarding journey, and I am profoundly grateful to all those who have supported me throughout this endeavor.

First and foremost, I would like to extend my deepest gratitude to my advisor, Professor Yiran Chen, for his invaluable guidance, encouragement, and unwavering support. His expertise and thoughtful insights have shaped the direction of my research, and his patience has allowed me the freedom to explore and grow. I am truly fortunate to have had such an advisor.

I am also sincerely thankful to my dissertation committee members, Professor Hai Li, Professor Jian Pei, Professor Rong Ge, Professor Maria Gorlatova, for their constructive feedbacks and thought-provoking suggestions. Their guidance has greatly enhanced the quality of this dissertation, and their support has been instrumental in refining my ideas.

My heartfelt thanks go to my colleagues and lab mates, for the countless discussions, collaborations, and shared experiences. Their camaraderie made each challenge more manageable and each milestone more meaningful. I am deeply appreciative to all my friends, who have provided encouragement and a welcome respite from the demands of research. The moments of laughter and understanding we shared have sustained me through the difficult times.

Lastly, but most importantly, I want to express my profound gratitude to my family. Their unwavering love, faith in my abilities, and constant support gave me the strength to persist through every obstacle. I could not have accomplished this without their belief in me.

Thank you all for walking this journey with me. This dissertation would not have been possible without your guidance, generosity, and encouragement.

%
%
%
\chapter{Introduction}
\label{chap:Introduction}

Deep learning \cite{lecun2015deep} has experienced a meteoric rise in recent years, achieving breakthroughs that push us closer to the realm of artificial general intelligence (AGI). A significant driver behind this success is our growing ability to model and generate complex data distributions, enabling tasks like realistic image synthesis and natural language generation \cite{ho2020denoising,rombach2022high,ramesh2022dalle2,saharia2022photorealistic,llama3modelcard,anil2023palm,openai-chatgpt,GPT4report,team2023gemini,touvron2023llama,wang2024large}. In classical machine learning and statistics, such data modeling has traditionally been handled through probabilistic modeling—building theoretically sound models to represent data distributions and then making predictions or decisions based on those distributions \cite{kingma2013auto,James2013}.  However, as AI systems tackle increasingly rich data, traditional probabilistic modeling is reaching its limits. The challenges are multifold: modern data distributions can be highly complex and multimodal (e.g. images with both continuous pixel intensities and discrete textual elements) \cite{zhang2020stochastic,zhang2024artistimprovinggenerationtextrich}, the distributions in AI systems are often dynamic rather than static (e.g. continually evolving user data or non-stationary streaming inputs) \cite{joren2024sufficient,zhang2023reaugkd}, and the assumptions of classical methods often ignore practical constraints such as computational efficiency, trustworthiness of outcomes \cite{zhang2024towards,zhang2024sled,flop_qian2021}. Moreover, many real-world target distributions are fundamentally inaccessible – we may lack sufficient data to fully characterize them \cite{zhang2024sled}, or there may be no explicit rules or first-principles to describe the phenomena of interest \cite{fedcbs,zhang2024sled}. These factors highlight a growing gap between the elegant theory of probabilistic modeling and the messy reality of modern deep learning applications. 

Given the mounting difficulties of applying traditional probabilistic modeling, AI research has increasingly shifted toward engineering-driven approaches \cite{liu2021pretrainpromptpredictsystematic}. In earlier eras, feature engineering – manually crafting input features – was paramount. This gave way to model engineering, where progress came from designing better neural network architectures and training algorithms. Most recently, prompt engineering has emerged as a powerful paradigm, especially with large language models: practitioners focus on finding the right prompts or instructions to guide the model’s behavior, rather than adjusting abstract probability functions. All of these approaches manipulate tangible components (input features, model structures, or prompts) that are more intuitive and hands-on, and typically less mathematically demanding, than directly specifying probability distributions. For example, prompt engineering allows even non-experts to steer AI models through natural language instructions – an intuitive interaction that lowers the technical barrier to controlling model outputs. This trend towards concrete engineering raises a fundamental question: \textit{Does probabilistic modeling lose its relevance in modern deep learning?} In other words, as we rely more on tweaking models and inputs, is there still a place for formally handling probability distributions in cutting-edge deep learning?

In this dissertation, we argue that instead of abandoning probabilistic thinking, we need to reframe it as an engineering discipline. To this end, we introduce \textbf{Probability Engineering} – a new paradigm that treats probability distributions as first-class engineering objects within the deep learning workflow. This approach is inspired by the success of feature, model, and prompt engineering: just as those treat a model’s components as things to adjust and optimize, Probability Engineering views the design and manipulation of probability distributions as a practical skill integrated into AI development. The key difference from classical probabilistic modeling is a shift in emphasis from mathematical purity to pragmatic utility. We modify, refine, and adapt distributions in whatever ways best serve the end goals of deep learning systems. In a narrow sense, Probability Engineering means pragmatically tweaking probability models under real-world constraints – even if these tweaks sacrifice some theoretical elegance for practical payoff. In a broader sense, one can view Probability Engineering as an evolution of probabilistic modeling itself, extending its toolkit to better align with deep learning’s needs (much like how software engineering extends computer science theory to build real applications). This dissertation formally explores the concept of Probability Engineering, defining its scope and methodologies and illustrating its value through multiple case studies. By the end, we aim to establish a systematic understanding of Probability Engineering and outline how it can shape the future of AI development.

Our first case study applies Probability Engineering to Bayesian Deep Learning \cite{ChenDC:NIPS15,wilson2022bayesiandeeplearningprobabilistic}, a field that explicitly embraces probability distributions to quantify uncertainty. Traditional Bayesian sampling methods like stochastic gradient MCMC \cite{WellingT:ICML11,ChenFG:ICML14,DingFBCSN:NIPS14,ChenDC:NIPS15} or variational inference often struggle with these multimodal and high-dimensional posteriors, leading to slow convergence and poor scalability. To address this, we introduce Stochastic Particle-Optimization Sampling (SPOS), a novel approach that bridges optimization and sampling to make Bayesian inference more tractable for deep networks. SPOS builds on particle-based variational inference (exemplified by Stein variational gradient descent \cite{liu2017stein}) but injects stochastic noise into the particle updates, avoiding the particle-collapse pitfalls of purely deterministic updates. In essence, we engineer the sampling dynamics by adding randomness that enhances exploration of the posterior. This engineered sampler achieves improved sampling efficiency and better convergence properties, backed by a non-asymptotic convergence theory. 

Next, we demonstrate Probability Engineering in contexts where the goal is tackling partially unknown or evolving distributions in real-world Edge AI systems \cite{Gill_2024,SINGH202371}. Two prominent examples we explore are Federated Learning and Knowledge Distillation. In Federated Learning, a model is trained across many decentralized clients (e.g. mobile devices or silos of data) with the constraint that raw data never leaves each client. Here, the global data distribution is essentially an amalgamation of many local distributions, which are often heterogeneous and unknown – clients may have wildly different data profiles, and the server has no direct access to this data to model it. Classical federated algorithms (like FedAvg) typically sample clients at random, which can lead to biased or inefficient training if, say, a random sample of clients yields a highly skewed aggregate dataset. We introduce an approach called Federated Class-Balanced Sampling (Fed-CBS) that applies Probability Engineering to the client selection process. Fed-CBS implicitly models the distribution of data across clients – in particular, it utilizes the class label distribution on each client in a privacy-preserving manner (without collecting the data centrally). Using this knowledge, Fed-CBS engineers the selection probability of clients each round to yield an overall training batch that is more representative and balanced. This method effectively mitigates the non-IID data problem and improves both the convergence speed and final accuracy of federated learning, all while respecting privacy constraints. The second scenario, Knowledge Distillation (KD) \cite{hinton2015distilling}, involves transferring knowledge from a large teacher model to a smaller student model. As modern teachers (e.g. large language models or very deep networks) incorporate ever-expanding datasets, their knowledge distribution becomes vast and sometimes evolving (for instance, as new data is added or when dealing with time-changing facts). Conventional KD techniques typically rely on the teacher’s soft predictions on a training set to guide the student, effectively encoding the teacher’s knowledge only in the student model’s parameters. This approach faces limitations: the student has finite capacity and may not capture all nuances, and importantly, once training is over, the rich knowledge of the teacher (such as intermediate representations or rare examples) is not directly accessible to the student. We propose ReAugKD (Retrieval-Augmented Knowledge Distillation) to address this gap. In ReAugKD, we attach an external knowledge memory to the distillation process, where key outputs or embeddings from the teacher are stored. The student model is trained not just to mimic the teacher’s outputs, but also to retrieve relevant information from this memory and align its representations with the teacher’s on-the-fly. This engineered augmentation allows the student to effectively tap into the teacher’s knowledge base during training (and even potentially during inference), thereby handling distributional differences or updates more gracefully. Our experiments show that ReAugKD leads to a student that better preserves the teacher’s knowledge and achieves superior performance.

We further explore Probability Engineering in the realm of Generative AI \cite{ho2020denoising,rombach2022high,ramesh2022dalle2,saharia2022photorealistic,llama3modelcard,anil2023palm,openai-chatgpt,GPT4report,team2023gemini,touvron2023llama,wang2024large}, focusing on two cutting-edge domains: large language models and text-to-image generation. For large language models (LLMs), one critical challenge is designing decoding strategies that produce accurate and truthful outputs. LLMs like GPT have internal representations that implicitly contain a wealth of latent knowledge, but standard generation procedures (e.g. greedy or probabilistic decoding from the language model head) can still lead to errors or hallucinations when the model’s output distribution isn’t properly aligned with factual reality. In a method we term SLED (self logits evolution decoding), we engineer the decoding process by leveraging signals from the model’s own hidden activations to guide generation. By identifying telltale activation patterns or directions associated with truthful knowledge, SLED adjusts the model’s output probabilities during decoding to favor factually correct completions. This form of probability engineering within the decoder results in notably improved factual accuracy of LLM outputs, as the model is steered to be consistent with its latent knowledge base. In the domain of text-to-image generation, we tackle a different kind of multimodal distribution problem. Here, models like diffusion models must handle data that combine continuous visual information with discrete text (for example, generating an image that contains written characters or symbols). The textual content embedded in images constitutes a discrete distribution (over vocabulary of characters or words) superimposed on the continuous image space, and treating these two aspects uniformly can be suboptimal. Our approach, called ARTIST, introduces a disentangled probabilistic modeling for such cases. We explicitly separate the probability distributions for the visual content and the textual content within images. This specialized probability manipulation for each modality ensures that, for instance, the model can generate legible and correct text in images without compromising the visual realism of the image. By handling the heterogeneous distributions in a unified framework, ARTIST achieves more coherent and high-fidelity text-to-image generation, demonstrating the power of Probability Engineering in complex multi-modal generative tasks.

In summary, through these diverse explorations, we develop and formalize the concept of Probability Engineering as a practical complement to both classical probabilistic modeling and modern deep learning engineering. We compare this paradigm to existing approaches – highlighting that, unlike standard probabilistic modeling which centers on an idealized model of data, Probability Engineering is inherently application-driven, constrained and informed by the needs of deep learning systems (speed, scale, accuracy, etc.). We outline core methodologies that emerge from our case studies, such as injecting stochasticity into deterministic inference algorithms, implicitly modeling unknown data distributions via observable proxies, and augmenting learning processes with external probabilistic knowledge sources. Taken together, the dissertation’s contributions serve as a first step toward establishing Probability Engineering as a structured approach in AI, one that can be taught, systematically applied, and further refined. Finally, we chart several future directions for this nascent field. These include potential applications in reinforcement learning, in trustworthy and explainable AI, and in the automation of generative AI. By laying out these possibilities, we hope to inspire subsequent research to build on the foundations of Probability Engineering, integrating the power of probabilistic reasoning with the hands-on ingenuity of engineering to tackle the next generation of AI challenges.

\chapter{Probability Engineering in Bayesian Deep Learning}
\label{chap:bayesian_dl}
In this chapter, we focus on how probability engineering ideas can be leveraged within Bayesian Deep Learning.

\section{Background}
\label{sec:bayes_background}

 In modern deep learning, sampling-based approximate Bayesian inference has become increasingly important due to its strong theoretical grounding and ability to quantify uncertainty. When dealing with big data and large-scale deep learning architectures, classic Markov Chain Monte Carlo (MCMC) techniques struggle to keep up with the sheer volume of data and number of parameters. In response, researchers have proposed \emph{stochastic gradient}-based Bayesian sampling algorithms, such as stochastic gradient MCMC (SG-MCMC)~\cite{WellingT:ICML11} and Stein variational gradient descent (SVGD)~\cite{liu2016stein}, which exploit mini-batch gradients to achieve more efficient sampling and better scalability. By reducing the computational overhead per iteration, these methods have paved the way for real-world deep learning applications.





\paragraph{Stochastic Gradient MCMC}\label{SGMCMC}
In Bayesian sampling, one aims at sampling from a posterior distribution $p(\bm{\theta}|\mathbf{x})\propto p(\mathbf{x}|\bm{\theta})p(\bm{\theta})$, where $\bm{\theta}\in\mathbb{R}^d$ represents the model parameter, and $\mathbf{X}\triangleq \{\mathbf{x}_j\}_{j=1}^N$ is the dataset. Let $p(\bm{\theta}|\mathbf{X})=(1/Z)\exp(-U(\bm{\theta}))$, where $U(\bm{\theta})=-\log p(\mathbf{X}|\bm{\theta})- \log p(\bm{\theta}) \triangleq -\sum_{i=1}^N\log p(\mathbf{x}_i | \bm{\theta}) - \log p(\bm{\theta})$ is referred to as the potential energy function, and $Z$ is the normalizing constant. We further define the full gradient $F$ and individual gradient $F_j$ used in this paper: $F_j(\bm{\theta})\triangleq  -\nabla_{\bm{\theta}}\log p(\mathbf{x}_j | \bm{\theta}) - \frac{1}{N}\nabla_{\bm{\theta}}\log p(\bm{\theta})$ and $F(\bm{\theta})\triangleq    \nabla_{\bm{\theta}}U(\bm{\theta})=\sum_{j=1}^{N}F_j(\bm{\theta})
$. Now, we can define a stochastic differential equation, an instance of It\'{o} diffusion, whose stationary distribution equals the target posterior distribution $p(\bm{\theta}|\mathbf{X})$. For example, consider the following 1st-order Langevin dynamic:

\begin{align}\label{eq:itodif}
	\mathrm{d}\bm{\theta}_t = -\beta^{-1}F(\bm{\theta}_t)\mathrm{d}t + \sqrt{2\beta^{-1}}\mathrm{d}\mathcal{W}_t~,
\end{align}
where $t$ is the time index, $\mathcal{W}_t \in \mathbb{R}^{d}$ is $d$-dimensional Brownian motion, and $\beta$ a scaling factor. By the Fokker-Planck equation \cite{Kolmogoroff:MA31,Risken:FPE89}, the stationary distribution of \eqref{eq:itodif} equals $p(\bm{\theta}|\mathbf{X})$.

SG-MCMC algorithms \cite{WellingT:ICML11,ChenFG:ICML14,DingFBCSN:NIPS14,ChenDC:NIPS15} are discretized numerical approximations of the It\'{o} diffusions \eqref{eq:itodif}. To make algorithms feasible in a big-data setting, the computationally-expensive term $F$ is replaced with its unbiased stochastic approximation via a random subset of the dataset in each iteration, {\it e.g.} $F$ can be approximated by a stochastic gradient: $G_k \triangleq \frac{N}{B} \sum_{j\in I_k}F_j(\theta_k) =-\nabla\log p(\theta_k)- \frac{N}{B}\sum_{j\in I_k} \nabla_{\theta_k}\log p(\mathbf{x}_{j}|\theta_k)$, where $I_k$ is a random subset of $\{1, 2, \cdots, N\}$ with size $B$. As an example, SGLD \cite{WellingT:ICML11} is a numerical solution of \eqref{eq:itodif}, with update equation:
\vspace{-0.4cm}
\begin{align}
    \theta_{k+1} = \theta_{k} - \beta^{-1}G_kh + \sqrt{2\beta^{-1}h}\xi_{k},
\end{align} where $h$ is the step size and $\xi_k \sim\mathcal{N}(\mathbf{0}, \mathbf{I})$.


\paragraph{Stein Variational Gradient Descent}

Different from SG-MCMC, SVGD initializes a set of particles, which are updated iteratively to approximate the posterior distribution. Specifically, we consider a set of particles $\{\bm{\theta}^{(i)}\}_{i=1}^M$ drawn from some distribution $q$. SVGD tries to update these particles by doing gradient descent on the interactive particle system via
{\small
$\bm{\theta}^{(i)} \leftarrow \bm{\theta}^{(i)} + h \phi(\bm{\theta}^{(i)})$, where $\phi = \arg\max_{\phi\in \mathcal{F}} \{\dfrac{\partial}{\partial h} KL(q_{[h\phi]}||p)\}_{\{h=0\}}$} is a function perturbation direction chosen to minimize the KL divergence between the updated density $q_{[h\phi]}$ induced by the particles and the posterior $p(\bm{\theta}|\mathbf{X})$. SVGD considers $\mathcal{F}$ as the unit ball of a vector-valued reproducing kernel Hilbert space (RKHS) $\mathcal{H}$ associated with a kernel $\kappa(\bm{\theta},\bm{\theta}^{\prime})$. In such a setting, \cite{liu2016stein} shows that $\phi(\bm{\theta}) = \mathbb{E}_{\bm{\theta}^\prime\sim q}[\kappa(\bm{\theta}, \bm{\theta}^\prime) F(\bm{\theta}^\prime)
+ \nabla_{\bm{\theta}^\prime} \kappa(\bm{\theta}, \bm{\theta}^\prime)]$.
When approximating the expectation $\mathbb{E}_{\bm{\theta}^\prime\sim q}[\cdot]$ with an empirical distribution formed by a set of particles $\{\bm{\theta}^{(i)}\}_{i=1}^M$ and adopting stochastic gradients $G_k^{(i)}\triangleq \frac{N}{B} \sum_{j\in I_k}F_j(\theta_k^{(i)})$, we arrive at the following update for the particles: 
\vspace{-0.2cm}
{\begin{align}  \label{eq:svgd_update}
	\theta_{k+1}^{(i)} = \theta_{k}^{(i)} +\dfrac{h}{M} \sum_{q=1}^M \left[ \kappa(\theta_{k}^{(q)}, \theta_{k}^{(i)}) G_k^{(i)}+ \nabla_{\theta_{k}^{(q)}} \kappa(\theta_{k}^{(q)}, \theta_{k}^{(i)}) \right]
\end{align}}
SVGD applies \eqref{eq:svgd_update} repeatedly for all the particles. 

Indeed, SG-MCMC and SVGD have been successfully adopted in various tasks, achieving impressive results in areas such as topic modeling~\cite{GanCHCC:icml15,LiuZS:NIPS16}, matrix factorization~\cite{ChenFG:ICML14,DingFBCSN:NIPS14,SBCG:ICML16}, differential privacy~\cite{WangFS:icml15,LiCLC:arxiv17}, Bayesian optimization~\cite{SpringenbergKFH:NIPS16}, reinforcement learning~\cite{haarnoja2017reinforcement,zhang2018policy,zhang2018scalable}, and training deep neural networks~\cite{PSGLD:AAAI16}. These successes highlight the potential of efficient sampling methods to enrich deep learning with robust uncertainty quantification and better exploration of parameter space.

Nonetheless, existing Bayesian sampling approaches still face notable drawbacks in practical scenarios. For instance, methods like SG-MCMC can produce highly correlated samples, reducing effective sample efficiency, while particle-based approaches like SVGD may suffer from \emph{particle collapse}, in which particles fail to maintain sufficient diversity to represent a complex posterior. Both issues lead to a degradation in performance, especially for multimodal or high-dimensional posteriors commonly encountered in deep learning models. Overcoming these limitations thus remains a significant challenge.

\section{Stochastic Particle-Optimization Sampling (SPOS)}

In the following sections, we will discuss how \textbf{probability engineering} can help mitigate these sampling-related shortcomings, offering strategies to ensure more effective exploration of the posterior and better alignment with the demands of modern deep learning pipelines. By treating the sampling process itself as an engineering target—tweaking particle updates, introducing adaptive noise—we can potentially unlock higher scalability and reliability in Bayesian Deep Learning with our proposed SPOS method.

We focus on the RBF kernel $\kappa(\bm{\theta},\bm{\theta}^{\prime})=\exp(-\frac{\| \bm{\theta}-\bm{\theta}^{\prime} \|^2}{2\eta^2})$ in Equation \ref{eq:svgd_update} due to its wide practical applications. Hence, we can rewrite the kernel $\kappa(\bm{\theta},\bm{\theta}^{\prime})$ with a simpler function $K(\bm{\theta})=\exp(-\frac{\| \bm{\theta} \|^2}{2\eta^2})$. According to the work of \cite{ChenZWLC:UAI18}, the stationary distribution of the $\rho_t$ in the following partial differential equation equals to our target distribution $p(\bm{\theta}|\mathbf{X})$ in Bayesian sampling : 
\begin{align}\label{accept}
\partial_t \rho_t = \nabla_{\bm{\theta}}\cdot (\rho_t\beta^{-1}F(\bm{\theta})+\rho_tE_{Y\sim\rho_t}K(\bm{\theta}-Y)F(Y)-\rho_{t}(\nabla K*\rho_{t})+\beta^{-1}\nabla_{\bm{\theta}}\rho_{t})~.
\end{align}
When approximating the $\rho_t$ in (\ref{accept}) with an empirical distribution formed by a set of particles $\{\bm{\theta}^{(i)}\}_{i=1}^M$, we derive the following diffusion process characterizing the SPOS algorithm 
{\begin{align*}
	&\mathrm{d}\bm{\theta}_{t}^{(i)} =-\frac{F(\bm{\theta}_t^{(i)})}{\beta}\mathrm{d}t - \frac{1}{M}\sum_{q=1}^{M}K(\bm{\theta}_{t}^{(i)} - \bm{\theta}_{t}^{(q)})F(\bm{\theta}_{t}^{(q)})\mathrm{d}t +\frac{1}{M}\sum_{q=1}^{M}\nabla K(\bm{\theta}_{t}^{(i)} - \bm{\theta}_{t}^{(q)})\mathrm{d}t+ \sqrt{\frac{2}{\beta}}\mathrm{d}\mathcal{W}_t^{(i)}.
\end{align*}} Note that if we set the initial distribution of all the particles $\bm{\theta}_0^{(i)}$ to be identical, the system of these \textit{M} particles is exchangeable. Hence, the distributions of all the $\bm{\theta}_ t^{(i)}$ are identical and can be denoted as $\rho_t$. When solving the above diffusion process with a numerical method and adopting stochastic gradients $G_k^{(i)}$, we arrive at the following update equation for SPOS:
\begin{align} \label{particle_num}
{\theta}_{k+1}^{(i)} = {\theta}_{k}^{(i)} -\frac{hG_k^{(i)}}{\beta}  - \frac{h}{M}\sum_{j=1}^{M}K(\theta_{k}^{(i)} - \theta_{k}^{(j)})G_k^{(j)}
+\frac{h}{M}\sum_{j=1}^{M}\nabla K({\theta}_{k }^{(i)} - {\theta}_{k }^{(j)}) + \sqrt{\frac{\xi_{k}^{(i)}}{\beta}} 
\end{align}
where $\xi_{k}^{(i)}\sim\mathcal{N}(\mathbf{0}, \mathbf{I})$. SPOS applies the update of \ref{particle_num} repeatedly for all the particles ${\theta}_{k}^{(i)}$, which has been summarized in Algorithm \ref{alg1}.

\begin{algorithm}[h]
	\caption{Stochastic Particle-Optimization Sampling (SPOS)}
	{\bf Input:} Initial particles $\{\theta_0^{(i)}\}_{i=1}^{M}$, step size $h_k$, batch size $B_k$ 
	\begin{algorithmic}[1]\label{alg1}
		\FOR{iteration $k$= 0,1,...,T}
		\STATE Update ${\theta}_{k+1}^{(i)}$ with \ref{particle_num} for $\forall i$.
		\ENDFOR
	\end{algorithmic}
	{\bf Output:}{$\{\theta_T^{(i)}\}_{i=1}^M$}
\end{algorithm}
\begin{figure}[t!]
	\centering
	\includegraphics[width=0.49\columnwidth]{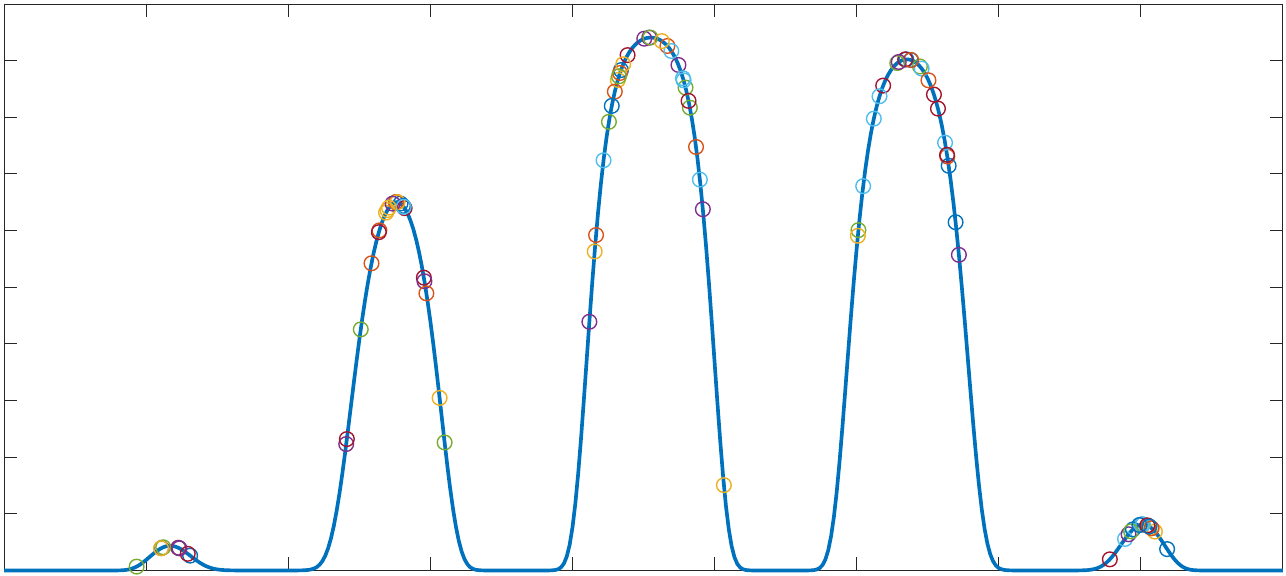}
	\includegraphics[width=0.49\columnwidth]{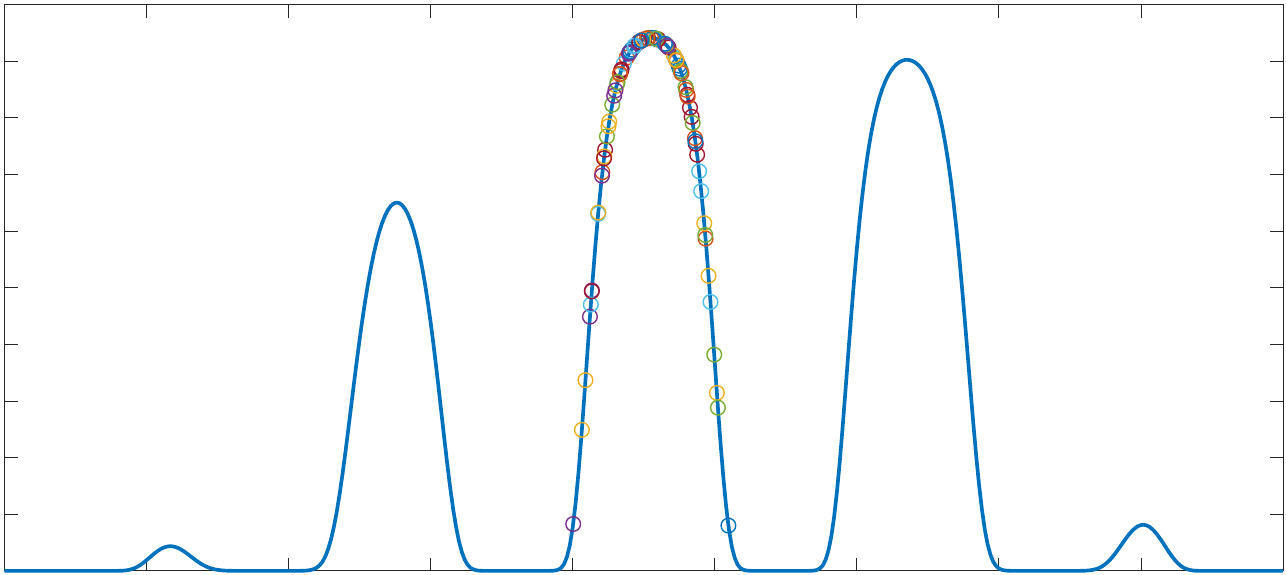}
    \vspace{3mm}
	\caption{Comparison of SPOS (left) and SVGD (right) on a multi-mode distribution.}
	\label{fig:multimode}
\end{figure}
Intuitively, the added Gaussian noise $\sqrt{\beta^{-1}\xi_{k}^{(i)}}$ enhances the ability of the algorithm to jump out of local modes, leading to better ergodic properties compared to standard SVGD. 
To illustrate the advantage of our proposed SPOS, we compare SPOS and SVGD on sampling a complex multi-mode distribution, with the density function given in Section~\ref{app:multimode} of the SM. 
Figure~\ref{fig:multimode} plots the final locations of the particles along with the true density, showing that particles in SPOS can reach different modes, while they are all trapped at one mode in SVGD. We also provide a convergence guarantee for SPOS.

We prove non-asymptotic convergence rates for the proposed SPOS algorithm under the 1-Wasserstein metric ${W}_1$ in Theorem \ref{theo:fixed}, a special case of the p-Wasserstein metric defined as ${W}_p(\mu,\nu)=\left(\inf_{\zeta \in \Gamma(\mu,\nu)}\int _{\mathbb{R}^d\times\mathbb{R}^d} \|X_{\mu}-X_{\nu}\|^p d\zeta(X_{\mu},X_{\nu})\right)^{1/p}
	$, where $\Gamma(\mu,\nu)$ is the set of joint distributions on $\mathbb{R}^d\times\mathbb{R}^d$ with marginal distribution $\mu$ and $\nu$. 

\begin{assumption}\label{assnew:ass1}
	Assume $F$, $K$ and $\nu_0$ satisfy the following assumptions:
	\vspace{-0.15cm}
	\begin{itemize}		

		\item $F$ is $L_F$-Lipschitz continuous {\it i.e.}, $\|F(\bm{\theta}) - F(\bm{\theta}^\prime)\| \leq L_F\|\bm{\theta} - \bm{\theta}^\prime\|$.
		\item There exists positive $m_F$ such that $\langle F(\bm{\theta})-F(\bm{\theta}^\prime), \bm{\theta}-\bm{\theta}^\prime \rangle \geq m_F \|\bm{\theta}-\bm{\theta}^\prime\|^2$.
		\item $K$ is $L_K$-Lipschitz continuous; $\nabla K $ is $L_{\nabla K}$-Lipschitz continuous.
		\item $F(\mathbf{0}) = \mathbf{0}$ and $K$ is an even function, {\it i.e.}, $K(-\bm{\theta}) = K(\bm{\theta})$
		\item The initial probability law of each particle has a bounded and strictly positive density $\nu_0$ with respect to the Lebesgue measure on $\mathbb{R}^d$, and $\gamma_0 \triangleq \log \int_{\mathbb{R}^d}e^{\|\bm{\theta}\|^2}\nu_0(\bm{\theta})d\bm{\theta} < \infty$
	\end{itemize}
\end{assumption}

\begin{theorem}\label{theo:fixed}
	Under Assumption~\ref{assnew:ass1} and setting $h_k=h_0$, $B_k=B_0$, if we denote the distribution of ${\theta}_{T}^{(i)}$ as $\mu_T$ and our target distribution as $\nu_{\infty}$,  ${W}_1(\mu_T, \nu_{\infty})$ is bounded as: 
	\begin{align}\label{eq:fixedbound}
	{W}_1(\mu_T, \nu_{\infty})\leq & \frac{c_1}{\sqrt{M}(\beta^{-1} - c_2)} + c_6Md^{\frac{3}{2}}\beta^{-3}(c_4\beta^2B^{-1}+c_5h)^{\frac{1}{2}}T^{\frac{1}{2}}h^{\frac{1}{2}} \nonumber\\
	&+ c_3\exp\left\{-2\left(\beta^{-1}m_F-L_F-2L_K\right)Th\right\},
	\end{align}
	where $(c_1, c_2, c_3, c_4, c_5, c_6, \beta)$ are positive constants such that $\frac{1}{\beta} > c_2$ and $\frac{m_F}{\beta} >L_F+2L_K$.
\end{theorem}

\section{Experimental Results}
To verify the effectiveness of SPOS for deep learning, we next conduct experiments for Bayesian learning of deep neural networks (DNNs) to empirically compare SGLD, SVGD, and SPOS for the posterior sampling of BNN weights with standard Gaussian priors. Following \cite{li2015stochastic}, 9 UCI public datasets are considered. We use the same setting as \cite{ZhangLCC:AISTATS18}. The datasets are randomly split into 90\% training and 10\% testing. We report the root mean squared error (RMSE) in Table~\ref{tab:reg_1}. The proposed SPOS outperforms both SVGD and SGLD.
\begin{table}[htp]
	\centering
	\caption{\small Averaged RMSE with standard deviations.}
	\label{tab:reg_1}
	\vskip 0.02in
	\scalebox{1}{
		\begin{tabular}{c|ccc}
			\hline
			& \multicolumn{3}{c}{Test RMSE} \\
			Dataset &SGLD&SVGD&SPOS \\
			\hline
			Boston & 3.114 {\scriptsize $\pm$ 0.144} & 2.961 {\scriptsize $\pm$ 0.109} & \textbf{ 2.829} {\scriptsize $\pm$ \textbf{0.126}}  \\
			Concrete & 5.508 {\scriptsize $\pm$ 0.275} & 5.157 {\scriptsize $\pm$ 0.082} & \textbf{ 5.071} {\scriptsize $\pm$ \textbf{0.150}} \\
			Energy& 0.842 {\scriptsize $\pm$ 0.060} & 1.291 {\scriptsize $\pm$ 0.029} & \textbf{ 0.752} {\scriptsize $\pm$ \textbf{0.029}} 
			\\
			Kin8nm & 0.080 {\scriptsize $\pm$ 0.001}   & 0.090 {\scriptsize $\pm$ 0.001} & \textbf{ 0.079} {\scriptsize $\pm$ \textbf{0.001}}\\
			Naval& 0.004 {\scriptsize $\pm$ 0.000} & 0.004 {\scriptsize $\pm$ 0.000} & \textbf{ 0.004} {\scriptsize $\pm$ \textbf{0.000}} \\
			CCPP& 4.059 {\scriptsize $\pm$ 0.080} & 4.127 {\scriptsize $\pm$ 0.027}& \textbf{ 3.939} {\scriptsize $\pm$ \textbf{0.049}} \\
			Wine &0.632 {\scriptsize $\pm$ 0.022} & 0.604 {\scriptsize $\pm$ 0.007} & \textbf{0.598} {\scriptsize $\pm$ \textbf{0.014}} \\
			Yacht& 1.183 {\scriptsize $\pm$ 0.263} & 1.597 {\scriptsize $\pm$ 0.099} & \textbf{ 0.840} {\scriptsize $\pm$ \textbf{0.087}}  \\
			Protein& 4.281 {\scriptsize $\pm$ 0.011} & 4.392 {\scriptsize $\pm$ 0.015} & \textbf{ 4.254} {\scriptsize $\pm$ \textbf{0.005}} \\
			\hline
			\end{tabular}}
	\end{table}

\begin{figure}[h]
	\centering
    \includegraphics[width=0.49\linewidth]{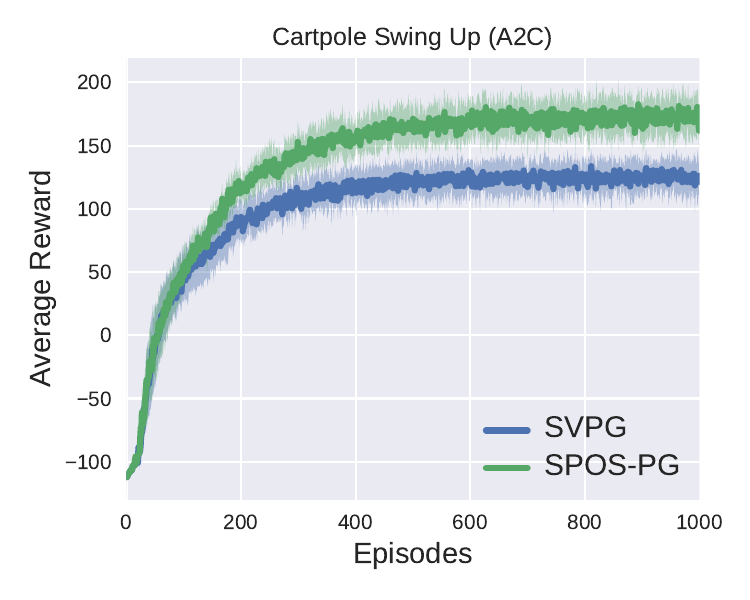}
	\includegraphics[width=0.49\linewidth]{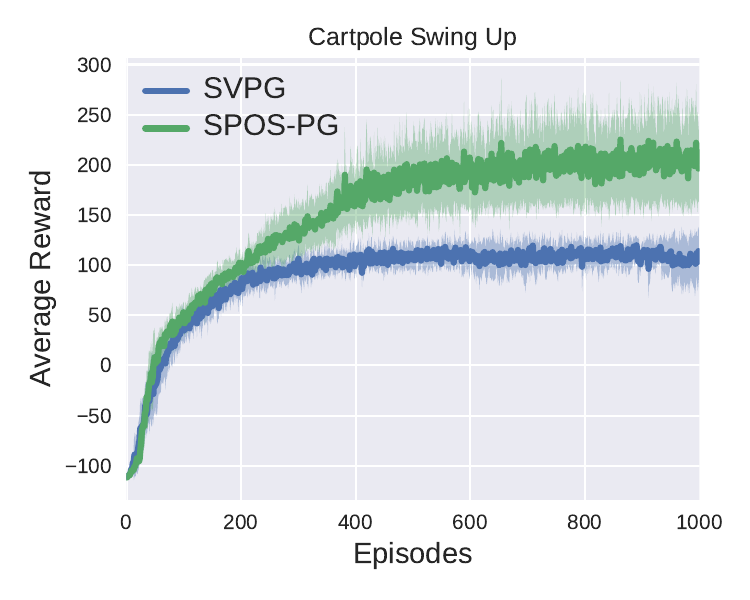}
	\caption{Policy learning with Bayesian exploration in policy-gradient methods with SVPG and SPOS-PG.}\label{fig:bayesian_exp1}
\end{figure}

Besides, we also apply SPOS for reinforcement learning and compare it with SVPG, an SVGD version of the policy gradient method \cite{liu2017stein}. 
We follow the same setting as in \cite{liu2017stein}, except that we use simpler policy-network architectures, as in \cite{houthooft2016vime}. We conduct experiments on the classical continuous control task: Cartpole. Figure~\ref{fig:bayesian_exp1} plots the cumulative rewards overtime on the Cartpole environment, which clearly shows the advantage of our method over SVPG.

\subsection{Summary} 

We applied probability engineering to modify the particle update process in SVGD and SGLD, resulting in a more efficient sampling method called SPOS that better meets practical requirements. The paper on SPOS \cite{zhang2020stochastic} was published in AISTATS 2020. We then carried out a series of follow-up studies to develop more efficient stochastic gradient–based Bayesian sampling methods for deep learning \cite{zhang2020variance,Zhang2020Cyclical,zhao2019self,NEURIPS2022_db174d37}, such as Variance Reduction in SPOS. In standard SPOS, each particle is updated by adopting $G_k^{(i)} \triangleq \frac{N}{B}\sum_{j\in I_k}F_j(\bm{\theta}_k^{(i)})$. Because one can only access $B \ll N$ data points in each step, the increased variance of the ``noisy gradient'' $G_k^{(i)}$ causes a slower convergence rate. A simple way to alleviate this is to increase $B$ by using larger mini-batches. Unfortunately, this brings more computational costs, an undesired side effect.
Thus more effective variance-reduction methods are needed for SPOS. Inspired by recent work on variance reduction in SGLD, {\it e.g.}, \cite{DubeyRPSX:nips16,Nilardri:2018,DIFAN:2018}, we propose three variants of variance-reduced SPOS, called SAGA particle-optimization sampling (SAGA-POS), SVRG particle-optimization sampling (SVRG-POS) and a variant of SVRG-POS which avoids full gradient computations, denoted as SVRG-POS$^+$, based on the SAGA \cite{SAGA:DA2014} and SVRG \cite{RTASGD:AG2013} from stochastic optimization. The details of these methods are summarized in the Algorithm \ref{algo:algo1}, \ref{algo:algo2} and \ref{algo:algo3} in Section \ref{app:vrspos}. Besides, we also proposed cyclical SG-MCMC methods, which enable automatic exploration of complex multimodal distributions. This line of work was presented as an Oral Presentation at ICLR 2020 \cite{Zhang2020Cyclical}. Beyond these, we also published a paper \cite{NEURIPS2022_db174d37} which employed probability engineering principles to refine contrastive learning techniques in a Bayesian manner at Neurips 2021 .
\chapter{Probability Engineering in Edge AI}
\label{chap:edgeAI}

Edge AI systems bring intelligence directly to devices at the network edge, reducing latency, improving privacy, and enabling real-time decision-making in distributed applications. Two prominent areas related to Edge AI are Federated Learning (FL), which trains models across decentralized devices without sharing raw data \cite{pmlr-v54-mcmahan17a}, and Knowledge Distillation (KD), which compresses complex models into lighter versions suitable for edge deployment while maintaining performance~\cite{hinton2015distilling}. In this chapter, we explore how Probability Engineering can be leveraged to address key challenges within these domains, specifically client selection in FL and the adaptation of knowledge distillation techniques, by engineering probability distributions to better align with practical demands.

\section{Federated Learning}\label{Fed-CBS}

With the booming of IoT devices, a considerable amount of data is generated at the network edge, providing valuable resources for learning insightful information and enabling intelligent applications such as self-driving, video analytics, anomaly detection, etc. The traditional wisdom is to train machine learning models by collecting data from devices and performing centralized training. Data migration usually raises serious privacy concerns.  Federated learning (FL) \cite{pmlr-v54-mcmahan17a} is a promising technique to mitigate such privacy concerns, enabling a large number of clients to learn a shared model collaboratively, and the learning process is orchestrated by a central server. In particular, the participating clients first download a global model from the central server and then compute local model updates using their local data. The clients then transmit the local updates to the server, where the local updates are aggregated and then the global model is updated accordingly.  

\subsection{Background}

In practice, due to limited communication and computing capabilities, one usually can not engage all the available clients in FL training to fully utilize all the local data. Therefore, most FL methods only randomly select a subset of the available clients to participate in the training in each communication round. However, in practice, the data held by different clients are often typically non-IID (independent and identically distributed) due to various user preferences and usage patterns. This leads to a serious problem that the random client selection strategy often fails to learn a global model that can generalize well for most of the participating clients under non-IID settings~\cite{goetz2019active,Cho2020ClientSI,Nishio2019ClientSF,yang2020federated}.  

Previous literature has made some efforts to improve client sampling for FL. In the method of ~\cite{goetz2019active}, the clients with more considerable local loss will have a higher probability of being sampled to participate in training. Power-of-Choice~\cite{Cho2020ClientSI} selects several clients with the largest loss from a randomly sampled subset of all the available clients. However, sampling the clients with more significant local loss may not guarantee that the final model can have a more minor global loss. Focusing on the diversity in client selection, the authors \cite{balakrishnan2021} select clients by maximizing a submodular facility location function defined over gradient space. 
\begin{figure*}[!htbp]
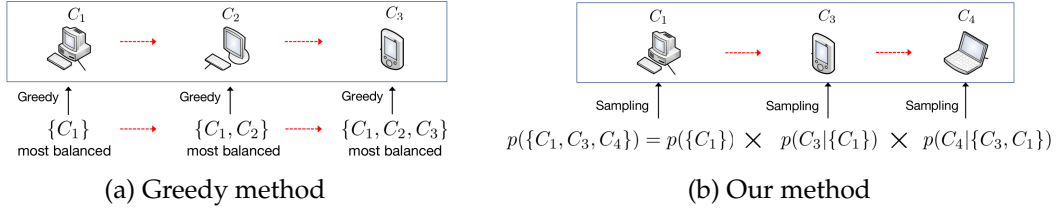

\begin{subfigure}{.49\textwidth}
  \centering
  \includegraphics[width=\linewidth]{Fig/greedy.pdf}
  \caption{Greedy method}
  \label{fig:greedy}
\end{subfigure}
\begin{subfigure}{.49\textwidth}
  \centering
  \includegraphics[width=0.99\linewidth]{Fig/ours.pdf}  
  \caption{Our method}
  \label{fig:our method}
\end{subfigure}
\caption{An example demonstrating the weakness of greedy method to deal with class imbalance.}
\label{fig:greedyandours}
\end{figure*}
The authors of \cite{ribero2020communication} model the progression of model weights by an Ornstein-Uhlenbeck process and design a sampling strategy for selecting clients with significant weight
updates. However, the work only considers identical data distribution. To the best of our knowledge, \cite{Duan2019AstraeaSF} and  \cite{yang2020federated} are the first two attempts to improve client sampling by reducing the class-imbalance. An extra virtual component called mediator, which has access to the local label distributions of the clients, is introduced in \textit{Astraea} of \cite{Duan2019AstraeaSF}, causing severe concerns about privacy leakage. The weakness of \cite{yang2020federated} is still obvious. First, their estimations of the clients' local label distribution are inaccurate.  Theorem 1 in \cite{yang2020federated}, which supports their estimations, can not be generalized to multi-class classification tasks since it has only been proved in the original paper \cite{anand1993improved} for two-class classification problems. Besides, their performance is not guaranteed due to the nature of the greedy algorithm, as shown in Figure \ref{fig:greedyandours}. Supposing we work on a 6-class classification task and aim to select 3 clients from 4 available clients $C_1, C_2, C_3, C_4$. Each of them has 30 images. The compositions of their local datasets are $[5,5,5,5,5,5]$, $[6,6,6,6,6,0]$, $[0,0,0,10,10,10]$ and  $[10,10,10,0,0,0]$ respectively. The greedy method in \cite{yang2020federated} is deterministic. It can only derive one result $\{C_1, C_2, C_3\}$ instead of the optimal solution $\{C_1, C_3, C_4\}$. But in our Fed-CBS, the optimal solution $\{C_1, C_3, C_4\}$ can be sampled with high probability.

Building on these observations, if we view each client's selection probability as a distribution, these baselines' poor training performance makes it apparent that they do not effectively capture this distribution—despite their theoretical or intuitive underpinnings. This limitation stems chiefly from the fact that the ideal distribution we seek is fundamentally inaccessible: there may be no explicit rules describing the underlying phenomena of interest, and we lack clear knowledge of which specific distribution would satisfy our requirements under heterogeneous data conditions. To address this gap, we apply probability engineering on it, wherein we first identify the relevant first principles and then design our solution according to those principles, thereby tackling the core challenges posed by heterogeneous data in federated learning.

\subsection{Federated Class-balanced Sampling (Fed-CBS)}

\begin{figure*}[!hbtp]
\centering
\begin{subfigure}{0.24\textwidth}
  \centering
  \includegraphics[width=1.0\linewidth]{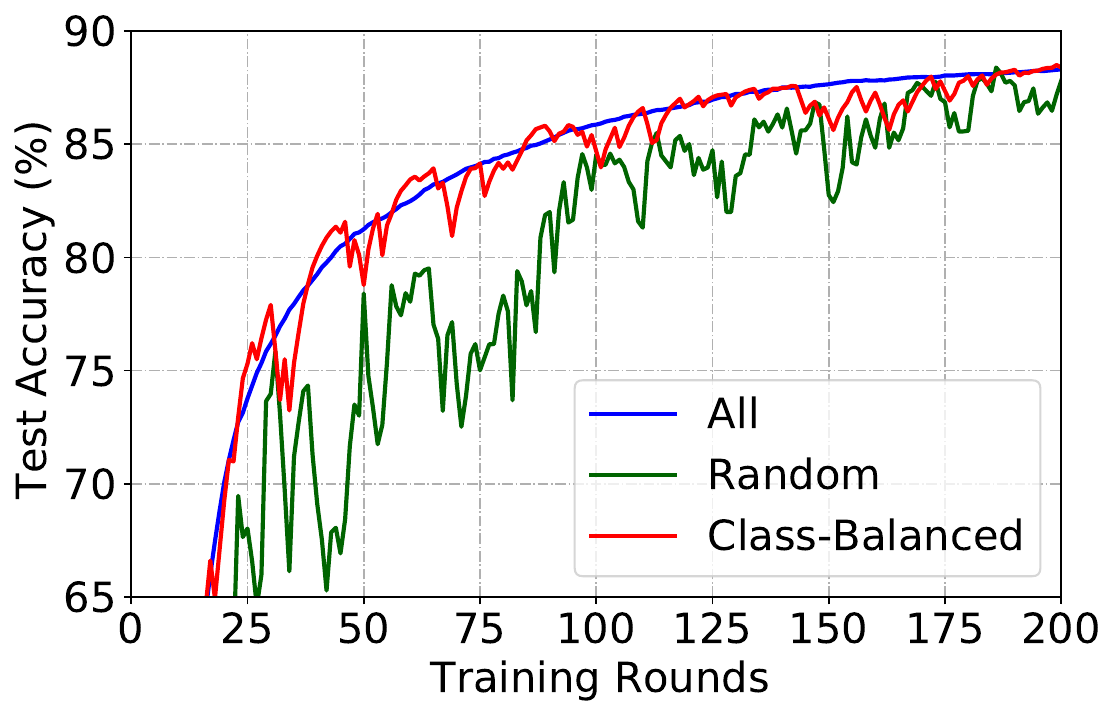}
  \caption{Global Balanced \& One-class}
  \label{fig:fig1-sub-first}
\end{subfigure}
\begin{subfigure}{.24\textwidth}
  \centering
  \includegraphics[width=1.0\linewidth]{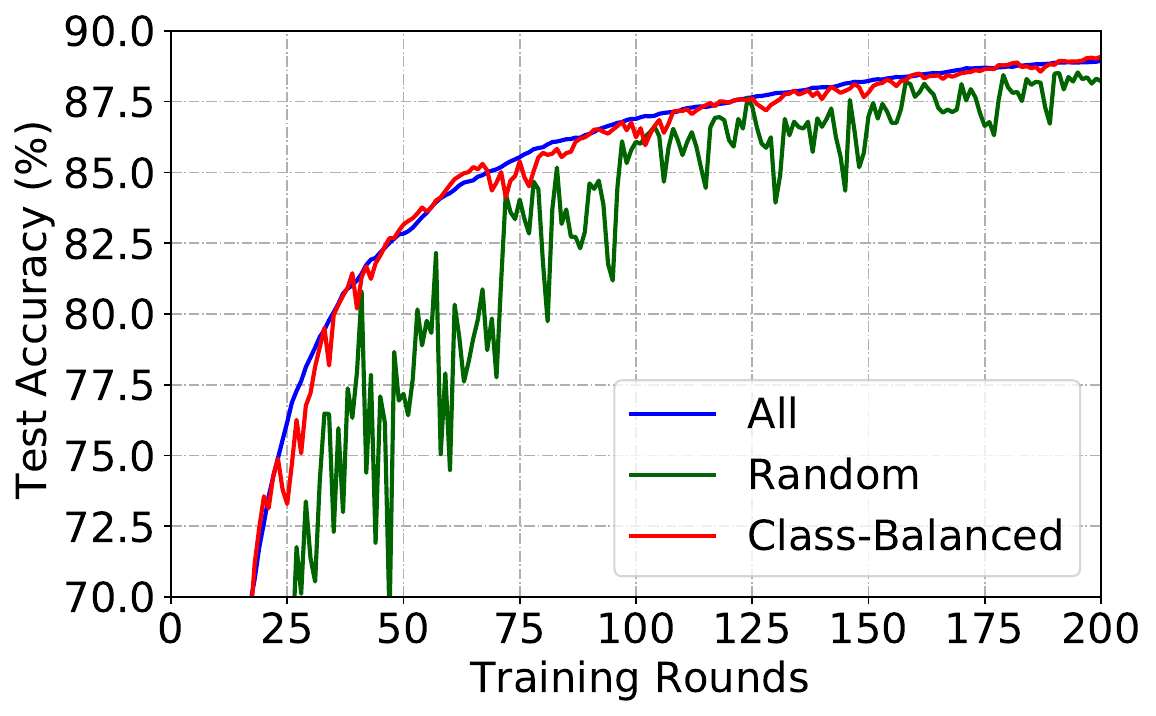}  
  \caption{Global Balanced \& Two-class}
  \label{fig:fig1-sub-second}
\end{subfigure}
\begin{subfigure}{.24\textwidth}
  \centering
  \includegraphics[width=1.0\linewidth]{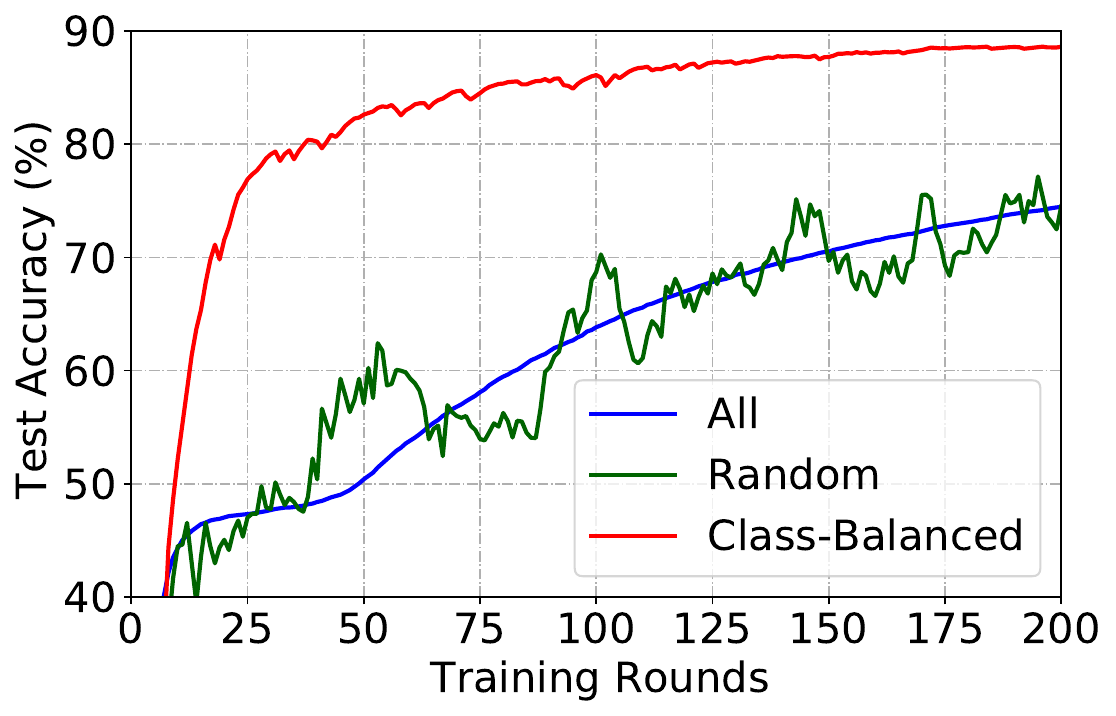}
  \caption{ Global Imbalanced \& One-class}
  \label{fig:fig1-sub-third}
\end{subfigure}
\begin{subfigure}{.24\textwidth}
  \centering
  \includegraphics[width=1.0\linewidth]{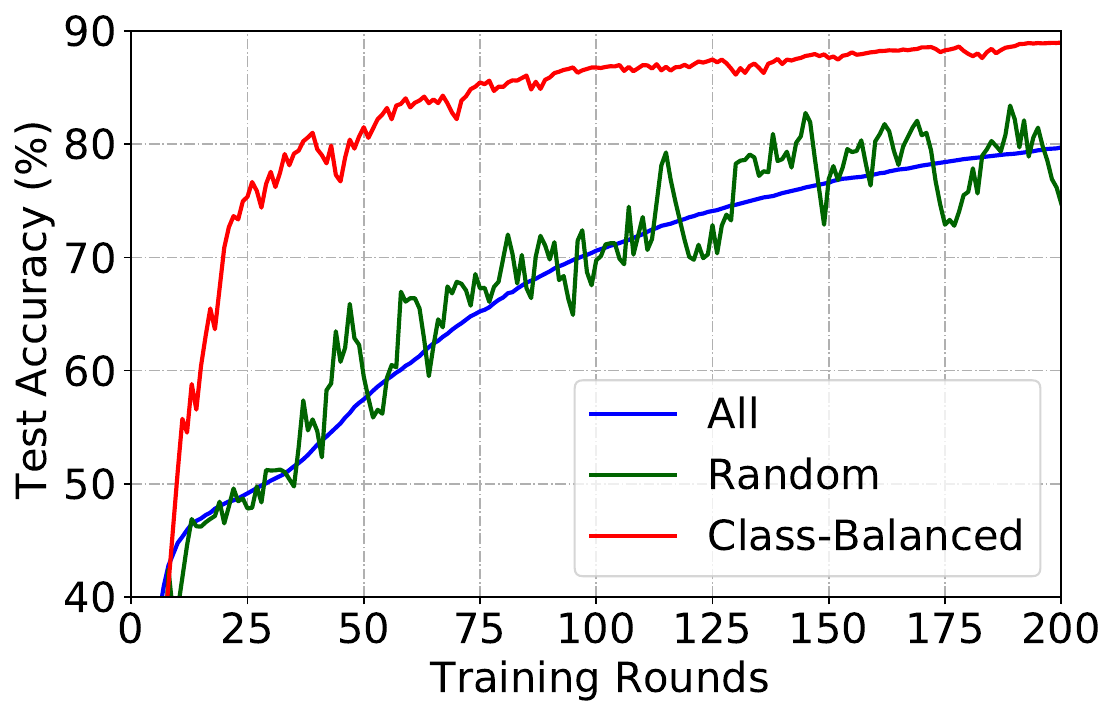}
  \caption{Global Imbalanced \& Two-class}
  \label{fig:fig1-sub-fourth}
\end{subfigure}
\caption{Three different FL client selection strategies on MNIST.
Each client has only one class of data in (a) and (c) and each client has two classes of data in (b) and (d). }
\end{figure*}

First, we unveil the essential reason for the performance degradation on non-IID data with the random client sampling strategy in FL training, \textit{i.e.}, the class-imbalance. We conduct some experiments on MNIST to verify this \footnote{Detailed experiment settings are listed in the Appendix (Section \ref{Sec:settingsec5})}. As shown in Figure \ref{fig:fig1-sub-first} and Figure \ref{fig:fig1-sub-second}, the random sampling mechanism shows the worst performance when the global label distribution is class-balanced. Suppose we keep the grouped dataset class-balanced by manually selecting the clients based on their local label distribution. In that case, we can obtain accuracy comparable to the case of fully engaging all the clients in training. Another natural corollary is that when the global dataset is inherently class-imbalanced, engaging all clients in training may lead to worse performance than manually keeping class balanced in the grouped dataset. The results in Figure \ref{fig:fig1-sub-third} and Figure \ref{fig:fig1-sub-fourth} prove our hypothesis and verify the importance of class-imbalance reduction. 

Motivated by this insight, we first propose a new metric, QCID, to measure class-imbalance. Then we present a method to derive such a measure in a privacy-preserving way. We then design our client sampling mechanism and show its superiority based on this measure.

Assume there are $B$ classes of data in an image classification task, where $B\geq 2$. In the $k$-th communication round, we assume there are $N_k$ available clients and we select $M$ clients from them. To make the presentation concise, we ignore the index ``k'' and assume the set of indices for the available clients is  $\{1,2, 3, ..., N\}$ and the $n$-th available client has its own training dataset $\mathcal{D}_n$. We adopt the following vector $\bm{\alpha}_n$ to represent the local label distribution of $\mathcal{D}_n$, $\bm{\alpha}_n=\left[\alpha_{(n,1)}, \alpha_{(n,2)},...,\alpha_{(n,b)},... ,\alpha_{(n,B)}\right]$, where $\alpha_{(n,b)}\geq 0$ and $\sum_{b=1}^B\alpha_{(n,b)}=1$. We aim to find a subset $\mathcal{M}$ of $\{1,2,3,..,N\}$ of size $M$, such that the following grouped dataset {\small $\mathcal{D}^{g}_{\mathcal{M}}=\bigcup\limits_{n\in \mathcal{M}} \mathcal{D}_n$} is more class-balanced. Assuming the $n$-th client's local dataset has $q_n$ training data in total, one can derive the following vector $\bm{\alpha}^g_{\mathcal{M}}=\frac{\sum_{n\in \mathcal{M} }q_n\bm{\alpha}_n}{\sum_{n\in \mathcal{M} }q_n}= \left[\frac{\sum_{n\in \mathcal{M} }q_n \alpha_{(n,1)}}{\sum_{n\in \mathcal{M} } q_n},...,\frac{\sum_{n\in \mathcal{M} }q_n \alpha_{(n,b)}}{\sum_{n\in \mathcal{M} } q_n},... ,\frac{\sum_{n\in \mathcal{M} }q_n \alpha_{(n,B)}}{\sum_{n\in \mathcal{M} } q_n}\right]$, which represents the label distribution of the grouped dataset $\mathcal{D}^{g}_{\mathcal{M}}$. Now we propose the following function to measure the magnitude of class-imbalance of $\mathcal{M}$, which we call \textit{Quadratic Class-Imbalance Degree (QCID)}:
\begin{align*}
\textit{QCID} (\mathcal{M}) \triangleq \sum_{b=1}^B(\frac{\sum_{n\in \mathcal{M} }q_n \alpha_{(n,b)}}{\sum_{n\in \mathcal{M} } q_n}-\frac{1}{B})^2=\frac{\sum_{n\in \mathcal{M},n^\prime \in \mathcal{M}} q_nq_n^\prime\bm{\alpha}_n\bm{\alpha}_{n^\prime}^T}{(\sum_{n\in \mathcal{M} }q_n)^2} -\frac{1}{B}.
\end{align*}
To derive the \textit{QCID} for any subset $\mathcal{M} \subseteq \{1,2,3,..,N\}$, we only need to know the following $N\times N$ matrix $\bm{S}$ with element $s_{n,n^\prime}$ being $\bm{\alpha}_n\bm{\alpha}_{n^\prime}^T$, which is the inner product between the local label distributions of the available clients $n$ and $n^\prime$. FHE~\cite{bgvfhe,fvfhe,halevi2015bootstrapping} enables to perform computation (addition and multiplication) on encrypted data. We provide a framework in Section \ref{FHEframework} as an example to show it is possible to deriving ${S}$ without knowing the values of local label distributions $\{\bm{\alpha}_i\}$ using FHE. 
\begin{align*}
 \bm{S}=
 \begin{bmatrix}
q_1q_1\bm{\alpha}_1\bm{\alpha}_{1}^T & q_1q_2\bm{\alpha}_1\bm{\alpha}_{2}^T  & \cdots   & q_1q_N\bm{\alpha}_1\bm{\alpha}_{N}^T   \\ 
q_2q_1\bm{\alpha}_2\bm{\alpha}_{1}^T & q_2q_2\bm{\alpha}_2\bm{\alpha}_{2}^T  & \cdots   & q_2q_N\bm{\alpha}_2\bm{\alpha}_{N}^T  \\
\vdots & \vdots  & \ddots   & \vdots  \\
q_Nq_1\bm{\alpha}_{N}\bm{\alpha}_{1}^T & q_Nq_2\bm{\alpha}_{N}\bm{\alpha}_{2}^T  & \cdots\  & q_Nq_N\bm{\alpha}_{N}\bm{\alpha}_{N}^T  \\
\end{bmatrix}
\end{align*}
We consider $\mathcal{M}$ as a sequence of random variables, \textit{i.e.} $\mathcal{M}=\{C_1,C_2,...,C_m,...,C_M\}$ and assign it with some probability. Our expectation is that $\mathcal{M}$ should have higher probability to be sampled if it is more class-balanced. This means $P(C_1=c_1,C_2=c_2,...,C_m=c_m,...,C_M=c_M)$ should be larger if $\mathcal{M}=\{c_1,c_2,...,c_M\}$ has a lower $QCID$ value. Our sampling strategy generates the elements in $\mathcal{M}$ in a sequential manner, {\it i.e.}, we first sample $\mathcal{M}_1=\{c_1\}$ according to the probability of $P(C_1=c_1)$, then sample $c_2$ to form $\mathcal{M}_2=\{c_1,c_2\}$ according to the conditional probability $P(C_2=c_2|C_1=c_1)$. The same procedure applies for the following clients
until we finally obtain $\mathcal{M}=\{c_1,c_2,...,c_M\}$. In the following, we will design proper conditional probabilities such that the joint distribution of client selection satisfies our expectation.

Let $T_{n}$ denote the number of times that client $n$ has been selected. Once client $n$ has been selected in a communication round, $T_{n} \rightarrow T_{n}+$ 1, otherwise, $T_{n} \rightarrow T_{n}$. In the $k$-th communication round, the first element is designed to be sampled with the following probability: $P(C_1=c_1) \propto \frac{1}{[QCID(\mathcal{M}_1)]^{\beta_1}}+\lambda \sqrt{\frac{3 \ln k}{2 T_{c_1}}},$ where $\beta_1>0,$ and $\lambda$ is the exploration factor to balance the trade-off between exploitation and exploration. The second term will add higher probability to the clients that have never been sampled before in the following communication rounds. After sampling $C_1$, the second client is defined to be sampled with probability $P(C_2=c_2|C_1=c_1) \propto \frac{\frac{1}{[QCID(\mathcal{M}_2)]^{\beta_2}}}{\frac{1}{[QCID(\mathcal{M}_1)]^{\beta_1}}+\alpha \sqrt{\frac{3 \ln k}{2 T_{c_1}}}}, \beta_2>0.$ For the m-th client ($2 < m \leq M$), we define $P(C_m=c_m|C_{m-1}=c_{m-1},...,C_2=c_2,C_1=c_1)\propto  \frac{[QCID(\mathcal{M}_{m-1})]^{\beta_{m-1}}}{[QCID(\mathcal{M}_m)]^{\beta_m }},$ where $\beta_{m-1},\beta_{m}>0.$
With the above sampling process, the final probability to sample $\mathcal{M}$ is $P(C_1=c_1,C_2=c_1,...,C_M=c_M)=P(C_1=c_1)\times P(C_2=c_2|C_1=c_1)\cdots\times P(C_M=c_M|C_{M-1}=c_{M-1},...,C_2=c_2,C_1=c_1)\propto {1}/{[QCID(\mathcal{M})]^{\beta_M }}$. Since $\beta_M>0$, this matches our goal that the $\mathcal{M}$ with lower $QCID$ value should have higher probability to be sampled. Fed-CBS is summarized in the following Algorithm.

\begin{figure*}[!hbtp]
  \centering
  \includegraphics[width=1\linewidth]{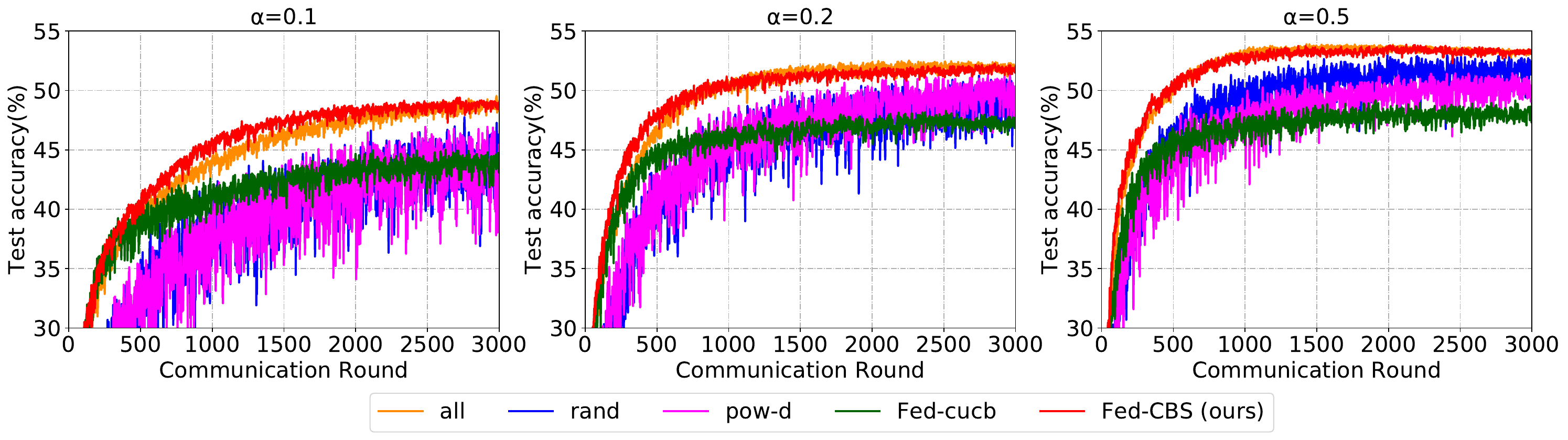}
\vspace{2mm}
\caption{Test accuracy on Cifar-10 under three heterogeneous settings.}
\label{fig:figcifar1}
\end{figure*}

\begin{algorithm}[!htbp]
\begin{algorithmic}
\caption{Fed-CBS}\label{Algorithm123}

 \STATE {\bfseries Initialization:} initial local model $\boldsymbol{w}^{(0)}$, client index subset $\mathcal{M}=\varnothing$, $K$ communication rounds, $k=0$, $T_n=1$\\
\WHILE{$k < K$}

 \STATE{\bfseries Client Selection}:
\FOR{$n$ {\bfseries in} $\{1,2,...,N\}$}
\IF{$n\in\mathcal{M}$}
   \STATE $T_n\rightarrow T_n+1$
\STATE \textbf{else} $T_n\rightarrow T_n$.
   \ENDIF
\ENDFOR
\STATE Update $\mathcal{M}$ using our proposed sampling strategy in Section \ref{Fed-CBS}\\
\STATE {Local Updates:}
\FOR{$n\in \mathcal{M}$}
\STATE $\boldsymbol{w}_n^{(k+1)}\leftarrow \ Update(\boldsymbol{w}^{(k)})$. \\
\ENDFOR
\STATE { Global Aggregation}:
\STATE $\boldsymbol{w}^{(k+1)}\leftarrow Aggregate(\boldsymbol{w}_n^{(k+1)})$ for $n \in\mathcal{M}$

\ENDWHILE
\end{algorithmic}    
\end{algorithm}
We conduct the simulation of the cross-device federated learning (CDFL) on CIFAR-10. 
In our experiment, we set 200 clients in total with a class-balanced global dataset. The non-IID data partition among clients is based on the settings of Dirichlet distribution parameterized by the concentration parameter $\alpha$ in \cite{Hsu2019MeasuringTE}. In each communication round, we uniformly and randomly set 30$\%$ of them (i.e., 60 clients) available and sample 10 clients from those 60 available ones to participate in the training. 
There are four baselines: random selection (rand), Power-of-choice Selection Strategy (pow-d)~\cite{Cho2020ClientSI}, the method in  \cite{yang2020federated} (Fed-cucb), the ideal setting where we select all the available clients (all). As a benefit of successfully reducing the class-imbalance, our method outperforms the other three baseline methods. Furthermore, it achieves comparable performance to the ideal setting where all the available clients are engaged in training, as shown in Figure \ref{fig:figcifar1}. Details of the experimental setup are listed in Section \ref{Sec:settingsec5}.

\begin{table*}[t]
\centering
\renewcommand{\arraystretch}{1}
\setlength\tabcolsep{3pt}
\scalebox{1}{\begin{tabular}{|c|c|c|c|c|c|c|}
\hline
\multicolumn{2}{|l|}{}                                                                                  & all            & rand        & pow-d            & Fed-cucb      & Fed-CBS          \\ \hline
\multirow{3}{*}{Communication Rounds}         & $\alpha$=0.1 & 757$\pm$155   & 951$\pm$202 & 1147$\pm$130    & 861$\pm$328  & \textbf{654$\pm$96}   \\ \cline{2-7} 
                                                                                         & $\alpha$=0.2 & 746$\pm$95     & 762$\pm$105   & 741$\pm$111    & 803$\pm$220           & \textbf{475$\pm$110}   \\ \cline{2-7} 
                                                                                         & $\alpha$=0.5 & 426$\pm$67     & 537$\pm$115   & 579$\pm$140   & 1080$\pm$309           & \textbf{384$\pm$74}   \\ \hline
\multirow{3}{*}{$\mathbb{E}[QCID](10^{-2})$} & $\alpha$=0.1 & 1.01$\pm$0.01 & 8.20$\pm$0.21  & 12.36$\pm$0.26 & 7.09$\pm$2.27  & \textbf{0.62$\pm$0.20} \\ \cline{2-7} 
                                                                                         & $\alpha$=0.2 & 0.93$\pm$0.03  & 7.54$\pm$0.27 & 10.6$\pm$0.48  & 5.93$\pm$1.01 & \textbf{0.51$\pm$0.12} \\ \cline{2-7} 
                                                                                         & $\alpha$=0.5 & 0.72$\pm$0.03  & 5.87$\pm$0.24 & 7.36$\pm$0.57  & 6.47$\pm$0.77 & \textbf{0.36$\pm$0.04} \\ \hline
\end{tabular}}
\vspace{3mm}
\caption{The communication rounds required for targeted test accuracy and the averaged QCID values. The targeted test accuracy is $45\%$ for $\alpha=0.1$, $47\%$ for $\alpha=0.2$ and $50\%$ for $\alpha=0.5$. The results are the mean and the standard deviation over 4 different random seeds.}
\label{roundqcid1}
\end{table*}

\subsection{Summary}

We have identified that the primary cause of performance degradation in federated learning (FL) under non-IID data conditions is class imbalance. To address this, we propose Federated Class-balanced Sampling (Fed-CBS), a client sampling mechanism that leverages probability engineering to select clients in a way that reduces class imbalance. Our extensive experiments demonstrate that Fed-CBS significantly outperforms existing approaches and achieves performance comparable to the ideal scenario where all available clients participate in training. We also provide theoretical convergence guarantees for Fed-CBS. In addition to Fed-CBS, we have conducted related research in federated learning \cite{zhang2024buildingfederatedgptfederated,hao2021towards,flop_qian2021,du2022rethinking,zhang2022next,tang2023fade,jia2025unlocking,yao2024federated,zhang2024mllm}, many of which apply probability engineering techniques and have been published in AI conferences.

\section{Knowledge Distillation}

Pre-trained language models, such as BERT \cite{devlin2018bert}, RoBERTa \cite{liu2019roberta}, GPT \cite{radford2018improving} and Electra \cite{clark2020electra} have achieved significant success on several different NLP tasks \cite{ding2019cognitive,wang2018glue} with fine-tuning. However, these models usually contain millions or billions of parameters, preventing their execution on resource-restricted devices. To deploy these models, Knowledge distillation (KD) is an effective compression technique to derive a smaller student model from a larger teacher model by transferring the knowledge embedded in the teacher's network. Previous KD methods typically store knowledge in the student's parameters and train the student by minimizing divergence between the student's and teacher's output prediction and intermediate activation distributions \cite{park2019relational,zhang2018deep}. However, the student's parametric memory is often limited and cannot be quickly expanded or revised. Moreover, after training, the teacher model's soft labels and activations, which contain essential task-specific knowledge, are not utilized by the student at inference time.

To address the issues mentioned above, we propose the \textit{Retrieval-augmented Knowledge Distillation} (ReAugKD) framework. ReAugKD introduces a non-parametric external memory in addition to the implicit parametric memory of the model and uses kNN retrieval to retrieve from this memory. The key intuition of ReAugKD is to enhance the effective capacity of the student by using an external memory derived from relevant task-specific knowledge of the teacher. While this external memory could include any task-specific knowledge, in this work, it is composed of the soft labels and embeddings generated by the teacher model. Our framework consists of an inference phase and a training phase.
In the inference phase, we aggregate the soft labels of those teacher embeddings in our memory that are most similar to the student embedding. We demonstrate the efficacy of our framework by achieving state-of-the-art results on the GLUE benchmark \cite{wang2018glue} with less than 3\% latency overhead over the baseline without retrieval augmentation.

Knowledge distillation can also be viewed as tackling distribution matching between the teacher and student. However, because the distributions in AI systems are often dynamic—driven by constantly changing user data or non-stationary inputs—our retrieval-augmented approach provides a more flexible way to capture and adapt to these evolving distributions. Specifically, ReAugKD comprises a training phase in which we teach the student how to retrieve relevant embeddings from the external memory effectively. In this phase, we introduce a novel relational KD loss that minimizes the divergence between both teacher–teacher and teacher–student embedding distributions, ensuring the student’s embedding space is well-aligned with the teacher’s for retrieval. Notably, this relational loss not only aligns the representation spaces but also improves generalization—even if retrieval augmentation is ultimately disabled. In other words, by infusing the student with the capability to draw on external memory, we enhance its ability to adapt and generalize, making ReAugKD a powerful illustration of probability engineering principles that cater to the naturally evolving distributions found in real-world knowledge distillation scenarios.


\subsection{Related Work}

\textbf{Knowledge distillation} KD can be broadly classified into task-specific KD, where the student model will be used for the same task as the teacher model \cite{mirzadeh2020improved,jin2019knowledge, zhang2018deep, sun2019patient} and task-agnostic KD where the student may be used for a different task, after finetuning on the new task \cite{jiao2019tinybert, sun2020mobilebert, sanh2019distilbert, wang2020minilmv2, zhang2018deep,TRPEDNN}. In this work, we show that ReAugKD can be applied to enhance task-specific distillation as well as when finetuning task-agnostic distilled models. Closest to our work is RKD \cite{park2019relational} that introduces a loss to transfer relational knowledge between teacher-teacher embedding and student-student embedding distributions. Our work differs in that we transfer relational knowledge between teacher-teacher embedding and teacher-student embedding distribution to enhance the student model's ability to retrieve from the external memory. MetaDistil \cite{zhou2022bert} is a strong task-specific distillation baseline that employs meta-learning to better transfer knowledge to the student. Unlike MetaDistill, we show that ReAugKD can significantly improve the student model's generalization without retraining the whole teacher with meta-learning.

\noindent \textbf{Retrieval-augmented language models} There has been growing interest in retrieval-augmented methods for Knowledge-Intensive generative NLP Tasks, such as text generation and question answering \cite{weston2018retrieve,lewis2020retrieval,guu2020retrieval,lin2022unsupervised}, where querying training examples during inference significantly improves likelihood. Closest to our work is BERT-kNN  \cite{kassner2020bert} which combines BERT with a kNN search over a large datastore of an embedded text collection, to improve cloze-style QA. In our work, we apply retrieval augmentation to enhance the capacity of student models during KD, and show improvement even on non-knowledge intensive tasks like GLUE.

\subsection{ReAugKD: Retrieval-Augmented Knowledge Distillation For Pre-trained Language Models}

\begin{figure*}[t]
    \centering
    \includegraphics[width=0.95\linewidth]{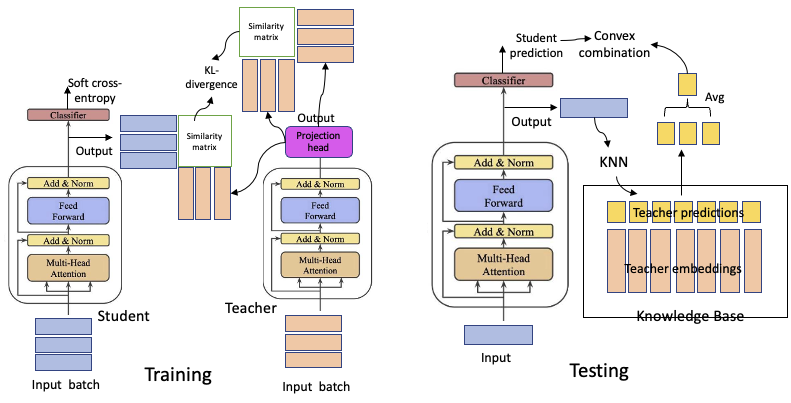}
    \vspace{5mm}
    \caption{Training and Inference (Testing) phases of Retrieval-augmented Knowledge Distillation (ReAugKD).}
    \label{framework}
\end{figure*}

\paragraph{Training Phase}

Our framework consists of two main phases, the training phase and the inference phase.
The training phase has two steps. In the first step, we prepare the teacher model for KD by adding a linear projection head $\mathcal{L}$ on the top of the teacher model encoder that has been finetuned for a specific downstream task. The input dimension of this projection head is the embedding dimension of the teacher. The output dimension is the embedding dimension of the student. We then freeze the other parameters of the teacher model and finetune the parameters in $\mathcal{L}$ with supervised contrastive loss \cite{khosla2020supervised}. This step a) reduces the dimension of the teacher's embeddings, to the student model dimension for retrieval, and b) uses supervised contrastive loss to derive a kNN classifier for BERT that is robust to natural corruptions, and hyperparameter settings \cite{li2021knn}. Fine-tuning $\mathcal{L}$ also greatly reduces the computational cost compared to retraining the whole teacher model \cite{zhou2022bert}.

In the second step, we perform KD by generating the teacher embeddings with $\mathcal{L}$ and teacher soft labels using the original teacher's classifier head for a batch of data. Then, we use the loss function we proposed in Section \ref{loss} to train our student model.

\paragraph{Loss function}\label{loss}

We present some mathematical notations to introduce our loss function. Given a batch of data $\{d_i\}, i=1,2,\cdots, N$, where $N$ is the batch size, we denote the embedding generated by the teacher's projection head as $z_i$ and the soft labels generated by the teacher's classifier as $\bar{y}_i$. Similarly, we adopt $x_i,{y}_i$ to denote the student's embeddings and predictions. Then we construct a probability distribution $q_{i,j}$ over each teacher's embeddings $z_j$ to capture the similarity with respect to an anchor point $z_i$,
\vspace{-1mm}
\begin{align}
    q_{i,j}=\frac{\exp{(z_i\cdot z_j)/\tau}}{\sum_{k=1}^{N}\exp{(z_i \cdot z_k)/\tau}},
\end{align} where the $\tau$ stands for temperature. Note that $\sum_{j=1}^N q_{i,j}=1$.  $q_{i,j}$ reflects the cosine distance relational knowledge among different embeddings generated by the teacher model in the batch. If $z_j$ is closer to $z_i$, cosine distance, $q_{i,j}$ will be larger. Similarly, given a student's embedding $x_i$ as an anchor point, we formulate another probability distribution $\bar{q}_{i,j}$ over each teacher's embeddings $z_j$ of the data in the batch.
\vspace{-1mm}
\begin{align}
    \bar{q}_{i,j}=\frac{\exp{(x_i\cdot z_j)/\tau}}{\sum_{k=1}^{N}\exp{(x_i \cdot z_k)/\tau}}.
\end{align}The $\bar{q}_{i,j}$ reflects the cosine distance relationship between different embeddings generated by the teacher model and the student's embedding. Our loss function aims to minimize the divergence of these two distributions $\bar{q}_{i,j}$ and ${q}_{i,j}$ since the teacher model is a strong kNN classifier after finetuning with supervised contrastive loss function in the first step of our training. In the ideal case, given a student's embedding $x_i$, the student retriever should retrieve the same set of embeddings as the corresponding teacher's embedding $z_i$. We adopt KL divergence to measure that divergence. In addition, we adopt the commonly-used cross-entropy loss to calculate the divergence between the student's prediction $y_i$ and the teacher's prediction $\bar{y}_i$.

Our loss function can be formulated as 
\begin{align}
    CE(y_i,\bar{y}_i) + \alpha KL(q_{i,j},\bar{q}_{i,j}),
\end{align} where $CE$ is the cross entropy loss and $KL$ is KL-divergence. $\alpha$ is  the hyperparameter controlling the trade-off between the two losses.

\paragraph{Inference Phase}
After training, we construct a knowledge base (KB) comprising of projected teacher embeddings and predictions. 
Given new data $d_i$ at inference time, we obtain $(x_i,{y}_i)$ using the student model. and use the HNSW algorithm \cite{malkov2018efficient} to derive the $K$ nearest teacher's embeddings and their corresponding soft labels $\{(z_k,\bar{y}_k)\}_{i= 1, 2, \cdots, K}$ from the KB. Then we compute the weighted average of these soft labels $Avg(\{\bar{y}\})_i$ based on $\bar{q}_{i,k}$
\vspace{-5mm}
     {  \begin{align*}
        Avg(\{y\})_i=\sum_{k=1}^K \frac{\bar{q}_{i,k}}{\sum_{k=1}^K \bar{q}_{i,k}}\bar{y}_k
    \end{align*}}
We derive a new prediction $\bar{y}_i^\prime$ for $d_i$ with $Avg(\{\bar{y}\})_i$.
    \begin{align*}
        \bar{y}_i^\prime = \beta \bar{y}_i + (1-\beta) Avg(\{\bar{y}\})_i,
    \end{align*}
    $\beta$ is the hyperparameter controlling the trade-off between the two predictions.

\subsection{Experimental Results}
We apply our method to distill BERT-Base \cite{devlin2018bert} into a 6-layer BERT with a hidden size of 768. We evaluate our proposed approach, ReAugKD, on the GLUE benchmark \cite{wang2018glue}. These datasets can be broadly divided
into three families of problems: single-set tasks that include linguistic acceptability (CoLA) and sentiment analysis (SST-2), similarity, and paraphrasing
tasks (MRPC and QQP); inference tasks that
include Natural Language Inference (MNLI and RTE); and Question Answering (QNLI). We compare our method with vanilla KD \cite{hinton2015distilling}, TAKD \cite{mirzadeh2020improved}, RCO \cite{jin2019knowledge}, RKD \cite{park2019relational}, DML \cite{zhang2018deep}, PKD \cite{sun2019patient} ProKT \cite{shi2020learning}, SFTN \cite{park2021learning} and MetaDistil \cite{zhou2022bert}. Following similar setting as MetaDistill, we perform a grid search over the sets of the weight of KD loss from \{0.9, 0.99\}, the predictions weight $\beta$ from \{0, 0.1, ... 1\} and the top-$k$ from 1 to 20. We set the student learning rate to 2e-5 and the batch size to 64.
\vspace{-1mm}
\paragraph{Experimental Results on GLUE}
We report the experimental results on the development set of the six GLUE tasks in Table \ref{tab:tab1}. Notably, our method achieves start-of-the-art results on five out of the six datasets with an average improvement of $0.34$\% over the previous best KD method MetaDistil \cite{zhou2022bert}. Although MetaDistil achieves slightly better performance on the MRPC dataset, our method has the advantage of not needing to conduct meta-learning on the whole large teacher model, which significantly increases extra training cost in terms of time and memory \cite{zhou2022bert}. In addition, we also observe a performance gain of $0.37$\% with the retrieval component of ReAugKD as compared to ReAugKD without retrieval which verifies the benefit of retrieval augmentation in our approach. Even without the retrieval process, the student model trained by our designed loss can still achieve comparable performance to MetaDistill on most datasets. Since our loss is designed to improve the student retrieval function, this demonstrates the importance of retrieval capability in KD.

\begin{table*}[!htbp]
\centering
\resizebox{1\columnwidth}{!}{%
\begin{tabular}{ccccccccl}
\hline
\multirow{2}{*}{Method}                                    & \multirow{2}{*}{\#Param} & \multicolumn{7}{c}{GLUE}                                                                                                                                                                                                                                                                                                                                     \\ \cline{3-9} 
                                                           &                          & \begin{tabular}[c]{@{}c@{}}CoLA\\ (8.5k)\end{tabular} & \begin{tabular}[c]{@{}c@{}}QNLI\\ (105k)\end{tabular} & \begin{tabular}[c]{@{}c@{}}QQP\\ (364k)\end{tabular} & \begin{tabular}[c]{@{}c@{}}RTE\\ (2.5k)\end{tabular} & \begin{tabular}[c]{@{}c@{}}SST-2\\ (67k)\end{tabular} & \begin{tabular}[c]{@{}c@{}}MRPC\\ (3.7k)\end{tabular} & Avg            \\ \hline
BERT-Base (teacher) \cite{devlin2018bert} & 110M                     & 58.9                                                  & 91.2                                                  & 91.4                                                 & 71.4                                                 & 93.0                                                  & \multicolumn{1}{c|}{87.6}                             & 82.25          \\
BERT-6L (student)\cite{turc2019well}      & 66M                      & 53.5                                                  & 88.6                                                  & 90.4                                                 & 67.9                                                 & 91.1                                                  & \multicolumn{1}{c|}{84.4}                             & 79.32          \\ \hline
\multicolumn{9}{c}{Task-specific Distillation}                                                                                                                                                                                                                                                                                                                                                                                                       \\ \hline
KD \cite{hinton2015distilling}            & 66M                      & 54.1                                                  & 89.2                                                  & 90.9                                                 & 67.7                                                 & 91.2                                                  & \multicolumn{1}{c|}{85.2}                             & 79.72          \\
PKD  \cite{sun2019patient}                & 66M                      & 54.5                                                  & 89.5                                                  & 90.9                                                 & 67.6                                                 & 91.3                                                  & \multicolumn{1}{c|}{84.7}                             & 79.75          \\
TinyBERT w/o DA   \cite{jiao2019tinybert} & 66M                      & 52.4                                                  & 89.8                                                  & 90.6                                                 & 67.7                                                 & 91.9                                                  & \multicolumn{1}{c|}{86.5}                             & 79.82          \\
RCO   \cite{jin2019knowledge}             & 66M                      & 53.6                                                  & 89.7                                                  & 90.6                                                 & 67.6                                                 & 91.4                                                  & \multicolumn{1}{c|}{85.1}                             & 79.67          \\
TAKD    \cite{mirzadeh2020improved}       & 66M                      & 53.8                                                  & 89.6                                                  & 90.7                                                 & 68.5                                                 & 91.4                                                  & \multicolumn{1}{c|}{85.0}                             & 79.83          \\
RKD      \cite{park2019relational}        & 66M                      & 53.4                                                  & 89.5                                                  & 90.9                                                 & 68.6                                                 & 91.7                                                  & \multicolumn{1}{c|}{86.1}                             & 80.03          \\
DML    \cite{zhang2018deep}               & 66M                      & 53.7                                                  & 89.6                                                  & 90.3                                                 & 68.4                                                 & 91.5                                                  & \multicolumn{1}{c|}{85.1}                             & 79.77          \\
ProKT \cite{shi2020learning}              & 66M                      & 54.3                                                  & 89.7                                                  & 90.9                                                 & 68.4                                                 & 91.3                                                  & \multicolumn{1}{c|}{86.3}                             & 80.15          \\
SFTN  \cite{park2021learning}             & 66M                      & 53.6                                                  & 89.5                                                  & 90.4                                                 & 68.5                                                 & 91.5                                                  & \multicolumn{1}{c|}{85.3}                             & 79.80          \\
MetaDistil  \cite{zhou2022bert}           & 66M                      & 58.6                                                  & 90.4                                                  & 91.0                                                 & 69.4                                                 & 92.3                                                  & \multicolumn{1}{c|}{\textbf{86.8}}                    & 81.42          \\
ReAugKD (ours)                                              & 66M                      & \textbf{59.4}                                         & \textbf{90.7}                                         & \textbf{91.24}                                       & \textbf{70.39}                                       & \textbf{92.5}                                         & \multicolumn{1}{c|}{86.3}                             & \textbf{81.76} \\
ReAugKD w/o retrieval                                      & 66M                      & 59.1                                                  & 90.6                                                  & 91.21                                                & 69.31                                                & 92.3                                                  & \multicolumn{1}{c|}{85.8}                             & 81.39          \\ \hline
\end{tabular}
}
\vspace{5mm}
\caption{Experimental results of ReAugKD and other previous works on the development set of GLUE. Numbers under each dataset
indicate the number of training samples. The results of the baselines are from \cite{zhou2022bert}. We report Matthew's correlation coefficient for CoLA and accuracy for other datasets.}
\label{tab:tab1}
\end{table*}

\subsection{Summary}

Our work demonstrates how probability engineering can transform knowledge distillation by explicitly modeling and adapting the distributions at play. In particular, we propose ReAugKD, a framework that augments a student’s parametric memory with a non-parametric memory built from teacher outputs. This memory enables the student to retrieve distributional information from the teacher at inference time, improving generalization on the GLUE benchmark. Furthermore, we introduce a novel relational loss function that aligns teacher–teacher and teacher–student embedding distributions, ensuring the student model is well-prepared to leverage retrieval-based signals. Even when retrieval is disabled at inference, this loss leads to significant improvements in generalization. Finally, we evaluate ReAugKD’s overhead in practical settings, showing that using approximate kNN retrieval imposes less than a $3\%$ latency increase—an acceptable cost for considerable accuracy gains. Overall, our approach highlights the power of probability engineering in handling the evolving and often complex distributions inherent in knowledge distillation scenarios. We have also extended this probability engineering approach in another work, where large multimodal models perform pretraining for smaller ones to fully leverage open-source raw data \cite{zhang2024mllm}, further underscoring the versatility of the proposed framework in handling evolving and often complex distributions in knowledge distillation scenarios.

\chapter{Probability Engineering in Generative AI}
\label{generative_AI}
\vspace{\baselineskip}
\section{High-quality Text Generation with Large Language Models}
\begin{figure*}[t!]
  \centering
  \includegraphics[width=0.7\textwidth]{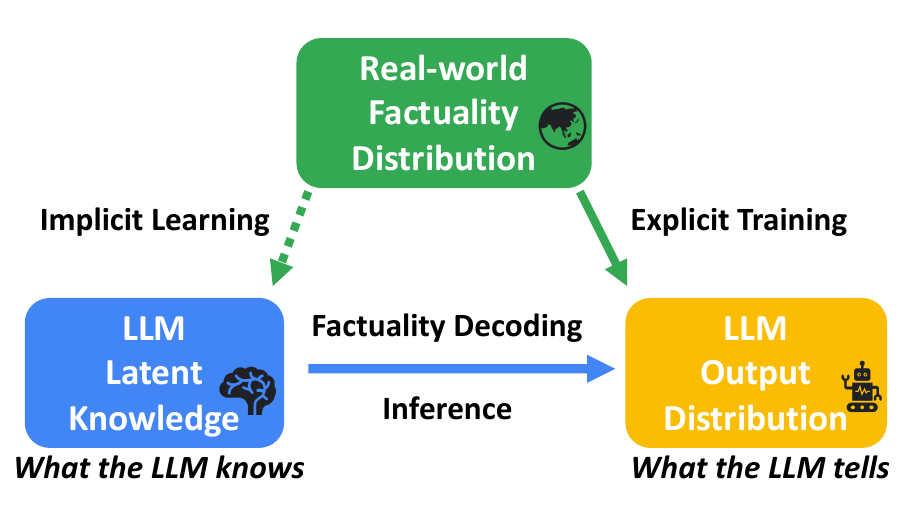} 
  \caption{Factuality decoding overview.} \label{fig:three-body}
\end{figure*}
Large Language Models (LLMs) have achieved remarkable breakthroughs in recent years, demonstrating exceptional performance across various domains~\cite{llama3modelcard,anil2023palm,openai-chatgpt,GPT4report,team2023gemini,touvron2023llama,touvron2023llama2}. However, a significant challenge associated with LLMs is their tendency to hallucinate or distort the truth, resulting in outputs that are not factual~\cite{huang2023survey,ji2023survey,zhang2023siren}. This issue of hallucination undermines the reliability and trustworthiness of LLMs in practical applications. A popular strategy for improving the LLM factuality involves refining the decoding process~\cite{shi2024thorough,welleck2024decoding}. Decoding focuses on how the model selects the next token during the generation process, which can significantly influence the factual accuracy of the output. The decoding methods can be cost-effective since (a) they do not rely on external knowledge and (b) no additional training is required. Furthermore, decoding methods can be synergistically combined with other techniques aimed at improving the LLM factuality, such as retrieving information from external knowledge bases~\cite{lei2023chain,lewis2020retrieval}, various fine-tuning strategies for better alignment~\cite{tian2024finetuning,touvron2023llama2}, or ensemble learning methods~\cite{du2024improving}.

Recent studies~\cite{kadavath2022language,NEURIPS2023_ITI,saunders2022self,wang2020language} suggest that LLMs sometimes have learned the factual content based on extensive pretraining or fine-tuning, although they fail to produce the correct answer when a user queries the model. This has inspired the development of several factuality decoding methods \cite{chuang2024dola,NEURIPS2023_ITI,li2022contrastive,zhang2023alleviating} to reveal what the model implicitly "knows." Figure \ref{fig:three-body} summarizes the underlying mechanism of these factuality decoding methods. The LLMs' output distribution is derived by applying the softmax function to the output logits from the final layer. During the training phase, this distribution is optimized based on the real-world factuality distribution represented by the training dataset. However, during the inference phase, "what the LLM tells" might still contain factual errors, which implies a discrepancy between the output distribution and the real-world factuality distribution. While the real-world distribution remains inaccessible during the inference phase, the model's latent knowledge ("what the model knows") may have implicitly learned some factual content correctly during the training phase \cite{kadavath2022language,wang2020language}. Therefore, a key challenge for factuality decoding strategies lies in effectively harnessing the latent knowledge embedded within LLMs to refine the output distribution (logits) during inference.

\begin{figure*}[h]
    \centering
    \includegraphics[width=0.97\textwidth]{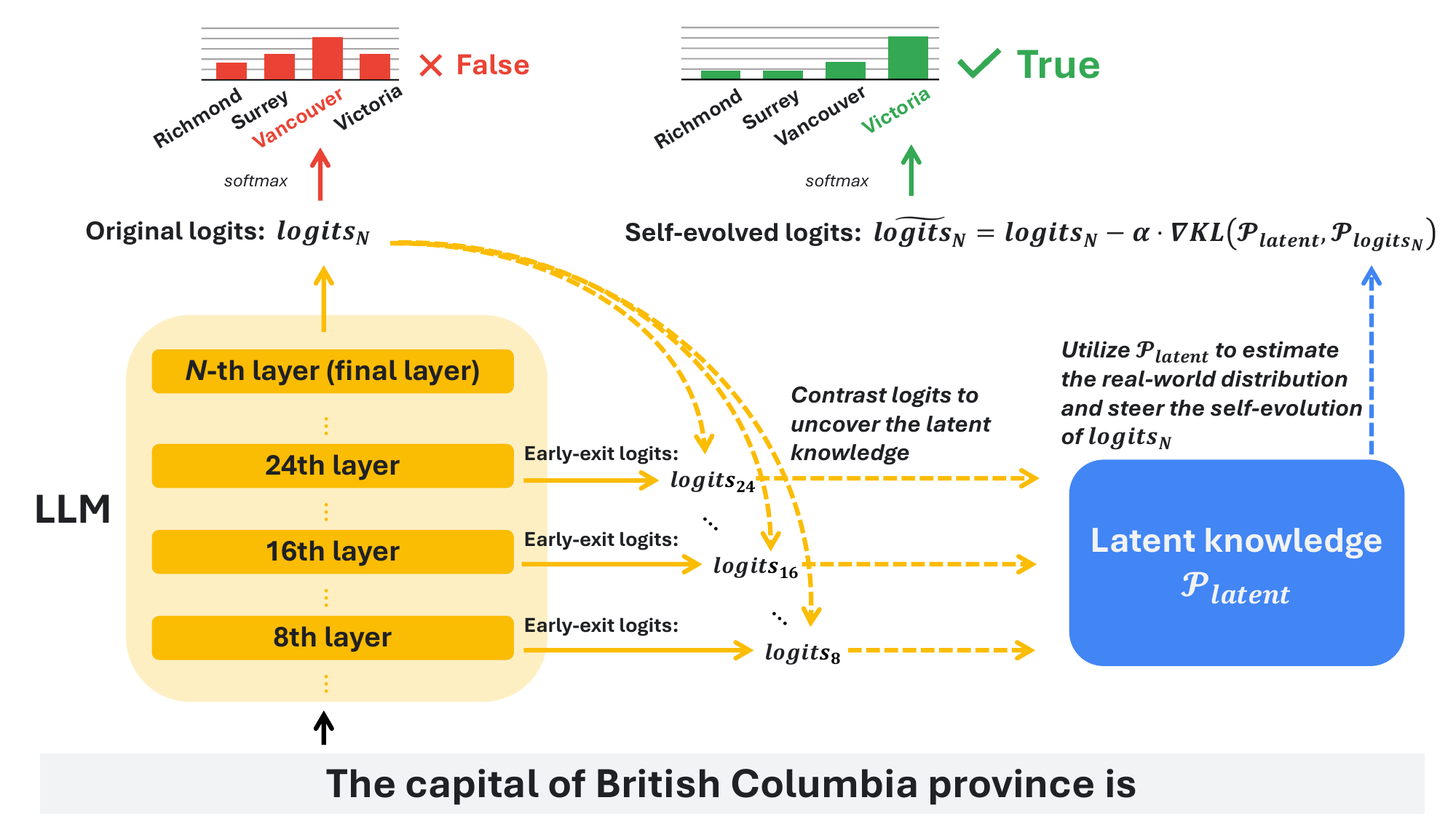}
    \vspace{5mm}
    \caption{Illustration of our Self Logits-Evolution Decoding (SLED) workflow. }
    \label{fig:sled_framework}
\end{figure*}

To address this challenge, we propose \textbf{S}elf \textbf{L}ogits \textbf{E}volution \textbf{D}ecoding (SLED), a novel factuality decoding approach that leverages the latent knowledge within LLMs by contrasting the final layer’s logits with early layers' logits. {During the decoding process, as LLMs progress from early to final layers, they progressively incorporate factual information stored in each layer into the output.} SLED tracks this evolution process to unearth the latent knowledge within LLMs, and enables the “self-evolution” of the output distribution further to align it more closely with real-world facts. Furthermore, our approach recognizes that the latent knowledge within LLMs, while valuable, may not always be perfect. Therefore, instead of simply replacing the original outputs with this latent knowledge, SLED integrates it into the original logits through an operation similar to “single-step gradient descent” over the output logits during the inference time. This operation minimizes the Kullback-Leibler (KL) divergence between the latent knowledge distribution and the output distribution, effectively balancing the two and mitigating potential drawbacks such as overfitting or biased outputs. Figure \ref{fig:sled_framework} illustrates the SLED workflow, highlighting how  SLED optimizes the output logits, leading to a more factual output distribution. We evaluate SLED on various LLMs (e.g., LLaMA 2 \cite{touvron2023llama2}, LLaMA 3 \cite{llama3modelcard}, Gemma \cite{Mesnard2024GemmaOM}) and benchmarks to demonstrate its state-of-the-art performance in layer-wise contrastive decoding methods. In summary, our main contributions are:

\subsection{Background}

There have been many advances in improving training and inference to develop better out-of-the-box LLMs~\cite{touvron2023llama, touvron2023llama2, llama3modelcard, team2023gemini, GPT4report, li2023textbooks,zhang2024buildingfederatedgptfederated,wang2024coreinfer,joren2024sufficient}. Unfortunately, LLMs still suffer from hallucinations and producing non-factual text. This has led researchers to develop many methods to improve factuality.

\paragraph{Retrieval, Fine-tuning, and Preferences.} Many techniques use additional knowledge graphs or fine-tuning data to increase factuality by updating the model parameters for this goal. One method is Retrieval-Augmented Generation (RAG) to use external knowledge to improve generation~\cite{chen2024benchmarking, cheng2024lift, ding2024survey, lewis2020retrieval}. Another option is to use post-generation retrieval and editing for improving attribution~\cite{gao2023rarr}. Other directions that use additional training or preference data are supervised fine-tuning (SFT)~\cite{ovadia2023fine, tian2024finetuning}, RLHF~\cite{ouyang2022training}, DPO~\cite{rafailov2024direct} or self-rewarding~\cite{yuan2024self}. Complementary to these approaches, we wish to improve the LLM output distribution directly without needing any additional data.

\paragraph{Decoding and Factuality Decoding} For each prefix, the LLM generates a probability distribution for the next token on a fixed vocabulary list, and a decoding method determines how the next token is derived based on the estimated distribution. Decoding methods were initially developed to enhance the fluency and coherence of text generation, such as Beam Search (BS), which maintains the $k$ most probable sequences at each time step. Common decoding methods also include Diverse Beam Search (DBS) \cite{Vijayakumar_Cogswell_Selvaraju_Sun_Lee_Crandall_Batra_2018}, Contrastive Decoding \cite{li2022contrastive}, Top-p Sampling \cite{holtzman2019curious} and so on.

Recently, the potential of decoding has extended beyond merely improving text readability, with some factuality decoding methods being proposed. These methods modify the generation process to focus on truthful statements rather than unsupported claims during the inference phase, aiming to reduce hallucinations. Notable recent works include Inference-Time Intervention (ITI) \cite{NEURIPS2023_ITI}, Induced-Contrastive Decoding \cite{zhang2023alleviating}, Decoding by Contrasting Layers (DoLa) \cite{chuang2024dola} and so on. ITI adjusts model activations during inference by following learned directions across a limited number of attention heads to improve truthfulness. Some researchers have extended previous Contrastive Decoding \cite{li2022contrastive} methods to improve factual accuracy, such as Frustratingly Easy Model Decoding \cite{yang2023frustratingly} and Induced-Contrastive Decoding \cite{zhang2023alleviating}, leveraging differences between expert and amateur models. Most closely related to our work is DoLa, which also employs contrasting logits from different layers. However, significant distinctions exist: Firstly, our method diverges in how to utilize those differences between logits to extract latent knowledge. Secondly, whereas DoLa directly substitutes the original output distribution with the latent knowledge distribution, our approach recognizes potential inaccuracies in this estimated distribution and adopts gradient descent within an optimization framework to integrate the model's latent knowledge with the original output.

\subsection{Self Logits Evolution Decoding}
A large language model,  equipped with $\mathit{N}$ layers and a vocabulary $\mathcal{V} = [v_1, v_2, \ldots, v_d]$, typically generates text in the next-token prediction fashion. For each given prefix, the
model computes the logits at the final ($\mathit{N}$-th) layer, $\mathit{logits}_\mathit{N} \triangleq (\ell_{(1,\mathit{N})}, \ell_{(2,\mathit{N})}, \ldots, \ell_{(d,\mathit{N})})$,  which are obtained by applying a linear transformation to the hidden states of the final layer, projecting the high-dimensional hidden state vectors onto the space of the vocabulary size. Subsequently, the output distribution $\mathcal{P}_{\mathit{logits}_\mathit{N}}$ at the final ($\mathit{N}$-th) layer for the next token is derived by applying softmax function on the logits, 
{\begin{align*}
    \mathcal{P}_{\mathit{logits}_\mathit{N}} \triangleq (p_{(1,\mathit{N})}, \ldots, p_{(d,\mathit{N})}) = softmax\left({\mathit{logits}_\mathit{N}}/{\tau}  \right),
\end{align*} }
where $\tau$ is the temperature parameter. Therefore, for each $p_{(i,\mathit{N})}~( 1 \leq i \leq d) $, we have \vspace{-3pt}
\begin{align*}
    p_{(i,\mathit{N})} = {\exp (\ell_{(i,\mathit{N})}/\tau)}/{S}, \ \text{where $\ S= \sum\nolimits_{j=1}^d \exp (\ell_{(j,\mathit{N})}/\tau).$} 
\end{align*}Similarly, we can also derive the logits from early layers by applying the same linear transformation mentioned above to their hidden states. For any early layer $n~(n < N$), we denote its logits as $\mathit{logits}_\mathit{n} \triangleq (\ell_{(1,\mathit{n})}, \ldots, \ell_{(d,\mathit{n})})$ and the corresponding distribution as $\mathcal{P}_{\mathit{logits}_\mathit{n}} \triangleq (p_{(1,\mathit{n})}, \ldots, p_{(d,\mathit{n})})$.

\begin{figure}[t]
    \centering
        \includegraphics[width=0.95\textwidth]{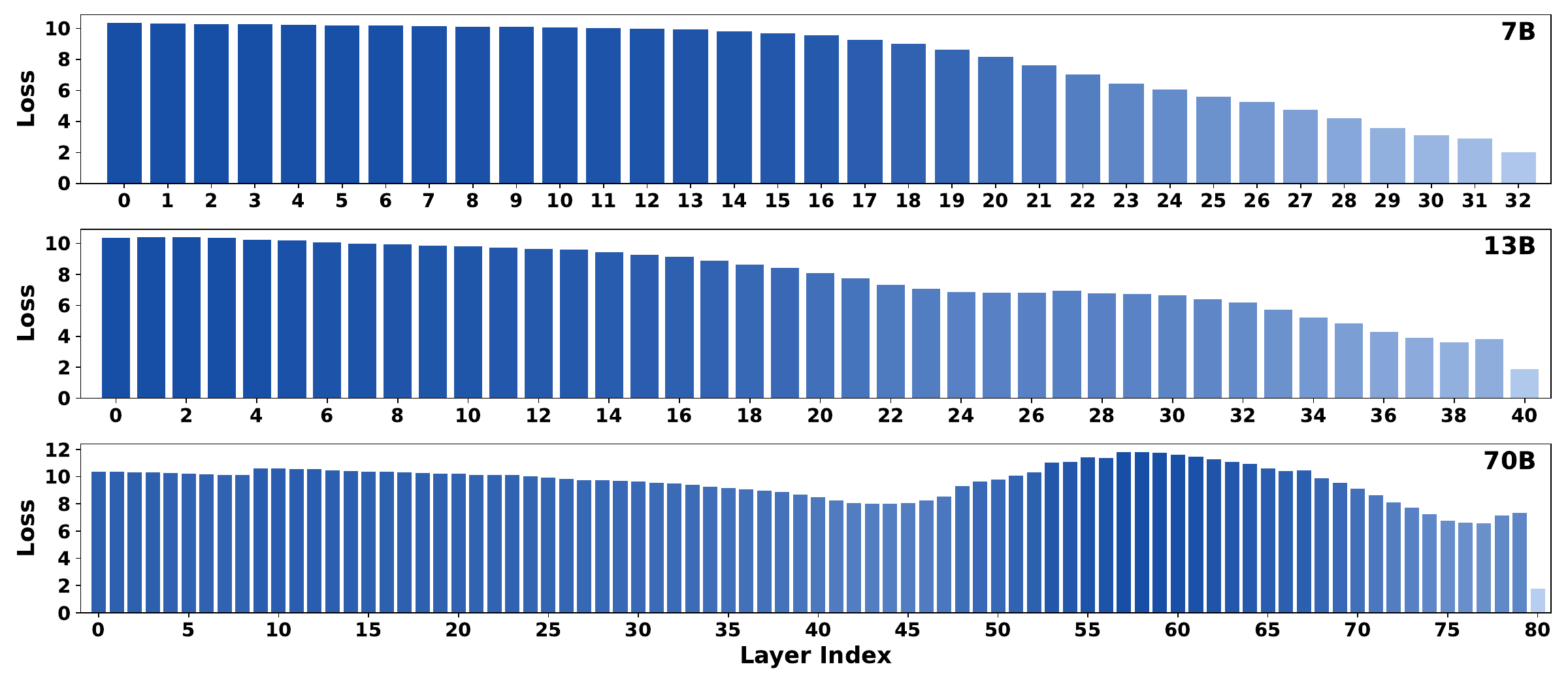}
        \vspace{5mm}
        \caption{The results on the next-token predictions of three LLaMA-2-base models using the logits from each layer individually. }
        \label{fig:layers_7B_13B}
\end{figure}

\paragraph{Logits Evolution}\label{logits_evolution}

To improve factual accuracy, it is crucial that the correct token \(v_i\) receives a higher value of $\mathit{logits}_\mathit{N}$ to ensure a higher probability value $p_{(i,\mathit{N})}$ in the output distribution $\mathcal{P}_{\mathit{logits}_\mathit{N}}$. From a mathematical perspective, this means aligning the model's output distribution $\mathcal{P}_{\mathit{logits}_\mathit{N}}$  closely with the real-world factuality distribution $\mathcal{P}_{\mathit{real}}$. Specifically, we can formulate this goal as optimizing the following loss function $\mathcal{L}$ regarding the $\mathit{logits}$:
\begin{align}\label{training_evolution}
     \mathcal{L} ( \mathit{logits}) \triangleq  KL(\mathcal{P}_{\mathit{real}},\mathcal{P}_{\mathit{logits}}), \text{where}\ \mathit{logits} = (\ell_1, ...,\ell_d),\ \mathcal{P}_{\mathit{logits}}=softmax(\mathit{logits}/{\tau})
\end{align}
We describe the above optimization as \textbf{Logits Evolution}. Interestingly, the training of LLMs also aims at minimizing the divergence (typically the $\mathit{KL}$ divergence, as the training loss function is often the cross-entropy loss) between the ground truth $\mathcal{P}_{\mathit{real}}$ and the output distribution $\mathcal{P}_{\mathit{logits}_\mathit{N}}$. During the training phase, the logits evolution is driven externally by the real-world distribution $\mathcal{P}_{\mathit{real}}$ presented in the training dataset, and the corresponding solution is $\mathit{logits}=\mathit{logits}_\mathit{N}$. However, $\mathcal{P}_{\mathit{real}}$ is not accessible during the inference phase. To address this challenge, SLED utilizes the model's latent knowledge to estimate $\mathcal{P}_{{real}}$ and enables "self-evolution" of the logits. We denote the estimation as $\mathcal{P}_{\mathit{latent}}$ and the self logits evolution can be achieved by the following gradient-descent operation:
\begin{align}\label{optimized_lr_delta}
     \widetilde{\mathit{logits}}_\mathit{N} = \mathit{logits}_\mathit{N} - \alpha \cdot \nabla_{\mathit{logits}_\mathit{N}} KL(\mathcal{P}_{\mathit{latent}}, \mathcal{P}_{\mathit{logits}_\mathit{N}}). 
\end{align}
The parameter \(\alpha\), termed the \textbf{Evolution Rate}, governs the magnitude of adjustments applied to $\mathit{logits}_\mathit{N}$ in the direction of the gradient $\nabla_{\mathit{logits}_\mathit{N}} KL(\mathcal{P}_{\mathit{latent}}, \mathcal{P}_{\mathit{logits}_\mathit{N}})$. In the following Section \ref{method_starting_point} and \ref{estimation_med}, we discuss how we derive the $\mathcal{P}_{\mathit{latent}}$ as the estimation of the real-world distribution $\mathcal{P}_{\mathit{real}}$.

\begin{figure}[t]
    \centering
        \includegraphics[width=1\textwidth]{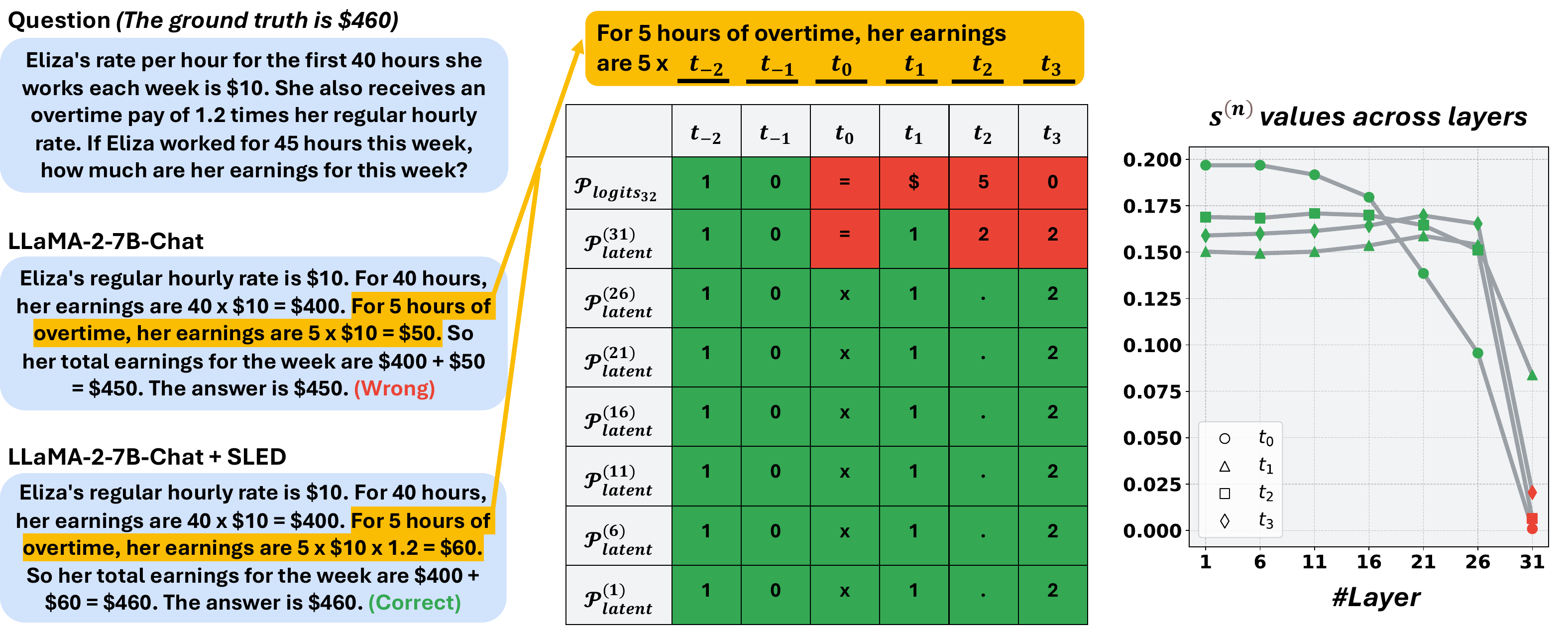}
        \vspace{5mm}
            \caption{An example from GSM8K demonstrating SLED's mechanism. }  \label{fig:pipeline}
\end{figure}

\paragraph{Estimate $\mathcal{P}_{\mathit{real}}$ by Tracking the Logits Evolution Direction throughout Layers} \label{method_starting_point}

The core principle of our method involves leveraging the difference between each early layer's logits and the final layer's logit, $\mathit{logits}_\mathit{n} - \mathit{logits}_\mathit{N}$ to approximate the gradient of \(KL(\mathcal{P}_{\mathit{real}},\mathcal{P}_{\mathit{logits}})\) at \(\mathit{logits}= \mathit{logits}_\mathit{n}\). Then we estimate $\mathcal{P}_{\mathit{real}}$ based on this approximation. 

This is inspired by a new perspective of interpreting the training phase of LLMs as the evolution of logits described in Problem \ref{training_evolution}. As mentioned above, the solution derived by the training phase is the final layer's logits $ \mathit{logits}= \mathit{logits}_\mathit{N}$, since the final layer's $\mathit{logits}_\mathit{N}$ directly engage with the real-world distribution $\mathcal{P}_{\mathit{real}}$ through the loss function in training. This implies that we can generally consider the final logits $\mathit{logits}_\mathit{N}$ to be a better solution than the logits from an early layer $\mathit{logits}_\mathit{n}$, with \( KL(\mathcal{P}_{\mathit{real}},\mathcal{P}_{\mathit{logits}_\mathit{N}}) < KL(\mathcal{P}_{\mathit{real}},\mathcal{P}_{\mathit{logits}_\mathit{n}}) \). We present some examples in Figure \ref{fig:layers_7B_13B} to demonstrate this. This analysis is performed on 200 true claims from the FACTOR dataset. The results verify that the logits distribution at the final layer is closer to the real-world distribution than all the early layers in terms of KL divergence. Based on this discussion, if we contrast the final layer's logits with the early layer's logits, we can consider the direction (orientation) of $\mathit{logits}_\mathit{n} - \mathit{logits}_\mathit{N}$ can approximately align with the direction of the gradient \(\nabla_{\mathit{logits}} KL(\mathcal{P}_\mathit{real}, \mathcal{P}_\mathit{logits}) |_{\mathit{logits}=\mathit{logits}_\mathit{n}}\). To further verify this motivation, we calculate the cosine similarity between $\mathit{logits}_\mathit{n} - \mathit{logits}_\mathit{N}$ and \(\nabla_{\mathit{logits}_\mathit{n}} KL(\mathcal{P}_\mathit{real}, \mathcal{P}_{\mathit{logits}_\mathit{n}})\) for thousands of tokens across different models in Figure \ref{fig:model_gradient_approx_cos}. We find that the majority of these values are positive, which means that the directions of these two vectors are close.

Hence, for each early layer \(n\), we propose to maximize the following function of cosine similarity and derive the \(\mathcal{P}^{(n)}_{\mathit{latent}}\) to estimate the $\mathcal{P}_{\mathit{real}}$.
\vspace{0.1cm}
\begin{align}\label{cos_approx_problem}
    \mathcal{P}^{(n)}_{\mathit{latent}} = \arg \max_{\mathcal{P}} \left( \text{CosSim} ( \mathit{logits}_\mathit{n} - \mathit{logits}_\mathit{N}, \nabla_{\mathit{logits}_\mathit{n}} KL(\mathcal{P}, \mathcal{P}_{\mathit{logits}_\mathit{n}}), 0 \right)
\end{align}
\vspace{-0.5cm}
\paragraph{Achieving the Self Logits Evolution in Three Phases}\label{estimation_med}

Based on the above analysis, we can introduce the procedures of SLED: First, we estimate \(\mathcal{P}^{(n)}_{\mathit{latent}}\) for each early layer \(n\) using the gradient approximation in Section \ref{method_starting_point}. Subsequently, we apply a weighted average on $\{\mathcal{P}^{(n)}_{\mathit{latent}}\}$ across all early layers $n < N$ to derive \(\mathcal{P}_{\mathit{latent}}\), which serves as the final estimation of the real-world distribution. Finally, we apply \(\mathcal{P}_{\mathit{latent}}\) in Equation \ref{optimized_lr_delta} to facilitate the self-evolution of \({\mathit{logits}}_\mathit{N}\), thereby derive the updated logits, $\widetilde{\mathit{logits}}_\mathit{N}$.
\begin{align*}
{\mathit{logits}}_\mathit{n} - {\mathit{logits}}_\mathit{N} &\overset{\text{in direction}}{\approx} \nabla_{\mathit{logits}_\mathit{n}} KL(\mathcal{P}_\mathit{real}, \mathcal{P}_{\mathit{logits}_\mathit{n}}) \\
&\xRightarrow[\text{Estimate}]{\textcolor[HTML]{4285F4}{\text{Phase 1}}} \ \mathcal{P}^{(n)}_{\mathit{latent}} \ \xRightarrow[\text{Ensemble}]{\textcolor[HTML]{4285F4}{\text{Phase 2}}} \ \mathcal{P}_{\mathit{latent}} \ \xRightarrow[\text{Self-evolution in Eq \ref{optimized_lr_delta}}]{\textcolor[HTML]{4285F4}{\text{Phase 3}}} \widetilde{\mathit{logits}}_\mathit{N}
\end{align*}

\textcolor[HTML]{4285F4}{Phase 1}: An exhaustive search for an exact solution to the complex optimization problem (Equation \ref{cos_approx_problem}) is computationally impractical. We can reduce the solution space by the following. Suppose the real-world factuality distribution dictates that the next word to be generated is the \(i\)-th token \(v_i\) from the vocabulary $\mathcal{V}$. Thus \(\mathcal{P}_{\mathit{real}} = \mathcal{P}_{e_i}\), where \(\mathcal{P}_{e_i}\) represents a standard basis vector (one-hot vector) with the \(i\)-th component set to 1 and all other components set to 0. Then, we can simplify the aforementioned optimization problem by limiting the solution space to $\{\mathcal{P}_{e_i} \}_{i=0}^{d}$ and decide which token $i$ should be selected. The corresponding gradient when $\mathcal{P}=\mathcal{P}_{e_i}$ has the following formulation.
\begin{proposition}
The gradient of $KL(\mathcal{P}_{e_i},\mathcal{P}_{\mathit{logits}})$ at \(\mathit{logits} =\mathit{logits}_\mathit{n}\) is:
{\begin{align}
    \nabla_{\mathit{logits}_\mathit{n}} KL(\mathcal{P}_{e_i}, \mathcal{P}_{\mathit{logits}_\mathit{n}}) =  (\mathcal{P}_{\mathit{logits}_\mathit{n}} - \mathcal{P}_{e_i})/{\tau}  = \left(p_{(1,\mathit{n})},\ldots, p_{(i,\mathit{n})} - 1, \ldots, p_{(d,\mathit{n})}\right)/{\tau}
 \label{gradeint_p_latent}
\end{align}}
\end{proposition}\vspace{-0.2cm}
We calculate the cosine similarity between the gradient $\nabla_{\mathit{logits}_\mathit{n}} KL(\mathcal{P}_{e_i}, \mathcal{P}_{\mathit{logits}_\mathit{n}})$ and the difference $\mathit{logits}_\mathit{n} - \mathit{logits}_\mathit{N}$ for each token in the vocabulary $\mathcal{V}$. Then we select the \(\mathcal{P}_{e_{i^*}}\) of which the gradient is closest to $\mathit{logits}_\mathit{n} - \mathit{logits}_\mathit{N}$ as the estimation \(\mathcal{P}^{(n)}_{\mathit{latent}}\). Mathematically, this involves selecting \(i^*\) according to the following criterion
\begin{align*}
    i^* = \arg \max_{1\leq i \leq d} \, \bar{m}^{(n)}_i, \ \text{where} \ \bar{m}^{(n)}_i = \max \left( \text{CosSim} ( \mathit{logits}_\mathit{n} - \mathit{logits}_\mathit{N}, \mathcal{P}_{\mathit{logits}_\mathit{n}} - \mathcal{P}_{e_i} ), 0 \right),
\end{align*}
and adopting \(\mathcal{P}^{(n)}_{\mathit{latent}} = \mathcal{P}_{e_{i^*}}\) as the "hard estimation" of \(\mathcal{P}_{\mathit{real}}\). Drawing from the concept of hard and soft targets in label smoothing and knowledge distillation, we further extend it to the "soft estimation",
\begin{align*}
   \mathcal{P}^{(n)}_{\mathit{latent}} =   ( m^{(n)}_1, \ldots , m^{(n)}_i, \ldots, m^{(n)}_d)/{m^{(n)}}, \ \text{where} \ m^{(n)}_i = (\bar{m}^{(n)}_i)^2 \  \text{and} \ m^{(n)}={\sum\nolimits_{i=1}^d m^{(n)}_i}
\end{align*}
 We square $\{\bar{m}^{(n)}_i\}$ to moderately amplify their differences. Prior studies prove that soft targets usually offer stronger generalization capabilities, more information, and more robustness to noise than hard targets \cite{hinton2015distilling,muller2019does,ClassificationSoftRobust,zhang2021delving}. Hence, we adopt the soft estimation in lieu of the hard estimation.
\textcolor[HTML]{4285F4}{Phase 2}: We ensemble \(\mathcal{P}^{(n)}_{\mathit{latent}}\) across all layers by computing a weighted average of the set \(\{\mathcal{P}^{(n)}_{\mathit{latent}}\}\) and adopt it as the final estimation of the $\mathcal{P}_{\mathit{latent}}$:
\begin{align*}
     \mathcal{P}_{\mathit{latent}} =  {\sum\nolimits_{n=0}^N s^{(n)} \mathcal{P}^{(n)}_{\mathit{latent}}}, \ \text{where} \ s^{(n)} = m^{(n)}/{(\sum\nolimits_{n=0}^N m^{(n)})}
\end{align*}
This estimation suggests that the weight $s^{(n)}$ of certain layer $n$ will be larger if the corresponding gradient approximation $\mathit{logits}_\mathit{n} - \mathit{logits}_\mathit{N}$ is more closely aligned with the gradients $\{\nabla_{\mathit{logits}_\mathit{n}} KL(\mathcal{P}_{e_i}, \mathcal{P}_{\mathit{logits}_\mathit{n}})\}$ for the tokens in the vocabulary. This in turn amplifies the influence of layer $n$ on the final estimation, which is a desirable effect in our method. One can further validate that for each component \(m_i\) in the final estimation \(\mathcal{P}_{\mathit{latent}} \triangleq (m_1, m_2, \ldots, m_d)\), the following relationship holds:
    $m_i = {\sum_{n=0}^N m^{(n)}_i}/{(\sum_{n=0}^N \sum_{j=1}^d m^{(n)}_j)}.$
    This property simplifies the description in Algorithm \ref{algo1}.      

\textcolor[HTML]{4285F4}{Phase 3}: Applying $\mathcal{P}_{\mathit{latent}}$ in Equation \ref{optimized_lr_delta} enables us to derive the gradient necessary for steering the self-evolution on the final layer's logits $\mathit{logits}_\mathit{N}$.
\begin{proposition}\label{proposition2}
The gradient of $KL(\mathcal{P}_{\mathit{latent}},\mathcal{P}_{\mathit{logits}})$ at \(\mathit{logits} =\mathit{logits}_\mathit{N}\) is:
{\begin{align*}
    \nabla_{\mathit{logits}_\mathit{N}} KL(\mathcal{P}_{\mathit{latent}}, \mathcal{P}_{\mathit{logits}_\mathit{N}}) = ( \mathcal{P}_{\mathit{logits}_\mathit{N}} - \mathcal{P}_{\mathit{latent}})/{\tau}  = \left(p_{(1,\mathit{N})}-m_1,\ldots, p_{(d,\mathit{N})}-m_d\right)/{\tau}
\end{align*}}
\end{proposition}
\vspace{-0.2cm}
Then we can derive the self-evolved logits $\widetilde{\mathit{logits}}_\mathit{N}$ 
\begin{align} 
     \widetilde{\mathit{logits}}_\mathit{N} \triangleq (\tilde{\ell}_{(1,\mathit{N})},\ldots, \tilde{\ell}_{(i,\mathit{N})}, \ldots, \tilde{\ell}_{(d,\mathit{N})}),\
     \text{where} \ \tilde{\ell}_{(i,\mathit{N})} = {\ell}_{(i,\mathit{N})} - {\alpha} (p_{(i,\mathit{N})}-m_i )/{\tau}.
\end{align} 
We provide a demo of SLED in Figure \ref{fig:pipeline}. SLED derives the estimations $\mathcal{P}^{(n)}_{\mathit{latent}}$ by contrasting final layer's logits $\mathit{logits}_\mathit{N}$ with early layers' logits $\{\mathit{logits}_\mathit{n}\}$. We list the token with the highest probability value from the $\mathcal{P}^{(n)}_{\mathit{latent}}$ for different early layers. As shown, SLED downplays incorrect tokens by assigning lower weights $s^{(n)}$ to the corresponding $\mathcal{P}^{(n)}_{\mathit{latent}}$. Conversely, if the estimation is correct, the weights are relatively larger. The parameter evaluation scale is set to 2.

\begin{algorithm}[t]
\caption{Self Logits Evolution Decoding} \label{algo1}
\begin{algorithmic}[1]

\STATE \textbf{Initialization:} LLM with $N$ layers, $inputs$, evolution rate $\alpha$, evolution scale $k>0$, $\eta \ll 0$, temperature parameter $\tau$, and the one-hoc vectors $\{\mathcal{P}_{e_i}\}$ defined in Section \ref{estimation_med}.

\STATE Feed the $inputs$ into the LLM to obtain the logits $\mathit{logits}_\mathit{n} = (\ell_{(1,\mathit{n})}, \ldots, \ell_{(d,\mathit{n})})$ and probabilities $\mathcal{P}_{\mathit{logits}_\mathit{n}} = (p_{(1,\mathit{n})}, \ldots, p_{(d,\mathit{n})}) = softmax(\mathit{logits}_\mathit{n}/\tau)$ at each layer $n$, where $n \leq N$.

\STATE Identify the tokens with the top-$k$ largest values in $\mathit{logits}_\mathit{N}$ and denote their indices by $I_k$.


\FOR{each early layer $n$, $(n<N)$}
    \STATE Compute differences for top-$k$ logits $\mathit{logits}_{\mathit{n}} - \mathit{logits}_{\mathit{N}}$.
    \STATE Calculate $m^{(n)}_i = \left[ \max \left( \text{CosSim} (\mathit{logits}_{\mathit{n}} - \mathit{logits}_{\mathit{N}}, \mathcal{P}_{\mathit{logits}_{\mathit{n}}} - \mathcal{P}_{e_i}), 0 \right) \right]^2, i \in I_k$.
\ENDFOR

\STATE Compute weighted average $m_i = \frac{\sum_{n=1}^N m^{(n)}_i}{\sum_{n=1}^N \sum_{j\in I_k} m^{(n)}_j}$ across different layers for each $i \in I_k$.

\FOR{each $i$ from $1$ to $d$}

\STATE Set $\tilde{\ell}_{(i,\mathit{N})} = {\ell}_{(i,\mathit{N})} - \frac{\alpha}{\tau} (p_{(i,\mathit{N})}-m_i )$ \textbf{if} {$i \in I_k$} \textbf{else} Set $\tilde{\ell}_{(i,\mathit{N})} = \eta \ll 0 $.

\ENDFOR

\STATE \textbf{Output:} The self-evolved logits are $\widetilde{\mathit{logits}}_\mathit{N} = (\tilde{\ell}_{(1,\mathit{N})},\ldots, \tilde{\ell}_{(i,\mathit{N})}, \ldots, \tilde{\ell}_{(d,\mathit{N})})$.

\end{algorithmic}
\end{algorithm}

\subsection{Experimental Results}
 
\paragraph{Benchmarks} 

We compare our method with baselines on several multiple-choice and open-ended generation tasks. For multiple-choice question tasks, we use the TruthfulQA \cite{lin-etal-2022-truthfulqa} and FACTOR (Wiki) ~\cite{muhlgay2023generating} datasets to assess the LLMs' factuality in short-answer/long-paragraph scenario, respectively. For open-ended generation tasks, we adopt TruthfulQA ~\cite{lin-etal-2022-truthfulqa}  and tasks involving chain-of-thought reasoning~\cite{wei2022chain}: StrategyQA~\cite{geva2021did} and GSM8K~\cite{cobbe2021training}.

\paragraph{Models \& Baselines} We evaluate the performance of SLED on six LLaMA-2 models \cite{touvron2023llama2} (\{7B,13B,70B\}-Base, \{7B,13B,70B\}-Chat), four LLaMA-3 family models \cite{llama3modelcard} (\{8B,70B\}-Base, \{8B,70B\}-IT), two Gemma models (2B,7B), two MoE models (Mixtral-8$\times$7B, Mixtral-8$\times$7B-IT)~\cite{jiang2024mixtralexperts}. We adopt the following baselines: 1) standard decoding (greedy decoding or sampling depending on the tasks), 2) DoLa \cite{chuang2024dola}, 3) Inference Time Intervention (ITI)~\cite{NEURIPS2023_ITI}, 4) Activation Decoding (AD)~\cite{chen2024context}, 5) Contrastive Decoding (CD) ~\cite{li2022contrastive}, and 6) Induce-then-Contrast Decoding (ICD)~\cite{zhang2023alleviating}.

\paragraph{Metrics} We adopt the factual accuracy evaluation implemented in ~\cite{chuang2024dola} for multiple-choice tasks and chain-of-thought reasoning tasks. For the open-ended generation task on TruthfulQA, we follow the evaluation procedure in ~\cite{chuang2024dola,lin-etal-2022-truthfulqa}, using “finetuned-GPT3-judge”s to measure the truthfulness, informativeness, and rejection rate of generated outputs respectively.

\begin{table}[t]
\centering
\caption{Comparison on LLaMA 2 model family. The best results are in bold for each dataset/metric. SLED outperforms DoLa and vanilla greedy decoding. 
}
\vspace{0.2cm}
\renewcommand{\arraystretch}{1.25}
\label{tab:mainresults_label}
\resizebox{\textwidth}{!}{\begin{tabular}{lccccccccccc}
\toprule
\multicolumn{1}{c}{\multirow{2}{*}{{Model\&Method}}} & \multicolumn{3}{c}{TruthfulQA (MC)}                           & \multirow{2}{*}{FACTOR}             & \multicolumn{4}{c}{TruthfulQA (Open-Ended)}                          & \multicolumn{2}{c}{CoT}         \\ \cmidrule(lr){2-4} \cmidrule(lr){6-9} \cmidrule(lr){10-11} 
\multicolumn{1}{c}{}                       & MC1        & MC2         & MC3              &                                     & \%Truth        & \%Info         & \%T*I       & \%Reject          & StrQA          & GSM8K          \\ \midrule
LLaMA-2-7B-Base                              & 33.17          & 59.42          & 31.78               & \multicolumn{1}{|c|}{58.15}          & 32.80          & 90.09          & {23.99}     & {8.45}            & \multicolumn{1}{|c}{60.96}          & 14.03          \\
+DoLa                               & 32.56          & \textbf{63.03} & {30.57}               & \multicolumn{1}{|c|}{62.49}          & 35.74          & \textbf{95.23} & {32.31}     & {2.57}            & \multicolumn{1}{|c}{60.61}          & 14.71          \\
+SLED (ours)                               & \textbf{34.15} & 62.57          & {\textbf{31.89}}  & \multicolumn{1}{|c|}{\textbf{67.27}} & \textbf{55.81} & 94.61          & {\textbf{52.87}} & {\textbf{0.12}} & \multicolumn{1}{|c}{\textbf{61.31}} & \textbf{15.01} \\ \midrule
LLaMA-2-7B-Chat                              & 35.62          & 57.46          & 32.07            & \multicolumn{1}{|c|}{56.78}          & 59.24          & 78.95          & {38.68}     & {17.50}           & \multicolumn{1}{|c}{63.67}          & 21.08          \\
+DoLa                                      & 33.41          & 61.93          & 30.35             & \multicolumn{1}{|c|}{56.65}          & 58.02          & 87.03          & {45.78}     & {13.10}           & \multicolumn{1}{|c}{64.32}          & 21.00          \\
+SLED (ours)                                & \textbf{37.08} & \textbf{63.86} & {\textbf{32.90}}  & \multicolumn{1}{|c|}{\textbf{64.70}} & \textbf{67.07} & \textbf{88.13} & {\textbf{55.69}} & {\textbf{11.02}} & \multicolumn{1}{|c}{\textbf{64.67}} & \textbf{21.15} \\ \midrule
LLaMA-2-13B-Base                             & 33.69          & 62.75          & 31.74             & \multicolumn{1}{|c|}{63.69}          & 31.21          & 91.55          & {23.26}     & {7.96}            & \multicolumn{1}{|c}{66.07}          & 28.66          \\
+DoLa                                      & 29.25          & 62.13          & 30.29               & \multicolumn{1}{|c|}{57.08}          & 37.58          & 92.41          & {30.11}     & {7.47}            & \multicolumn{1}{|c}{65.55}          & 18.88          \\
+SLED (ours)                                & \textbf{34.15} & \textbf{63.62} & {\textbf{31.89}}  & \multicolumn{1}{|c|}{\textbf{70.91}} & \textbf{38.31} & \textbf{94.85} & {\textbf{33.29}} & {\textbf{5.02}} & \multicolumn{1}{|c}{\textbf{66.81}} & \textbf{29.34} \\ \midrule
LLaMA-2-13B-Chat                             & 36.47          & 63.05          & \textbf{32.77}            & \multicolumn{1}{|c|}{62.06}          & 60.34          & 86.54          & {47.12}     & {13.59}           & \multicolumn{1}{|c}{69.87}          & 36.47          \\
+DoLa                                      & 34.52          & 63.24          & 31.48             & \multicolumn{1}{|c|}{58.08}          & 60.22          & 90.33          & {51.16}     & {9.67}            & \multicolumn{1}{|c}{67.90}          & 34.57          \\
+SLED (ours)                                & \textbf{37.09} & \textbf{63.75} & {32.60}  & \multicolumn{1}{|c|}{\textbf{67.50}} & \textbf{63.65} & \textbf{95.23} & {\textbf{58.87}} & {\textbf{5.26}} & \multicolumn{1}{|c}{\textbf{69.96}} & \textbf{36.54} \\ \midrule
LLaMA-2-70B-Base                             & 33.66          & 61.10          & 32.33             & \multicolumn{1}{|c|}{72.78}          & 55.45          & 62.55          & {18.48}     & {36.74}            & \multicolumn{1}{|c}{75.20}          & 56.33          \\
+DoLa                                      & 26.93          & 60.33          & 29.42               & \multicolumn{1}{|c|}{61.92}          & \textbf{60.95}          & 70.62          & {32.07}     & {17.72}            & \multicolumn{1}{|c}{73.45}          & 43.37          \\
+SLED (ours)                                & \textbf{35.13} & \textbf{64.92} & {\textbf{33.52}}  & \multicolumn{1}{|c|}{\textbf{77.49}} & {59.24} & \textbf{82.99} & {\textbf{43.70}} & {\textbf{13.10}} & \multicolumn{1}{|c}{\textbf{75.20}} & \textbf{57.09} \\ \midrule
LLaMA-2-70B-Chat                             & 35.98          & 64.18          & {32.99}            & \multicolumn{1}{|c|}{69.07}          & 49.57          & 81.27          & 31.33     & {29.13}           & \multicolumn{1}{|c}{77.25}          & 54.59          \\
+DoLa                                      & 31.58          & 54.40          & 32.31             & \multicolumn{1}{|c|}{58.28}          & 61.44          & 77.97         & {39.90}     & {21.28}            & \multicolumn{1}{|c}{74.41}          & 49.05        \\
+SLED (ours)                                & \textbf{38.31} & \textbf{66.71} & \textbf{34.66}  & \multicolumn{1}{|c|}{\textbf{73.98}} & \textbf{62.55} & \textbf{84.70} & {\textbf{47.74}} & {\textbf{14.98}} & \multicolumn{1}{|c}{\textbf{77.38}} & \textbf{54.81} \\ 
\bottomrule
\end{tabular}}
\end{table}

\subsubsection{Evaluation on a Broad Range of LLM Benchmarks}\label{IOD-results}

\begin{table}[]
\centering
\caption{Using SLED with other LLM families also improves the factuality.}
\vspace{0.2cm}
\renewcommand{\arraystretch}{1.2} 
\resizebox{\textwidth}{!}{
\begin{tabular}{lllll|llllll}
\toprule
\multirow{2}{*}{Model} & \multirow{2}{*}{FACTOR} & \multicolumn{3}{c|}{TruthfulQA} & \multirow{2}{*}{Model} & \multirow{2}{*}{FACTOR} & \multicolumn{3}{c}{TruthfulQA} \\ \cline{3-5} \cline{8-10} 
                       &                         & \multicolumn{1}{c}{MC1}      & \multicolumn{1}{c}{MC2}      & \multicolumn{1}{c|}{MC3}      &                         &                         & \multicolumn{1}{c}{MC1}      & \multicolumn{1}{c}{MC2}      & \multicolumn{1}{c}{MC3}      \\ \midrule
LLaMA-3-8B             & 64.33                   & 33.78    & 63.00    & 32.59    & Mixtral-8$\times$7B           & 71.41                   & 35.13    & 49.98   & \textbf{34.17}    \\
+DoLa                  & 68.04                   & 33.29    & 63.35    & 32.16    & +DoLa                   & 58.28                   & 32.44    & 35.91    & 33.68    \\
+SLED (ours)           & \textbf{68.67}          & \textbf{35.13}    & \textbf{64.09}    & \textbf{32.50}  & +SLED (ours)             & \textbf{74.92}          & \textbf{35.86}    & \textbf{57.26}    & 32.96 \\ \midrule
LLaMA-3-8B-IT          & 59.49                   & 38.92    & 68.16    & 36.50    & Mixtral-8$\times$7B-IT        & 70.51                   & 37.94    & 62.51    & 35.25    \\
+DoLa                  & 61.06                   & 35.86    & 65.30   & 33.78    & +DoLa                   & 56.15                   & 32.19    & 39.17    & 33.76    \\
+SLED (ours)           & \textbf{67.17}          & \textbf{42.23}    & \textbf{69.03}    & \textbf{37.97} & +SLED (ours)             & \textbf{75.55}          & \textbf{41.73}    & \textbf{68.52}    & \textbf{37.70} \\ \midrule
LLaMA-3-70B            & 78.72                   & 35.62    & 65.66    & \textbf{34.18}    & Gemma-2B                  & 50.87                   & 23.38    & 37.16    & 17.42    \\
+DoLa                  & 77.56                   & 33.29    & 64.83    & 32.81    & +DoLa                   & 32.93                   & \textbf{26.07}    & 48.97    & 26.55    \\
+SLED (ours)           & \textbf{80.83}          & \textbf{37.58}    & \textbf{66.19}    & 34.11 & +SLED (ours)             & \textbf{57.05}          & 25.21    & \textbf{50.20}    & \textbf{26.94} \\ \midrule
LLaMA-3-70B-IT         & 73.95                   & 44.80    & 70.29    & 41.02    & Gemma-7B                 & 60.42                   & 31.58    & 47.63    & 22.75    \\
+DoLa                  & 71.51                   & 38.43    & 68.70    & 35.21    & +DoLa                   & 36.07                   & 25.21    & 43.14    & \textbf{26.13}    \\
+SLED (ours)           & \textbf{76.85}          & \textbf{48.35}    & \textbf{74.03}    & \textbf{43.16} & +SLED (ours)             & \textbf{65.56}          & \textbf{32.31}    & \textbf{49.88}    & 25.22 \\
\bottomrule
\end{tabular}
}\label{results_llama3}
\end{table}

\paragraph{Multiple-Choices Tasks} The objective of these tasks is to employ decoding methods that enable LLMs to assign higher probabilities to correct completions/answers over incorrect alternatives. We demonstrate the effectiveness of SLED for both Short-Answer Factuality on the TruthfulQA and Long-Paragraph Factuality on the FACTOR dataset. For both DoLa and our SLED, we contrast the results from the final layer against all preceding layers. We randomly sample approximately 5\% of the data for validation regarding parameter selection. The results, as shown in Table \ref{tab:mainresults_label}, indicate that SLED achieves superior outcomes in almost all metrics across six LLaMA-2 models. Notably, SLED achieves better performance under the MC1/MC3 metrics on TruthfulQA, which are more sensitive to fluctuations and pose a greater challenge. For long sentences in FACTOR, our method shows improvements over baselines by 5-13\%. These results not only underscore the benefits of our method for factuality but also demonstrate its robustness across different lengths of text.

\paragraph{Open-Ended Generation Tasks}
In open-ended settings, we prompt the model to generate answers for the same questions from TruthfulQA, following the settings outlined in \cite{lin-etal-2022-truthfulqa, chuang2024dola, li2022contrastive}. In Table \ref{tab:mainresults_label}, we compare the performance of six LLaMA-2 models using standard greedy decoding, (greedy) DoLa, and (greedy) SLED. All the generated answers are then evaluated by a fine-tuned GPT-3 model for both truthfulness and informativeness scores. Considering that a 100\% truthful score can be easily achieved by simply responding with 'I have no comment,' which would result in a 0\% informative score and thus is not very useful, we have introduced additional metrics—\%Truth $\times$ Info and the rejection ratio \%Reject —to demonstrate that SLED is a mutual-gains approach to achieve better both truthful and informative scores. We have improved the overall \%Truth x Info scores by 3-20\% across different models and reduced the rejection ratio by up to 95\%. These enhancements demonstrate that our method effectively avoids the 'rejection pitfall,' making it more helpful.

\paragraph{Adaptation to Chain-of-thought Reasoning Tasks} Although the StrategyQA and GSM8K tasks are also open-ended and require factual accuracy, the primary focus here is to evaluate how different decoding methods adapt to the Chain-of-Thought (COT) approach for handling complex reasoning tasks. We maintain a repetition penalty of 1, as we will discuss the repetition flaws associated with DoLa in Section \ref{exp-analysis}. StrategyQA demands multi-hop reasoning, and as shown in Table \ref{tab:mainresults_label}, our method boosts accuracy across six models, whereas DoLa generally worsens it without a repetition penalty. GSM8K, a benchmark for math word problems that require arithmetic reasoning, also shows consistent accuracy improvement with SLED in 7B, 13B and 70B models.

\subsubsection{Evaluation Across Diverse LLM Configurations}\label{sec:llama3}

As discussed above and shown in Table \ref{tab:mainresults_label}, our method, SLED, demonstrates strong generalization capabilities across the LLaMA-2 model family, proving robust from 7B to 70B model sizes. In Table \ref{results_llama3}, we further showcase SLED's impressive performance on the more recent LLaMA-3 family models, both at 8B and 70B sizes, in terms of long paragraph factuality and short answer factuality. Interestingly, SLED is also applicable to different pre-trained models, such as Gemma at both 2B and 7B sizes, and can even be adapted to the increasingly popular Mixture of Experts (MoE) architectures. These results confirm the exceptional adaptability of our method across various LLM configurations.

\subsubsection{Evaluation on Integrating SLED with Other LLM Factuality Decoding Methods}\label{boost-fd-results}

SLED exclusively focuses on contrasting differences between layers without altering other parts of the model. Thus, it remains compatible with other techniques that incorporate additional strategies or utilize auxiliary models. This compatibility allows SLED to be seamlessly integrated into existing methods, enhancing factuality further without the need for modifications to SLED. We integrate SLED with the following approaches: ITI, AD, CD and ICD. Table \ref{tab:combined_w_others} shows that SLED leads to accuracy improvements from 1\% to 12\% across four LLaMA-2 models.

\begin{table}[]
\renewcommand{\arraystretch}{0.96}
\centering
\caption{Comparison of decoding strategies on TruthfulQA datasets. SLED can also be seamlessly combined with other decoding strategies to improve performance further.} \label{tab:combined_w_others}
\vspace{0.2cm}
\small
\begin{tabularx}{1\textwidth}{l|X X X|X X X|X X|X X} 
\toprule
{Model}& \multicolumn{3}{c|}{{LLaMA-2-7B-base}} & \multicolumn{6}{c}{{LLaMA-2-7B-chat}} \\ 
\cmidrule(lr){0-1} \cmidrule(lr){2-4} \cmidrule(lr){5-11}
{Method}& \begin{tabular}[c]{@{}c@{}}AD\end{tabular} & \begin{tabular}[c]{@{}c@{}}AD\\+DoLa\end{tabular} & \begin{tabular}[c]{@{}c@{}}AD\\+SLED \end{tabular} & \begin{tabular}[c]{@{}c@{}}AD\end{tabular} & \begin{tabular}[c]{@{}c@{}}AD\\+DoLa\end{tabular} & \begin{tabular}[c]{@{}c@{}}AD\\+SLED\end{tabular} & \begin{tabular}[c]{@{}c@{}}ITI\end{tabular} & \begin{tabular}[c]{@{}c@{}}ITI\\+SLED\end{tabular} & \begin{tabular}[c]{@{}c@{}}ICD\end{tabular} & \begin{tabular}[c]{@{}c@{}}ICD\\+SLED\end{tabular} \\ 
\midrule
MC1 & 32.80 & 25.58 & \textbf{33.29} & 35.37 & {33.41} & \textbf{36.23} & 36.60 & \textbf{43.33} & 46.32 & \textbf{46.87} \\ 
MC2 & 59.59 & 39.06 & \textbf{62.55} & 58.14 & {50.31} & \textbf{63.15} & 65.62 & \textbf{65.75} & 69.08 & \textbf{72.09} \\ 
MC3 & 31.05 & 17.89 & \textbf{31.80} & 31.84 & {23.15} & \textbf{32.23} & 34.89 & \textbf{37.66} & 41.25 & \textbf{43.64} \\ 
\bottomrule
\vspace{0.01cm}
\end{tabularx}
\begin{tabularx}{\textwidth}{l|X X X|X X|X X X|X X} 
\toprule
{Model}& \multicolumn{5}{c|}{{LLaMA-2-13B-base}} & \multicolumn{5}{c}{{LLaMA-2-13B-chat}} \\ 
\cmidrule(lr){0-1}\cmidrule(lr){2-6} \cmidrule(lr){7-11}
{Method}& AD & \begin{tabular}[c]{@{}c@{}}AD\\+DoLa\end{tabular} & \begin{tabular}[c]{@{}c@{}}AD\\+SLED\end{tabular} & CD  & \begin{tabular}[c]{@{}c@{}}CD\\+SLED\end{tabular} & AD & \begin{tabular}[c]{@{}c@{}}AD\\+DoLa\end{tabular} & \begin{tabular}[c]{@{}c@{}}AD\\+SLED\end{tabular} & CD & \begin{tabular}[c]{@{}c@{}}CD\\+SLED\end{tabular} \\
\midrule
MC1 & 33.90 & 24.72 & \textbf{33.90} & 30.11 & \textbf{33.78} & \textbf{36.84} & 34.72 & {36.35} & 28.15 & \textbf{36.47} \\
MC2 & 62.93 & 37.74 & \textbf{63.69} & 50.31 & \textbf{63.22}  & 63.75 & 50.42 & \textbf{64.83} & 54.87 & \textbf{64.93} \\
MC3 & 31.61 & 17.66 & \textbf{31.38} &  28.18 &  \textbf{32.21} & 32.69 & 23.83 & \textbf{32.85} &  29.75 & \textbf{33.39} \\
\bottomrule
    \end{tabularx}
\end{table}

\begin{table}[t]
\centering
\caption{Accuracy of LLaMA 2 13B Base on StrategyQA with Varying Repetition Penalties}\label{Repetition_ratio}
\vspace{0.2cm}
\renewcommand{\arraystretch}{1.2}
\resizebox{\textwidth}{!}{
\begin{tabular}{ll|cccccccc}
\hline
{Metric} & {Method} & {1} & {1.02} & {1.04} & {1.06} & {1.08} & {1.1} & {1.2} & {2} \\ \hline
\multirow{2}{*}{Accuracy(\%)} & DoLa & 65.55 & 65.98 & 66.37 & 65.98 & 65.59 & 66.37 & 67.16 & 66.64 \\
                              & SLED (Ours) & 66.81 & \textbf{69.39} & 68.51 & 68.47 & 67.07 & 65.72 & 60.87 & 54.75 \\ \hline
\multirow{2}{*}{Repetition-4(\%)} & DoLa & 7.63 & 7.19 & 6.45 & 5.98 & 5.50 & 5.10 & 3.73 & 2.05 \\
                              & SLED (Ours) & \textbf{3.73} & \textbf{2.45} & \textbf{1.89} & \textbf{1.36} & \textbf{1.05} & \textbf{0.69} & \textbf{0.20} & \textbf{0.10} \\ \hline
\multirow{2}{*}{Repetition-Sen(\%)} & DoLa & 2.16 & 2.04 & 1.66 & 1.37 & 1.12 & 0.89 & 0.23 & 0.03 \\
                              & SLED (Ours) & \textbf{0.88} & \textbf{0.39} & \textbf{0.10} & \textbf{0.02} & \textbf{0.03} & \textbf{0} & \textbf{0} & \textbf{0} \\ \hline
\end{tabular}}
\end{table}

\subsubsection{Ablation Studies and Analysis} \label{exp-analysis}

\paragraph{Mitigating Repetition Issues}
Table \ref{Repetition_ratio} demonstrates that our method, SLED, effectively addresses a significant issue in DoLa: repetitive content in open-ended generation tasks. Our approach outperforms DoLa without the need for excessive repetition penalty. While a slight increase in the repetition penalty further enhances the performance of our method, excessive penalties, such as 1.1, tend to degrade it. This suggests that SLED does not inherently require heavy adjustments for repetition issues. In contrast, DoLa's performance improves with higher penalties (e.g., 1.1, 1.2, 2), indicating a more critical need for addressing repetitive content. We also employ two intuitive metrics, Repetition-4 and Repetition-Sen, to gauge the severity of repetition issues, following prior research \cite{xu2022learning}. Regardless of the repetition penalty imposed, our method consistently exhibits lower repetition rates. Table~\ref{tab:repetition} includes some examples of generated text to illustrate this further.

\paragraph{Layer Selection}

\begin{figure}[t]
    \centering
        \caption{Evaluating using different premature layers for SLED and DoLa on a 10\% subset of the GSM8K dataset.}
    \begin{subfigure}[b]{0.46\textwidth}
        \includegraphics[width=1\textwidth]{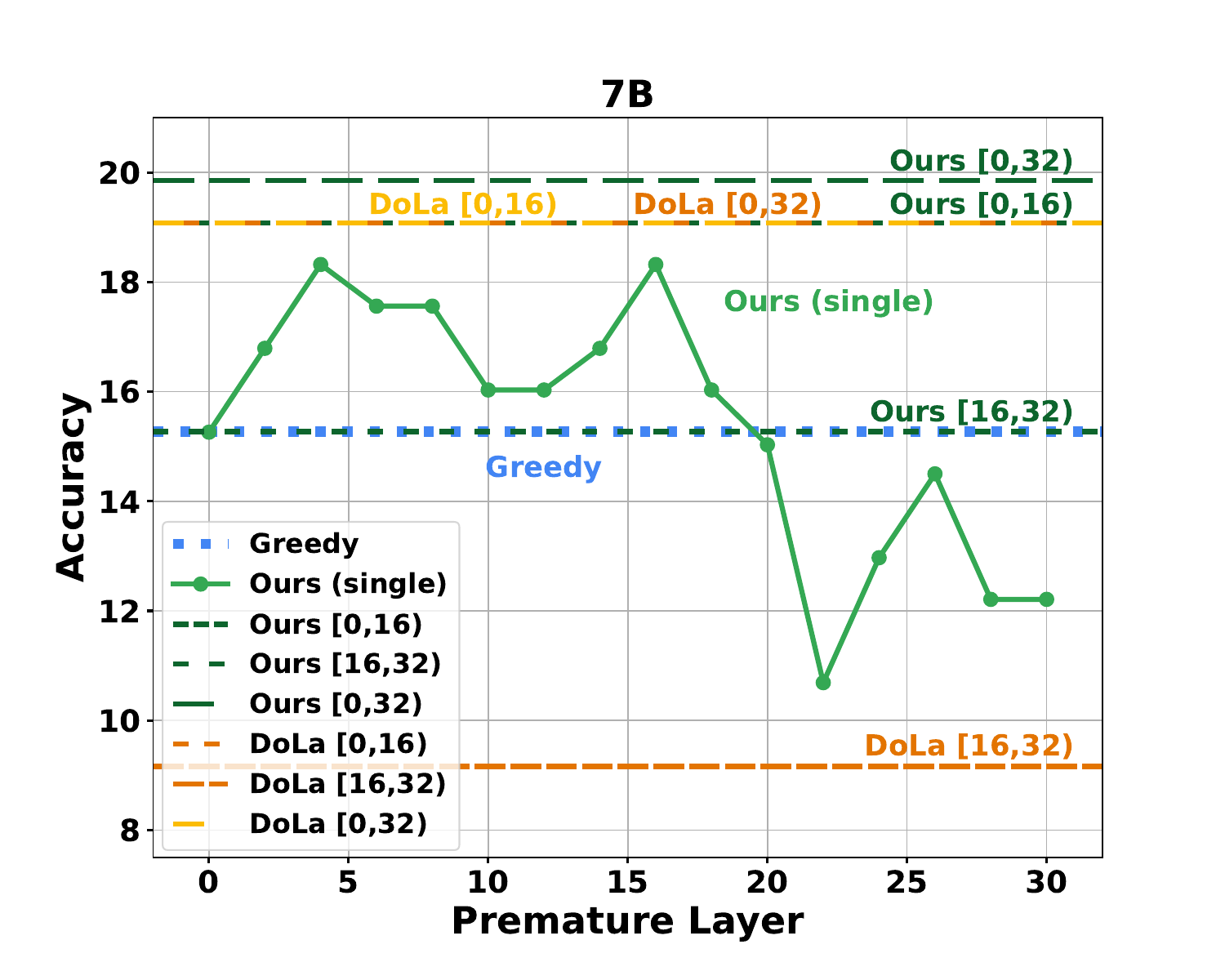}
        \label{fig:sub1}
    \end{subfigure}
    \begin{subfigure}[b]{0.46\textwidth}
        \includegraphics[width=1\textwidth]{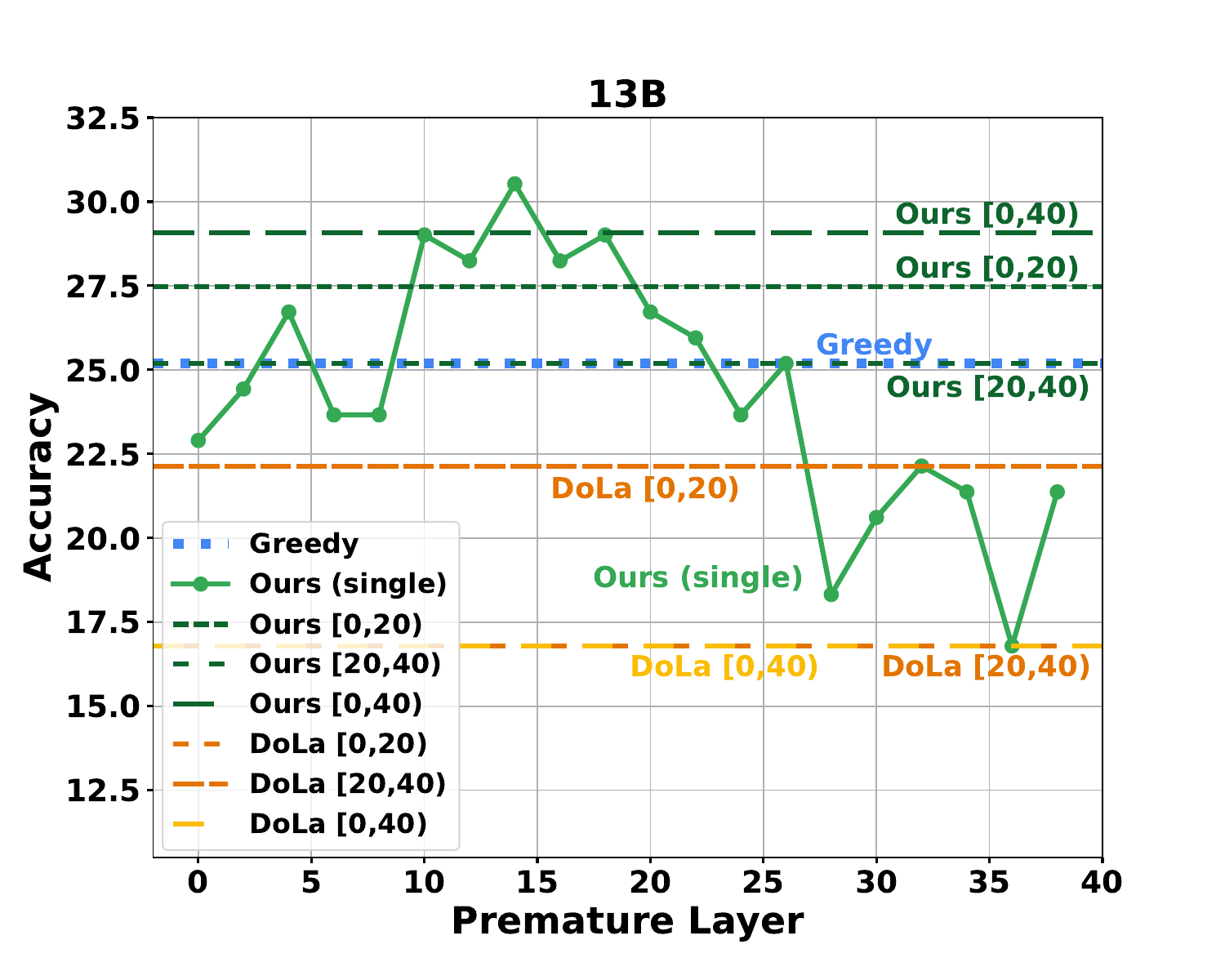}
        \label{fig:sub2}
    \end{subfigure}
    \label{fig:pre_layers}
\end{figure}

How to choose a good candidate set is still a paradoxically difficult task when applying DoLa. Our method does not exhibit this issue. Instead of selecting a single premature layer from the candidate set like DoLa, SLED contrasts the final layer with all layers in the candidate set and then ensembles all the results. Figure \ref{fig:pre_layers} shows that setting a larger candidate set, such as all the 32 layers for LLaMA-2-7B-Base, yields better performance than focusing solely on either the first $[0,16)$ or second half $[16,32)$. Contrasting all layers for SLED is better than using only the first half [0, 16) or the second half [16, 32). Hence, there are no improvements for SLED from strategic layer subset selection. This implies that our layer-wise contrast approach captures more useful information in a more scientific manner. Furthermore, our tests confirm the robustness of our method even when the candidate set is minimal, such as a single layer, consistently demonstrating strong performance. Our settings mirror those of DoLa.

\paragraph{Parameter Analysis}

\begin{figure}[t]
    \centering
    \caption{WE explore the impact of evolution scale and rate based on the factual accuracy of a subset of the FACTOR dataset. (G: Greedy, D: DoLa)}
    \vspace{5mm}
    \hspace{-0.6cm}
    \begin{subfigure}[b]{0.28\textwidth}
        \includegraphics[width=0.95\textwidth]{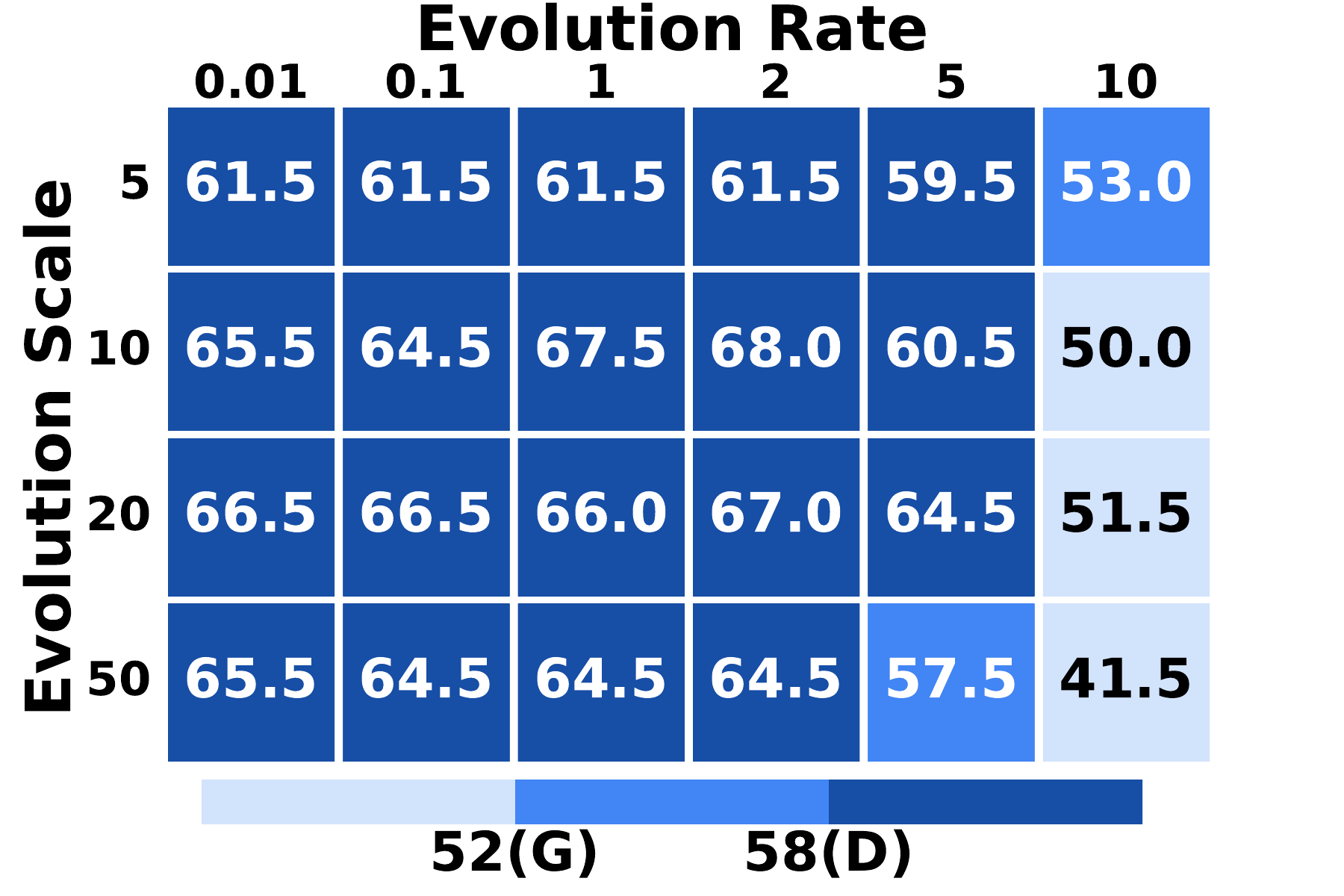}
        \caption{LLaMA 2 7B Base}
        \label{fig:hp-factor-sub1}
    \end{subfigure}
    \hspace{-0.6cm}
    \begin{subfigure}[b]{0.28\textwidth}
        \includegraphics[width=0.95\textwidth]{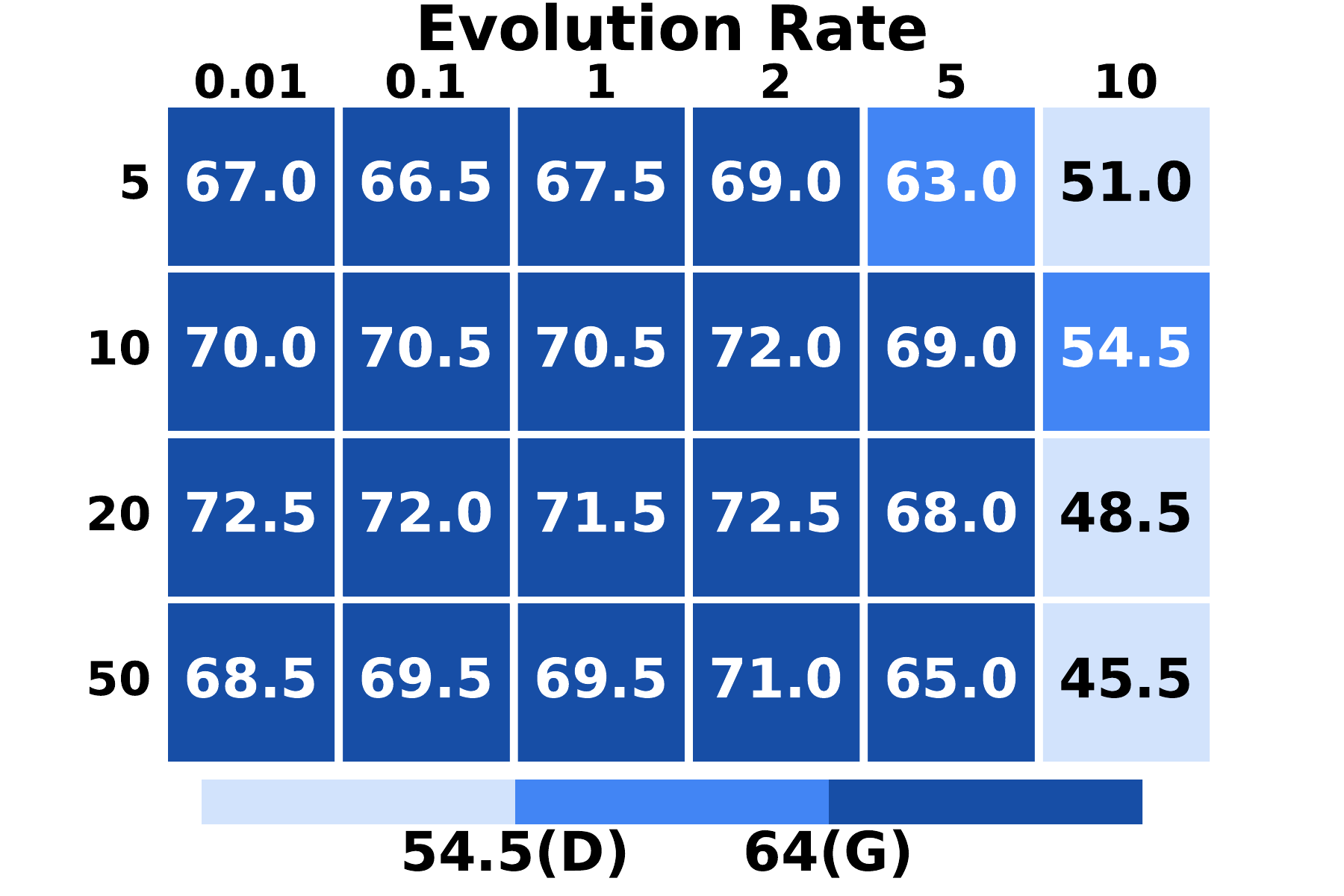}
        \caption{LLaMA 2 7B Chat}
        \label{fig:hp-factor-sub2}
    \end{subfigure}
     \hspace{-0.6cm}
    \begin{subfigure}[b]{0.28\textwidth}
        \includegraphics[width=0.95\textwidth]{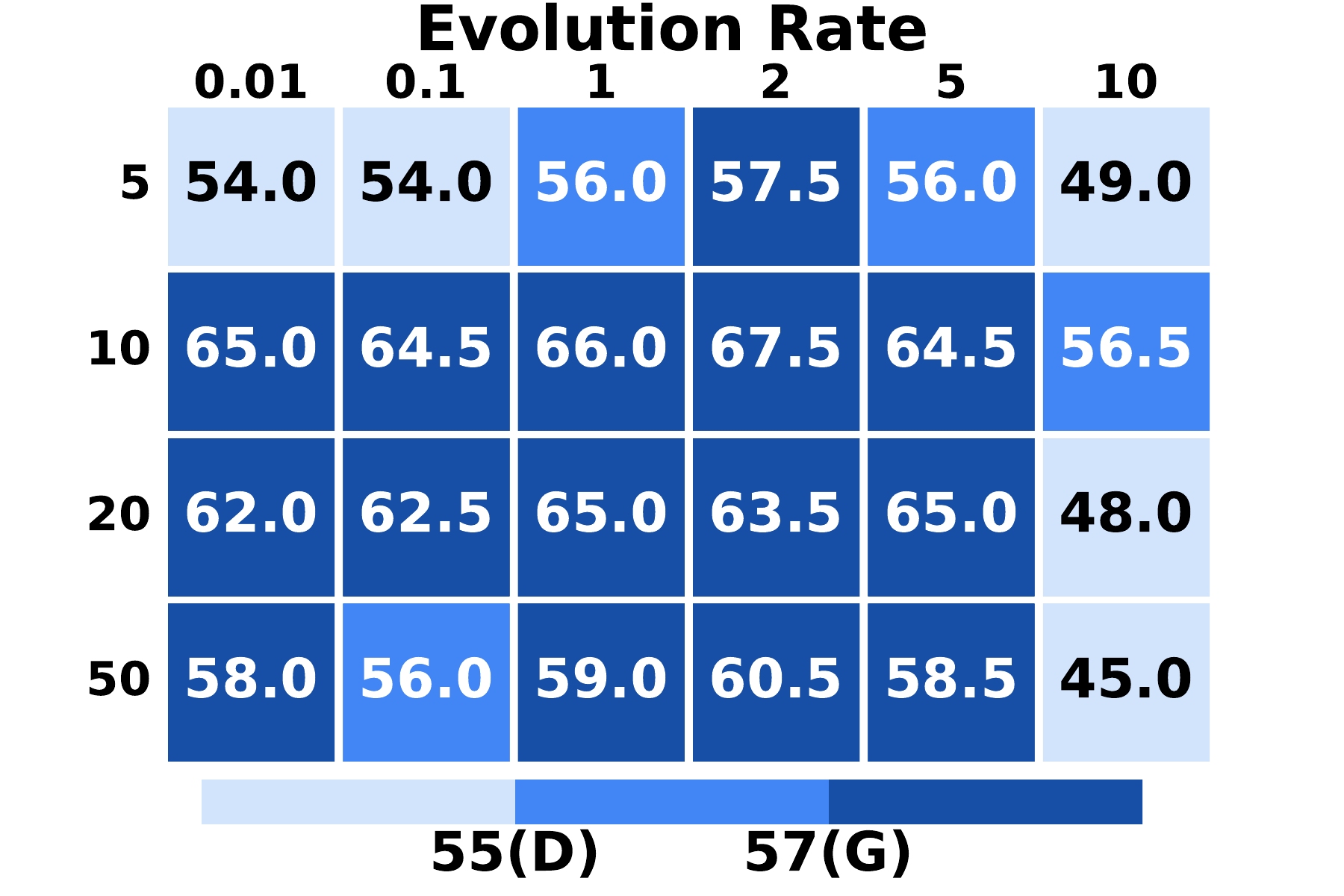}
        \caption{LLaMA 2 13B Base}
        \label{fig:hp-factor-sub3}
    \end{subfigure}
     \hspace{-0.6cm}
    \begin{subfigure}[b]{0.28\textwidth}
        \includegraphics[width=0.95\textwidth]{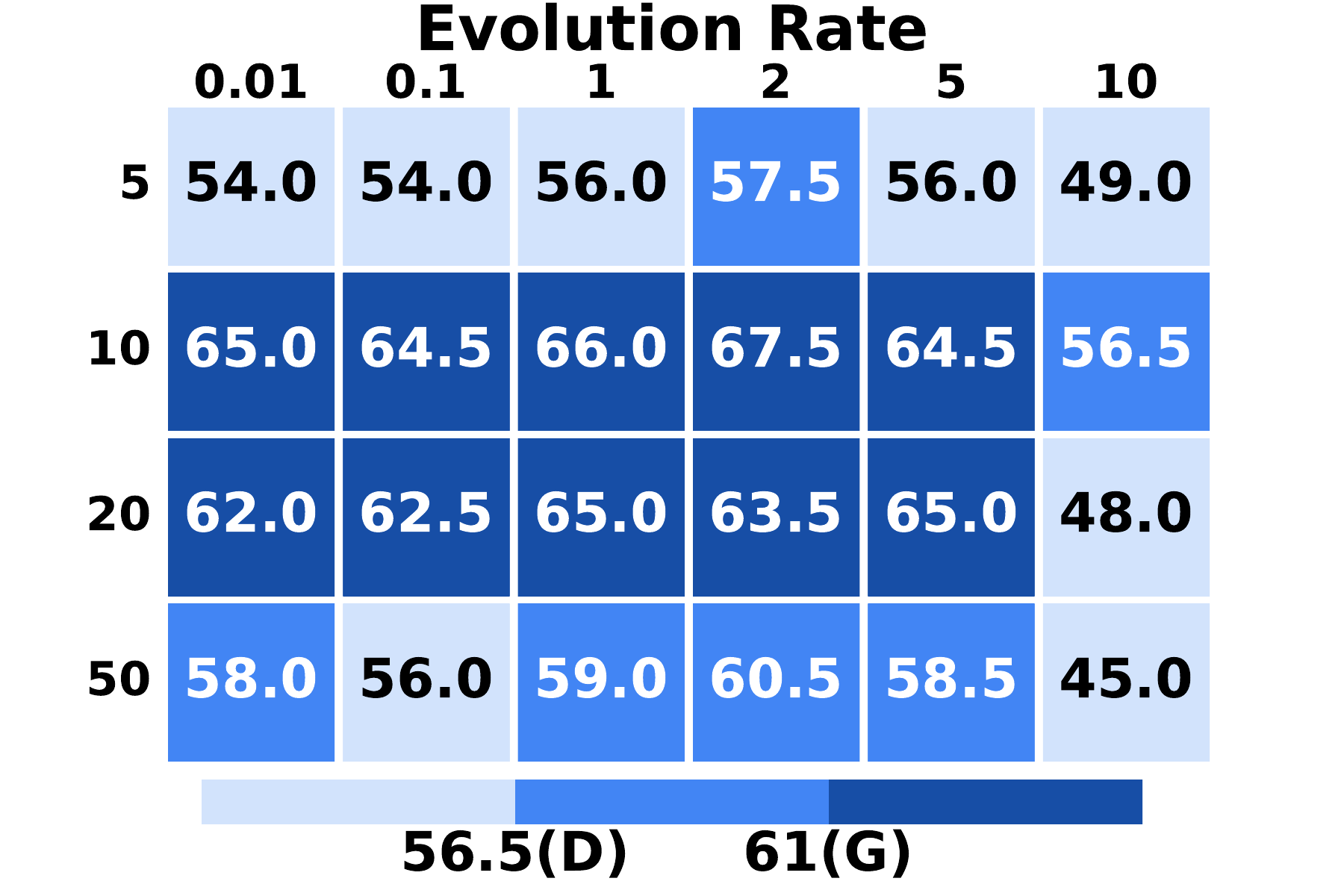}
        \caption{LLaMA 2 13B Chat}
        \label{fig:hp-factor-sub4}
    \end{subfigure}
     \hspace{-0.8cm}
    \label{fig:hp-factor}
\end{figure}

We next investigate the impact of parameters — evolution rate $\alpha$ and evolution scale $k$ — on the performance of SLED using a subset of the FACTOR dataset. We test evolution rates from $\{0.01, 0.1, 1, 2,5, 10\}$ and evolution scale values from $\{5, 10, 20, 50\}$. Without extreme evolution rates (e.g., 10), our method performs well, confirming its robustness. As analyzed in our methodology and Eq.~\ref{optimized_lr_delta}, the evolution rate balances the logit distribution ($\mathcal{P}_\mathcal{N})$ with the latent knowledge distribution ($\mathcal{P}_{\mathit{latent}}$). A lower evolution rate works better for larger models (13B) and chat models as their logits already better represent real-world distributions.

\paragraph{Latency}

Our method, SLED, does not incur significant latency overhead. The latencies presented in Table~\ref{Tab:latency} demonstrate that our method, SLED, just increases the decoding time of DoLa by factors ranging from 0.1\% to 10\%. Notably, even with an atypical setting such as evolution scale $ = 100$, which is seldom used, the increase remains around 10\%. The latency for DoLa and SLED is much higher compared to the vanilla greedy decoding because we set all early layers as candidate layers set for both DoLa and SLED for a fair comparison.

\begin{table}[htbp]
\centering
\caption{Latency (ms/token) comparison across different configurations. (ES: evolution scale)}\label{Tab:latency}
\vspace{0.2cm}
\renewcommand{\arraystretch}{1.1}
\resizebox{\textwidth}{!}{
\begin{tabular}{lcccccc}
\toprule
\textbf{Model} & \textbf{Greedy} & \textbf{DoLa} & \textbf{SLED (ES=5)} & \textbf{SLED (ES=20)} & \textbf{SLED (ES=50)} & \textbf{SLED (ES=100)} \\
\midrule
LLaMA-2-7B  & 23.64 & 29.93 & 30.41 & 31.15 & 32.70 & 34.63 \\
LLaMA-2-13B & 30.41 & 39.57 & 39.61 & 41.14 & 43.30 & 45.09 \\
LLaMA-2-70B & 82.63 & 136.42 & 138.33 & 140.24 & 143.12 & 148.85 \\
\bottomrule
\end{tabular}
}
\end{table}

\subsection{Summary}


We introduced Self Logits Evolution Decoding (SLED), a novel inference-time method to improve factual accuracy in large language models (LLMs) without the need for external knowledge retrieval (e.g., RAG) or additional fine-tuning (e.g., SFT). The core intuition is to refine the original output logits by leveraging the latent knowledge already embedded in the LLM's internal activations. By engineering the output probability distributions directly from internal representations, SLED provides a powerful demonstration of how Probability Engineering can significantly enhance factuality and robustness in high-quality text generation. Our extensive experiments across multiple benchmarks confirm that SLED achieves state-of-the-art performance, surpassing both vanilla decoding strategies and existing factuality-enhancing methods.

These insights underscore Probability Engineering’s effectiveness for enhancing generation tasks, setting the stage for broader applicability. Indeed, our other projects on large language models (or large multimodal models) \cite{zhang2024towards,kuo2025proactive,zhang2024min,kuo2025hcothijackingchainofthoughtsafety,lin2024speechprune,joren2024sufficient,wang2024coreinfer,zhang2022join,kuo2023dacbertleveragingdependencyagreement} further illustrate the critical role Probability Engineering can play, highlighting its potential in reliably leveraging complex, large-scale vision-language datasets to address real-world demands.

\section{Text-to-Image Generation with Diffusion Models}

The field of text-to-image generation has made remarkable progress, especially with the rise of diffusion models \cite{ho2020denoising,rombach2022high,ramesh2022dalle2,saharia2022photorealistic,yang2022diffusion,zhou2023shifted,ruiz2023dreambooth}. These models not only have demonstrated prowess in generating high-fidelity images, but have also showcased versatile applications, such as image inpainting \cite{lugmayr2022repaint,esser2021imagebart,jing2022subspace} , denoising \cite{xie2023diffusion,Yang2023RealWorldDV}, video
generation \cite{han2022card,blattmann2023videoldm}, and style transfer \cite{wang2023stylediffusion,zhang2022inversion}. Along with these advancements, the ability to generate images containing legible text, another realm of significance, is becoming increasingly worthy of attention. Specifically, this need is evident in everyday scenarios, where images with text are commonplace, from advertisements to road signs, posters, and book covers. Crafting these text-rich images manually demands skilled expertise and a significant time commitment. However, a significant shortcoming lies in the current state-of-the-art diffusion-based generative models; they often render text portions that are virtually unreadable, akin to gibberish, undermining the aesthetic and functional value of the generated images. Therefore, if generative AI, powered by diffusion models, can produce such images, it could revolutionize design workflows, inspire creativity, and alleviate designers' workload.
\begin{figure*}[t!]
    \centering
    \includegraphics[width=0.95\linewidth]{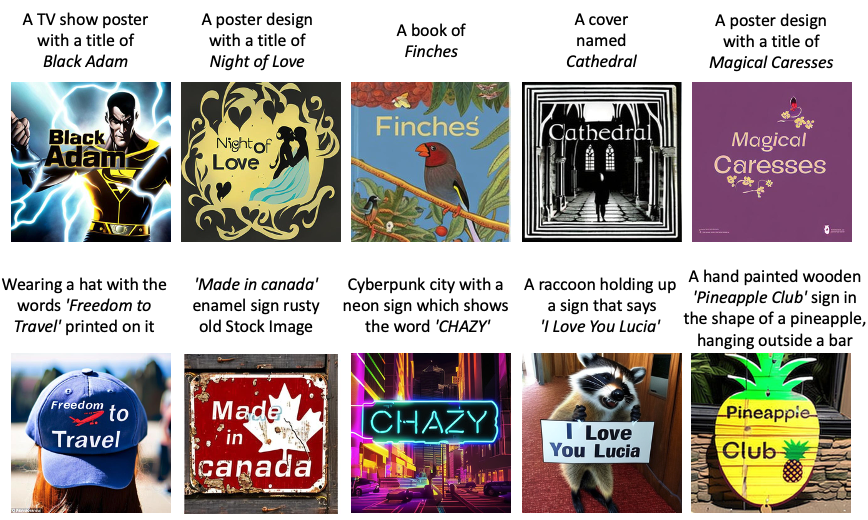}
    \vspace{5mm}
    \caption{Generated examples from our ARTIST. }
    \label{fig:examples}
\end{figure*}

To tackle the challenges posed by the quality of text generated by diffusion models, two primary pathways have been explored. Firstly, the traditional approach involves leveraging image-editing tools to superimpose text onto images directly. However, this frequently introduces unnatural artifacts, especially when dealing with intricate textures or varying lighting conditions in the background of the image. In contrast, recent research efforts aim to refine the diffusion models themselves for improved text quality. For example, recent innovations such as Imagen \cite{saharia2022photorealistic}, eDiff-I \cite{balaji2022ediffi}, and DeepFloyd \cite{DeepFloyd} discovered that the use of T5 series text encoders \cite{raffel2020exploring} led to better results compared to using the CLIP text encoder \cite{radford2021learning}. Similarly, Liu et al.\cite{liu2022character} integrated character-aware text encoders to enhance the quality of text rendering. However, these advancements primarily revolve around optimizing text encoders and do not necessarily grant more control over the holistic generation process. On a parallel track, GlyphDraw \cite{ma2023glyphdraw} has enhanced model controllability by conditionally focusing on the positioning and architecture of Chinese characters. Still, its utility remains limited, as it cannot cater to scenarios that require multiple text bounding boxes, making it less suitable for prevalent text-image formats like posters and book covers. TextDiffuser \cite{chen2023textdiffuser}  represents a recent advancement in the realm of enhancing the quality of text in generated images. While it undeniably marks a significant step forward, it is not without limitations. One of the prominent challenges is the model's dependence on manual efforts when it comes to recognizing key terms from prompts, making the process less efficient than desired. Additionally, even though TextDiffuser exhibits improved capabilities, the accuracy rate of its Optical Character Recognition (OCR) evaluation on generated text still leaves room for optimization.

In this work, our primary objective is to develop a more efficient system for text-to-image generation. This system would eliminate the need for manual post-production adjustments, ensuring superior text quality within generated images. Upon meticulous investigation, we have identified the primary challenges we aim to address. Firstly, automation of accurately discerning which words from a provided text prompt should be incorporated into the image. In platforms such as TextDiffuser, this identification requires human involvement, typically by highlighting specific terms with quotation marks, diminishing the platform's efficiency and automation. Second, there is the challenge of adeptly generating images that seamlessly integrate top-quality text, ensuring adherence to a predetermined layout, and emphasizing in the generated images.

\begin{figure*}[t]
    \centering
    \includegraphics[width=0.95\linewidth]{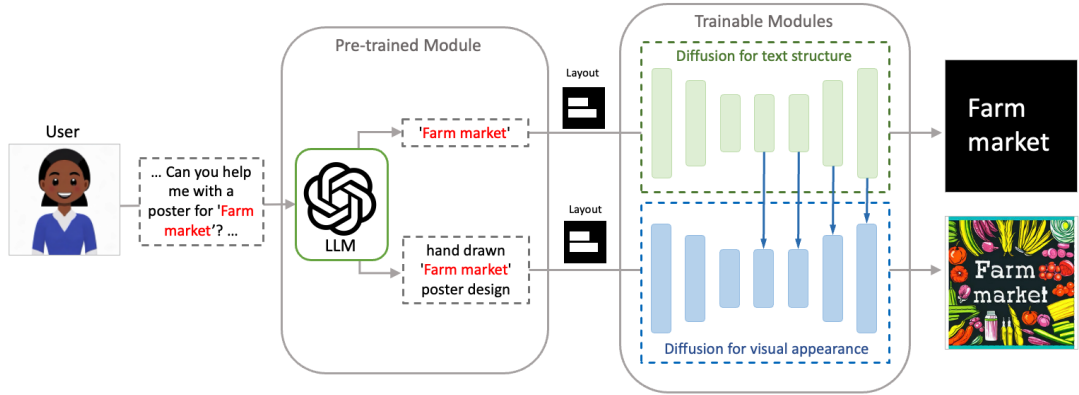}
    \vspace{5mm}
    \caption{Illustration of the proposed ARTIST framework. }
    \label{fig:framework}
\end{figure*}

To address the aforementioned challenges, we turned to the latest advancements in natural language processing and large language models (LLMs)~\cite{OpenAI2023GPT4TR,touvron2023llama,wang2024coreinfer,joren2024sufficient,yao2024federated,zhang2022join,kuo2023dacbertleveragingdependencyagreement,zhang2023reaugkd}. LLMs have showcased outstanding expertise in understanding and processing intricate linguistic patterns, making them perfectly suited for our needs. Inspired by their prowess, we formulated a strategy that employs large language models to surmount the initial hurdle of pinpointing the keywords. To tackle the subsequent challenge, we introduce a novel two-stage approach named ARTIST, detailed in Figure \ref{fig:framework}. Specifically, ARTIST's dual stages are dedicated to mastering text structure and refining visual aesthetics in that order. The moniker ``ARTIST" encapsulates the essence of \textit{the Ability of Rendering Text can be Improved by diSentanglemenT}. By merging our proposed methodology with LLMs' capabilities, we have achieved a marked improvement in the quality of text embedded within images.

\subsection{Background}

\paragraph{Text-to-image Generation}
Recent advances in text-to-image generation models can be categorized into two primary categories: autoregressive frameworks \cite{ramesh2021zero,yu2022scaling} and diffusion-based models \cite{ramesh2022dalle2,saharia2022photorealistic}, have been substantial. The latter, particularly diffusion-centric models, have gained significant momentum in recent times. The crux of these models lies in the conversion of textual prompts into latent representations, subsequently relying on diffusion mechanisms to formulate images. Several renowned models, such as Stable Diffusion \cite{rombach2022high}, DALL-E 2 \cite{ramesh2022dalle2} and Imagen \cite{saharia2022photorealistic} have set benchmarks in this domain. The challenge of controlled generation remains at the forefront. Contemporary image generation models, with an emphasis on diffusion, heavily prioritize text-based guidance, making it a promising vector for refining their controllability. The inherent complexities of gaining exact control through textual means are evident. For example, ControlNet \cite{zhang2023adding} provides an architectural blueprint that adapts pretrained diffusion models to accommodate a variety of input conditions. Although offering a higher level of flexibility, obtaining such condition signals often necessitates manual intervention and might be restrictive for broader conceptual applications. Methods such as GLIGEN \cite{li2023gligen} advocate for open-set image generation by employing grounding tokens to define the spatial parameters of objects. We agree with this approach, integrating elements of the grounding token methodology. 

\paragraph{Text Rendering \& OCR}

Despite the rapid development of diffusion-based Text-to-image Generation, existing methodologies still struggle with generating precise and consistent textual renderings. Various approaches, like Imagen \cite{saharia2022photorealistic}, eDiff-I \cite{balaji2022ediffi}, and DeepFolyd \cite{DeepFloyd}, have leveraged the prowess of expansive language models (notably large T5 \cite{raffel2020exploring}) to bolster their textual accuracy. The study in \cite{liu2022character} highlights a limitation in which traditional text encoders overlook token length, prompting them to propose a character-sensitive alternative. Simultaneously, GlyphDraw \cite{ma2023glyphdraw} focuses on producing superior images integrated with Chinese texts, guided by textual positioning and glyph imagery. GlyphControl \cite{yang2024glyphcontrol} further enhances this approach by adjusting the text alignment according to its location, implicitly incorporating elements like font size and text box positioning. A recent study, AnyText \cite{tuo2023anytext}, utilizes a diffusion pipeline with an auxiliary latent module and a text embedding module to improve the text generation, editing, and integration with the image background. The techniques of Textdiffuser \cite{chen2023textdiffuser} harnesses the Transformer model to discern keyword layouts, promoting multiline text generation. Then, it further employs character-based segmentation masks as a prior, offering flexibility in control to cater to user specifications. However, it still depends on manual efforts when it comes to recognizing key terms from prompts. Although Textdiffuser-2 \cite{chen2023textdiffuser-2} employs LLMs to enhance the interpretation of prompts, the improvements in the quality of generated images with text remain modest. This indicates that although the system has improved in understanding user inputs, converting these advancements into better visual results still demands additional optimization. Hence, we believe it is necessary to develop a more effective framework that captures the intricacies of text representation and seamlessly integrates textual structures into images, aiming for more coherent outputs.

 Optical Character Recognition (OCR) is a long-established academic endeavor \cite{white1983image,cash1987optical}. In the last decade, this field has seen remarkable progress, impacting a range of applications including recognition of car license plates \cite{Iraqi_car_license}, autonomous vehicle navigation \cite{Detecting_symbols6957755,wu2023ocr}, and its incorporation into foundational models such as GPT \cite{huang2023language,shen2023hugginggpt}. In this work, OCR serves as a pivotal evaluation metric used to critically assess the quality of the generated text and provide a comprehensive understanding of our model's performance in realistic scenarios.

\subsection{ARTIST: Improving the Generation of Text-rich Images with Disentangled
Diffusion Models and Large Language Models}
As revealed in previous works \cite{ramesh2022dalle2,saharia2022imagen,chen2023textdiffuser}, generating an image with text rendered on it is still challenging. We suspect that this happens because of two major reasons: 

\begin{itemize}
    \item It is challenging to simultaneously learn visual appearance and text structure with a single model;
    \item Existing datasets are not able to cover all the words and their possible combinations, making it hard to learn text structure from these limited noisy data.
\end{itemize}

In this work, we propose to mitigate the aforementioned challenges by utilizing separate modules to learn text structure and visual appearance. Furthermore, these two modules are trained separately, making it possible to learn text structure with synthetic data constructed by ourselves, which also tackles the problem of limited data. To ensure disentanglement and prevent information leakage, our text module and visual module take different prompts as inputs. However, it can be inefficient and user-hostile if these prompts have to be manually designed by the user themselves at inference. Fortunately, because of the recent success of large-language models (LLMs)~\cite{brown2020GPT3,touvron2023llama2, zhang2024towards,zhang2024sled,zhang2024min,zhang2024mllm}, we propose to utilize pre-trained LLMs to infer user's intention, provide accurate prompts for both modules. With the help of LLM, user's input can be either precise or vague, leading to a better interactive experience. 

Our proposed framework is termed ARTIST, because it illustrates that \textit{the Ability of Rendering Text can be Improved by diSentanglemenT}.
Our proposed framework is illustrated in Figure \ref{fig:framework}, with details discussed below. A large-language model (LLM) is utilized to analyze the user's intention. Two diffusion models will be trained to learn text structure and other visual appearance respectively. Given a user input, the LLM will output keywords, layout and text prompts, which will be fed into our trainable modules to generate target images. As we show in Section \ref{artist_exp}, our proposed framework outperforms the previous state-of-the-art (SOTA) in terms of image fidelity, image-prompt alignment, and accuracy of generated texts.

\subsubsection{LLM-based Prompt Understanding} \label{sec:LLM_PU}
TextDiffuser~\cite{chen2023textdiffuser} proposes to train a transformer model to extract texts that are expected to be shown on generated images. However, there appears to be a limitation associated with this method: their model is only capable of detecting words enclosed by quotation marks. This happens because most keywords from the training dataset are enclosed by quotation marks inside the captions. A model trained on these samples will lack generalization ability or even be overfitting. Moreover, their model even fails to generate the desired image according to the prompt ``a poster of Batman with the word Batman on it", as it does not understand that the word ``Batman" should be presented on the image.

On the contrary, we propose to provide pre-trained LLM with an open-domain prompt and rely on its capability to autonomously identify the essentials. Because of its vast open-domain knowledge, LLM is able to better understand user intents and thus can generalize to complicated scenarios. Given vague prompts, LLM is able to discern and propose which words or text elements to incorporate, resulting in more coherent and aesthetically appealing suggestions.

\subsubsection{Learning Text Structure}
\label{sec: learntext}
A diffusion model, denoted as a text module, is introduced to learn text structures. Specifically, this text module is trained to take bounding boxes as input and generates a black-white image with only text on it.
Because we do not require it to learn any visual effects, the prompt for this module is the word to be rendered and the dataset can be simply constructed using standard rendering libraries
Specifically, we construct two large-scale datasets to train this module, as described below.
\begin{figure}[t]
    \centering
    \includegraphics[width=0.6\textwidth]{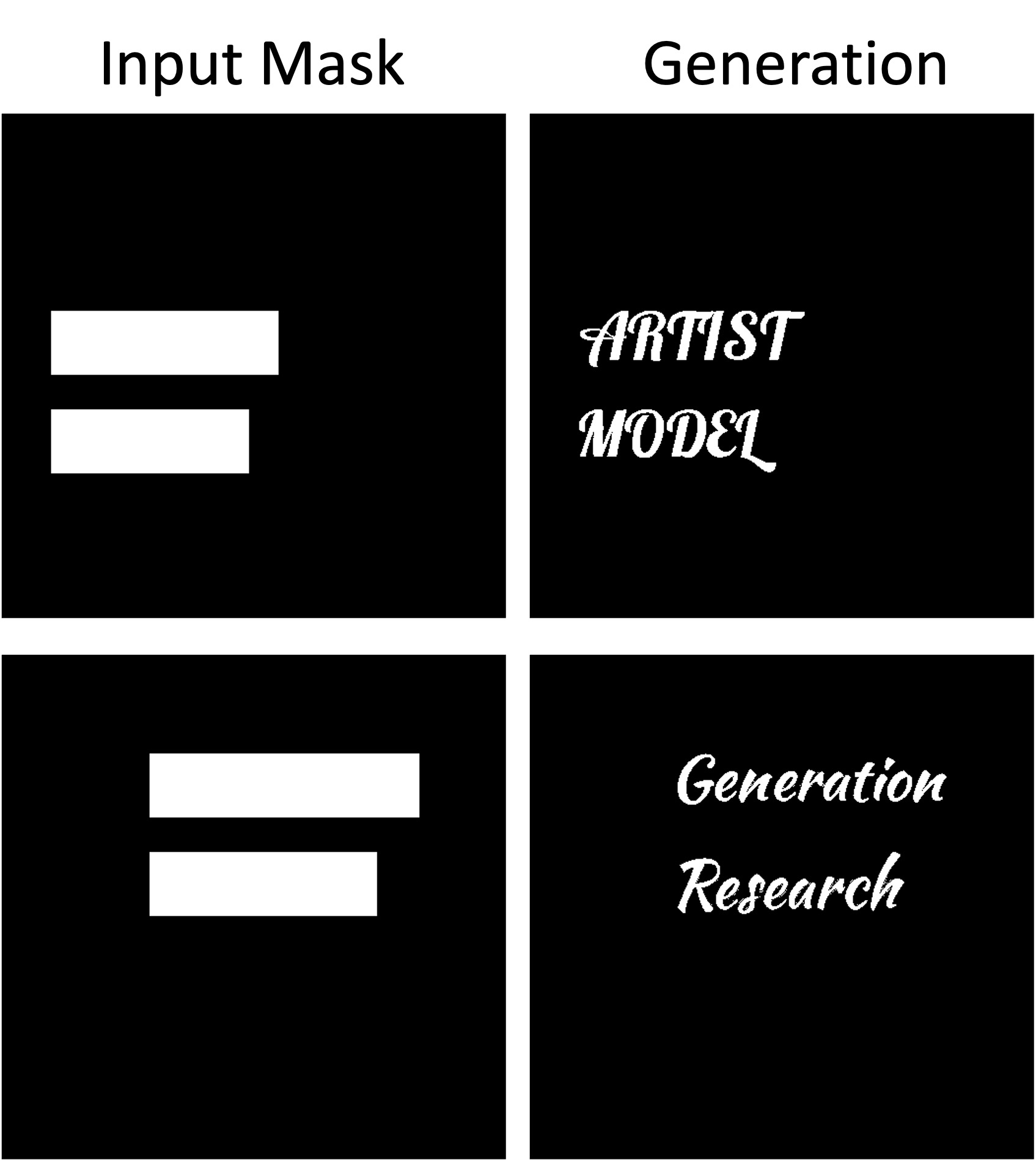}
    \vspace{5mm}
     \caption{Generated examples from our text module, along with input masks.}\label{fig:text_module}
\end{figure}
\paragraph {Word-level dataset:} Our first dataset consists of around 10 millions of black-white images with only single word on them. Specifically, to construct each data sample, we first randomly select a word from the vocabulary of CLIP text encoder, then render this word with random font and size on a black image. Meanwhile, we can also obtain the ground-truth bounding box for each rendered word. This dataset will be further augmented during training by randomly moving both the word and bounding box to a new position, resulting in more effective samples;\vspace{-2mm}
  \paragraph {Sentence-level dataset:} Although our word-level dataset contains massive structure information of English words, it contains no layout information about how should these words be placed and combined on the target image. To this end, we construct our second dataset which contains 50 millions of black-white images by utilizing MARIO-10M dataset~\cite{chen2023textdiffuser}. Specifically, we use the ground-truth text and layout information from MARIO-10M samples, and render the same text with randomly selected fonts on black images following the same layout.

After constructing both datasets, we train the text module in two stages, inside the latent space of a VAE~\cite{kingma2013auto} following Rombach et al.~\cite{rombach2022high}. In the first stage, the diffusion model is trained to take a single bounding box and a target word as inputs, then generate a black-white image with the word rendered on it. Then, this module will be fine-tuned on our sentence-level dataset as the second stage so that it can take multiple bounding boxes and multiple s as inputs. 
Our s are denoted as \( \mathcal{P} = \{p_1, p_2, \ldots, p_n\} \), composed of the words to be generated in the image. Our input mask is an image containing the corresponding bounding boxes \( \{m_1, m_2, \ldots m_n\} \) that indicate the position of each word. 
During training, the original text-only image and input mask image are first encoded into latent space features $\mathbf{z}$ and $\mathbf{m}$. Then we sample a time step $t \sim$ Uniform $\left(0, T\right)$ and a Gaussian noise $\boldsymbol{\epsilon}$ to corrupt the original feature, yielding $\mathbf{z}_t =\sqrt{\bar{\alpha}_t} \mathbf{z}+\sqrt{1-\bar{\alpha}_t} \boldsymbol{\epsilon}$ where $\overline{\alpha}_T$ is the coefficient of the diffusion process introduced in \cite{ho2020denoising}. $\mathbf{z}_t$ and $\mathbf{m}$ are concatenated in the feature channel as input for diffusion model, which will be trained with the diffusion loss between the sampled $\boldsymbol{\epsilon}$ and the predicated noise $\boldsymbol{\epsilon}_\theta$:
\begin{align} 
    \mathcal{L}_{\text{text}}= \mathbb{E}\left[\left\|\boldsymbol{\epsilon}-\boldsymbol{\epsilon}_\theta\left(\mathbf{z}_t, \mathbf{m}, \mathcal{P}, t \right)\right\|_2^2\right].
    \vspace{-2mm}
\end{align}
Some generated examples are shown in Figure \ref{fig:text_module}, which illustrates that our text module can successfully generate a target image with desired texts on it in different styles. In our implementation, MARIO-10M dataset~\cite{chen2023textdiffuser} is used to train this module following previous work for a fair comparison. $\overline{\mathcal{P}}, \mathcal{P}, \mathbf{m}$ have already been prepared in MARIO-10M. At inference time, LLM will be utilized to infer $\overline{\mathcal{P}}, \mathcal{P}, \mathbf{m}$ as mentioned in Section \ref{sec:LLM_PU}. 

\subsubsection{Learning Visual Appearance}
After training the text module, we would like to utilize its learned knowledge to generate high-fidelity images containing text. To this end, we propose to inject intermediate features from our text module into our visual module, which is also a diffusion model.
For each intermediate feature from the mid-block and up-block layers of text module, we propose to use a trainable convolutional layer
to project the feature and add it element-wisely onto the corresponding intermediate output feature of the visual module. We have also tested different architectures, the comparison will be provided in later experiment section. 

During the training of the visual module, the text module will be frozen, and only the newly introduced layers and visual module will be fine-tuned.
Differently from the text module described in Section \ref{sec: learntext}, whose training text only contains the target words to be rendered, the training data for our visual module has to contain visual descriptions of the image so that the model can successfully learn to generate visual contents based on the user's input. 

Let $\overline{\mathcal{P}}$ be the prompt which contains visual descriptions of the image, $\mathcal{P}$ be the s as defined in Section \ref{sec: learntext}.
We feed $\mathcal{P}$ and the input mask $\mathbf{m}$ as mentioned in Section \ref{sec: learntext} into the pre-trained text module. 
The intermediate features from the text module denoted as $\{f_i(\mathbf{z}_t, \mathbf{m}, \mathcal{P}, t)\}_{i=1}^k$ will be injected into the visual module, where $k$ stands for the number of intermediate features. The visual module will be trained with diffusion loss:
\begin{align}
\vspace{-4mm}
    \mathcal{L}_{\text{visual}} = \mathbb{E}\left[ \left\| \epsilon - \epsilon_\theta\left(x_t, \{f_i(z_t, m, \mathcal{P}, t)\}_{i=1}^k, m, \overline{\mathcal{P}}, t\right)\right\|_2^2 \right] \nonumber \vspace{-6mm}
\end{align}
where $\mathbf{x}_t=\sqrt{\bar{\alpha}_t} \mathbf{x}+\sqrt{1-\bar{\alpha}_t} \boldsymbol{\epsilon}$ denotes the corrupted VAE feature of ground-truth image.

\subsection{Experimental Results}\label{artist_exp}
\begin{table*}[t]
    \centering
    \caption{Results on MARIO-Eval benchmark, our ARTIST outperforms previous methods.}
    \label{tab:main_results}
    \scalebox{0.75}{
        \begin{tabular}{cccccccc}
        \toprule
        Metrics & SD & Fine-tuned SD & ControlNet & DeepFloyd & TextDiffuser & ARTIST-TD & ARTIST\\
        \midrule
            FID $(\downarrow)$ & 51.295 & \textbf{28.761} & 51.485 & 34.902 & 38.758 &  36.579 & 38.43\\
            CLIP Score $(\uparrow)$ & 0.3015 & 0.3412 & 0.3424 & 0.3267 & 0.3436 & 0.3466& \textbf{0.3482}\\
            OCR Accuracy $(\uparrow)$ & 0.0178 & 0.0154 & 0.2705 & 0.0457 & 0.5712 & {0.6298} & \textbf{0.7373} \\
            OCR Precision $(\uparrow)$ & 0.0192 & 0.1777 &0.5391 & 0.1738 & 0.7795 & 0.8237 & \textbf{0.8681} \\
            OCR Recall $(\uparrow)$ & 0.0260 & 0.2330 & 0.6438 & 0.2235 & 0.7498 & {0.7986} & \textbf{0.8677} \\
            OCR F-measure $(\uparrow)$ & 0.0221 & 0.2016 & 0.5868 & 0.1955 & 0.7643 & {0.8110} & \textbf{0.8679}\\
        \bottomrule
        \end{tabular}
    }
\end{table*}

\begin{figure*}[t!]
    \centering
    \includegraphics[width=0.95\linewidth]{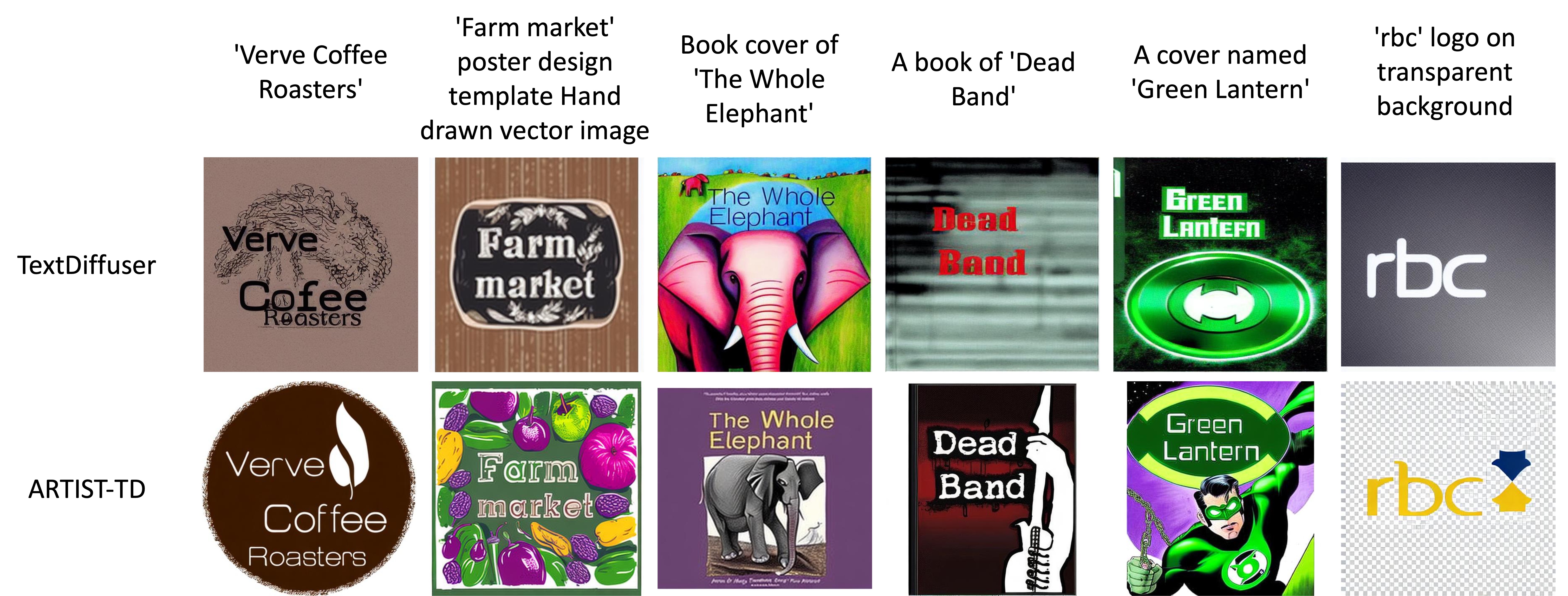}
    \vspace{5mm}
    \caption{Comparison with TextDiffuser on MARIO-Eval benchmark. Layout generated by TextDiffuser is used as input conditions for both models for fair comparison. }
    \label{fig:comparison}
\end{figure*}

\paragraph{Implementation Details} Our experiments are conducted on 8 Nvidia A100 GPUs, with Hugging Face Diffusers~\cite{von-platen-etal-2022-diffusers}. Both text and visual modules are initialized from the pre-trained Stable Diffusion \cite{rombach2021LDM} checkpoint. The text module is first pre-trained on our word-level synthetic dataset for 400,000 steps, then further fine-tuned for 200,000 steps on our sentence-level synthetic dataset. With the frozen text module, our visual module is trained for 250,000 steps on MARIO-10M dataset \cite{chen2023textdiffuser}. AdamW \cite{loshchilov2017AdamW} optimizer is used in all training stages, with a learning rate of 1e-5 and a weight decay of 1e-2. The batch size is set to 128.

We first compare our proposed framework with previous methods, including Stable Diffusion (denoted as SD), ControlNet, DeepFloyd and TextDiffuser in terms of common metrics such as OCR accuracy and FID.  We also finetune a Stable Diffusion on MARIO-10M dataset for a more comprehensive comparison. Then, we directly compare our method with the latest approaches like AnyText \cite{tuo2023anytext} and TextDiffuser 2 \cite{chen2023textdiffuser-2} through human evaluation, as it encapsulates the capabilities of the above traditional metrics while also assessing subjective aspects crucial for user experience and practical application effectiveness. Two variants of our proposed framework are evaluated, which are denoted as ARTIST-TD and ARTIST, respectively, indicating whether LLM is utilized. Specifically, ARTIST-TD directly uses the pretrained transformer from TextDiffuser instead of LLM to obtain bounding boxes and keywords based on input prompts. Thus, comparing ARTIST-TD with TextDiffuser can straightforwardly show the effectiveness of our training strategy, as they share exactly the same layout and keyword conditions.

Note that although there are two separate diffusion models in our framework, our computation requirement is still similar to the previous SOTA TextDiffuser. This is because TextDiffuser also utilizes an extra U-Net, which is designed for character-aware loss as a regularization term.

\paragraph{Results on MARIO-Eval benchmark} To start with, we conduct experiments on MARIO-Eval benchmark proposed in ~\cite{chen2023textdiffuser}, which contains 5,414 prompts in total. 4 images are generated for each prompt to compute the CLIP score, Fréchet Inception Distance (FID)~\cite{heusel2017FID} and OCR evaluations. Specifically, CLIP score is obtained by calculating the cosine similarity between generated images and prompts by using the features extracted with pre-trained ViT/B-32 CLIP model. OCR evaluation is performed with MaskSpotterv3~\cite{liao2020maskspotter} following \cite{chen2023textdiffuser}. The main results are presented in Table~\ref{tab:main_results}, from which we can see that our ARTIST outperforms the previous SOTA TextDiffuser in all metrics, even without the help of LLM. Some generated examples are provided in Figure \ref{fig:examples}. Our framework adeptly identifies the text intended to be generated in the image from the given prompts, regardless of explicit marking by quotes. The generated text is legible and complements the visual elements, enhancing the overall coherence of the design. More results are provided in the Appendix because of the limited space.  

For a more straightforward comparison, we also provide some generated examples in Figure~\ref{fig:comparison}. Although both TextDiffuser and ARTIST-TD are given the same layout and keyword conditions, we can see that ARTIST-TD generates images with better harmonization between painted text and background. ARTIST-TD is also able to generate many visual features that correspond to keywords such as ``Green Lantern'' and ``transparent'' in the prompt. This is because the learning of text structure and visual appearance is better disentangled, which leads to more efficient learning of both aspects compared to TextDiffuser.
\begin{figure*}[t!]
    \centering
    \includegraphics[width=0.9\linewidth]{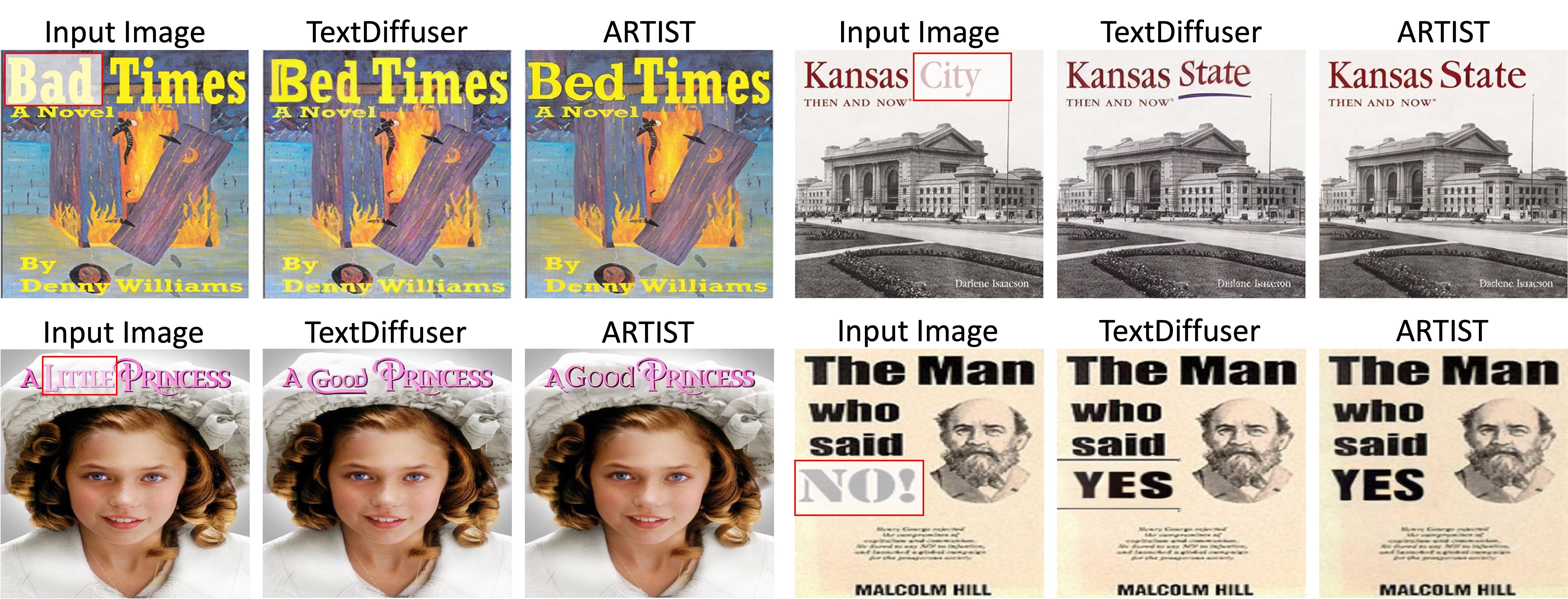}
    \vspace{5mm}
    \caption{Generated examples in inpainting task, where the masked regions are indicated by red rectangles. Prompts used are ``a book cover of Bed Times", ``a book cover for Kansas State", ``a poster for A Good Princess" and ``a poster for The Man Who Said YES".}
    \label{fig:inpainting} 
\end{figure*}

\paragraph{Image Inpainting} Similar to TextDiffuser, the proposed ARTIST can also be trained to perform an image-inpainting task. We trained another ARTIST following the setting from Chen et al.~\cite{chen2023textdiffuser} so that the model is asked to perform image inpainting instead of image generation with a probability of 0.5 during training. The resulting ARTIST is able to perform both tasks at inference time. Some generated examples are provided in Figure \ref{fig:inpainting}, along with the comparison between ARTIST and TextDiffuser. Similarly to the whole-image generation task, our ARTIST leads to more accurately rendered texts and more harmonized images.

\paragraph{ARTIST-Eval benchmark} As the reader may notice from Figure \ref{fig:examples} and \ref{fig:comparison}, most keywords in MARIO-Eval prompts are enclosed by quotation marks. In other words, a model can easily obtain promising accuracy in keyword prediction even if it is over-fitting and simply extracts words enclosed by quotation marks. Thus MARIO-Eval benchmark is not a reasonable benchmark to justify model performance on open-domain instructions. To this end, we propose a new ARTIST-Eval benchmark, which contains 500 pairs of prompts and keywords. Details about constructing the benchmark and some examples are provided in the Appendix \ref{app:artist_benchmark}. 
The keywords in ARTIST-Eval benchmark, which are expected to be shown in the resulting images, 
are designed to be contained in the prompts so that we can compare all the methods fairly.
However, keywords are not always enclosed by quotation marks. Similar to MARIO-Eval benchmark, 4 images are generated for each prompt to calculate CLIP score and OCR evaluations. The results on ARTIST-Eval benchmark are presented in Table \ref{tab:artist_benchmark}. From the results we can see that both ARTIST-TD and ARTIST outperform related methods, while ARTIST obtains a huge performance boost, indicating the benefits of using LLM.
\begin{table*}[t!] 
    \centering
    \caption{Results on ARTIST benchmark, our proposed framework outperforms all previous methods.}
    \label{tab:artist_benchmark}
    \scalebox{0.75}{
        \begin{tabular}{cccccccc}
        \toprule
        Metrics & SD & Fine-tuned SD & ControlNet & DeepFloyd & TextDiffuser & ARTIST-TD & ARTIST \\
        \midrule
            CLIP Score $(\uparrow)$ & 0.3387 & 0.3440 & 0.3105 & 0.3419 & 0.3100 & 0.3225 & \textbf{0.3545}\\
            OCR Accuracy $(\uparrow)$ & 0.0065 &  0.0300 &  0.0830 & 0.0545 & 0.1345 & 0.2185 & \textbf{0.6530} \\
            OCR Precision $(\uparrow)$ & 0.1031 &  0.1931 & 0.1885 & 0.2225 & 0.2160 & 0.2850 & \textbf{0.8166} \\
            OCR Recall $(\uparrow)$ &  0.1213 &  0.2352 &  0.2304  & 0.2669 & 0.1994 &  0.2779 & \textbf{0.8090} \\
            OCR F-measure $(\uparrow)$ & 0.1114 &  0.2121 & 0.2074 &  0.2427 & 0.2073 & 0.2814 & \textbf{0.8128} \\
        \bottomrule
        \end{tabular}
    }
\end{table*}

\paragraph{Keywords Identification} To quantitatively evaluate the contribution of LLM, we conduct experiments of keywords identification. Different models are asked to extract keywords from prompts of ARTIST-Eval benchmark.
The results are shown in Table \ref{tab:keywords_identification}, where we can conclude that LLM indeed leads to huge improvement in detecting keywords.
\begin{table}[ht]
    \centering
    \caption{Large-language models improves keywords identification by a large margin. }
    \vspace{5mm}
    \label{tab:keywords_identification}    
    \scalebox{1.1}{
        \begin{tabular}{ccccc}
        \toprule
         \multirow{2}{*}{Models} & \multicolumn{4}{c}{Keywords Identification Evaluation}\\
         & Acc $(\uparrow)$ & P $(\uparrow)$ & R $(\uparrow)$ & F1 $(\uparrow)$ \\
        \midrule
            TextDiffuser & 0.6320 & 0.6412  & 0.6397 & 0.6404 \\
            GPT-3.5-Turbo & 0.8300 & 0.9582 & 0.9496 & 0.9539 \\
            GPT-4 & 0.9380 & 0.9729 & 0.9893 & 0.9810 \\
        \bottomrule
        \end{tabular}
    }
\end{table}

\begin{figure}[h]
    \centering
    \includegraphics[width=0.95\linewidth]{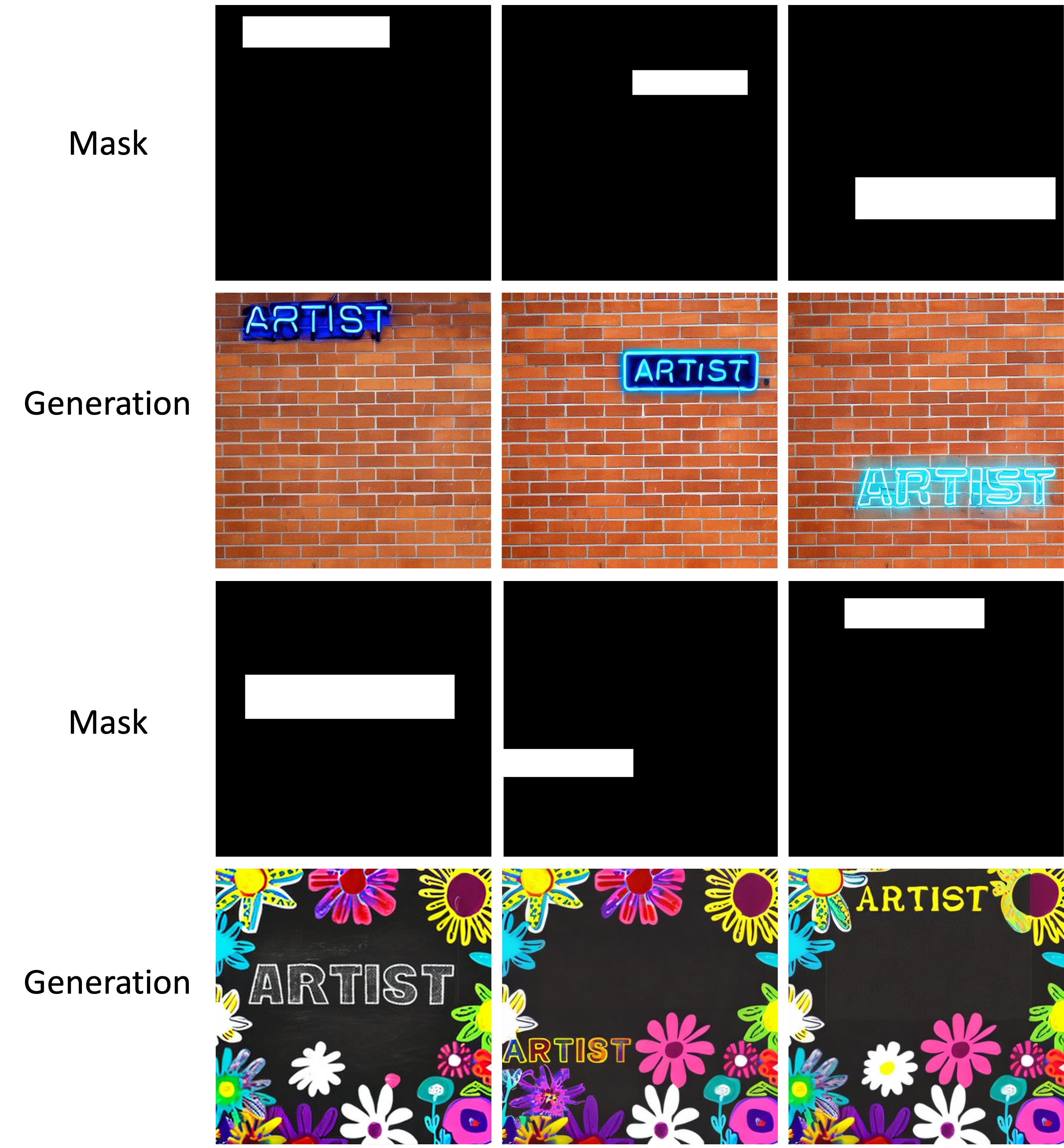}
    \vspace{5mm}
    \caption{The generated text will follow the layout information in terms of mask positions and sizes. Prompts used here are ``a neon light `ARTIST' on the brick wall" and ``word `ARTIST' surrounded by hand-drawn flowers".}
    \label{fig:mask_generation}
\end{figure}

\paragraph{Layout Adherence}
We conduct an ablation study to validate the precision of our model in adhering to the provided layout during inference. Illustrative examples depicted in Figure \ref{fig:mask_generation} demonstrate that the text generated by our method consistently aligns with the specified position and dimensions. This consistency underscores the model's capability to follow layout guidelines accurately.

\subsection{Summary}
We proposed ARTIST, a novel framework that significantly enhances the text-rendering ability of diffusion models. Our proposed framework utilizes pretrained large language models to infer the user's intention, provide accurate prompts, and improve the interactive experience. We also introduced a disentangled architecture design and training strategy which leads to better learning of both text structure and visual appearance. Our experimental results demonstrate that ARTIST outperforms the previous state-of-the-art in terms of image fidelity, image-prompt alignment, and accuracy of generated texts. 
In the future, we aim to improve the disentangled learning further, investigate the interpretability of the disentangled representations learned by ARTIST
and their potential for downstream tasks. Overall, we believe that ARTIST represents a significant step forward in the field of text-rich image generation and has the potential to enable new applications in the future. Ultimately, we envision Probability Engineering—exemplified by frameworks like ARTIST—becoming integral to realizing practical, high-quality multimodal generation, empowering more reliable and flexible AI systems in diverse real-world scenarios.

\chapter{Probability Engineering: An Overview}
\vspace{\baselineskip}
\section{Wide-Ranging Applications and Conceptual Foundations}

\noindent
\textbf{Case Studies and Scope of Probability Engineering.}
Throughout this dissertation, we have presented multiple case studies (including federated learning, knowledge distillation, Bayesian sampling, diffusion models, and large language models) to demonstrate the breadth of \emph{probability engineering}. These examples highlight how probability engineering can address challenges such as non-IID data distributions, resource constraints, evolving knowledge bases, or complex multimodal settings. By treating probability distributions as \emph{engineering artifacts}, we systematically modify, split, or refactor them to meet practical requirements, bridging the gap between theoretical probabilistic modeling and real-world AI applications.

\noindent
\textbf{Defining Probability Engineering.}
We use \emph{probability engineering} to describe the practice of \emph{designing, adapting, and utilizing} probability distributions with a specific, application-driven purpose. Unlike classical probabilistic modeling, which focuses on faithful distribution fitting or maximum-likelihood estimation, probability engineering \emph{prioritizes} domain constraints, computational efficiency, privacy, robustness, or other practical requirements. In other words, we do not merely aim to approximate the “true” distribution; rather, we incorporate engineering principles to ensure the modeled distribution aligns with the demands of deep learning systems in real-world deployments.

\noindent
\textbf{Comparisons with Traditional Probability Modeling.}
Traditional probability modeling is rooted in rigorous statistical foundations, often seeking to capture or infer underlying distributions with maximum fidelity. Probability engineering, by contrast, is \emph{application-oriented}, taking a more pragmatic stance: 
\begin{itemize}
    \item \emph{Objective:} Classical methods aim for minimal divergence from the real data distribution; probability engineering may \emph{trade off} exactness for computational or operational benefits.
    \item \emph{Scope:} Traditional modeling typically assumes relatively stable conditions; probability engineering readily accommodates \emph{dynamic distributions}, \emph{privacy constraints}, or \emph{limited data access}.
    \item \emph{Methodology:} Classical approaches often revolve around parametric modeling, Bayesian inference, or MCMC-like sampling; probability engineering introduces \emph{distribution manipulation} strategies, specialized data structures (e.g., memory modules), domain-specific constraints and so on to shape the distributions.
\end{itemize}

\noindent
\textbf{Relationship to Other “Engineering” Paradigms.}
In modern deep learning, we have seen “feature engineering,” “model engineering,” and “prompt engineering” all gain traction. Probability engineering intersects with these paradigms in that it, too, focuses on \emph{application-driven improvements}—but it does so at the level of probability distributions rather than features, model architectures, or textual prompts. Its \emph{advantage} is its \emph{breadth and flexibility}: almost any component or stage in the AI pipeline that involves uncertain or variable phenomena can be viewed through the lens of probability engineering. Conversely, its \emph{challenge} is that distributional manipulation can be more abstract than crafting features or building prompts, making it less intuitive or approachable for many practitioners. Nevertheless, once properly mastered, probability engineering offers a unifying perspective that can incorporate and extend many of the benefits from other engineering paradigms in AI.

\section{An Outline of the Probability Pioneering Process}

\noindent
\textbf{(1) Clarify Real-World Needs and Constraints.}
The first step is to identify exactly which challenges are not being met by existing modeling or system practices. These may include privacy constraints, computational limits, dynamic or streaming data conditions, or robust performance requirements. Importantly, the impetus might arise because classical probability modeling \emph{fails} to address certain real-world nuances, or because a specific engineering constraint (e.g., memory usage, non-stationarity, fairness) cannot be satisfied by vanilla distribution fitting.

\noindent
\textbf{(2) Identify the Relevant Distributions and Determine the Impacts.}
Next, we locate in the AI system the \emph{probability distribution(s)} that govern or influence the targeted requirement. This involves analyzing how data or model parameters flow through the system and deciding \emph{which distributions} (e.g., the posterior in Bayesian deep learning, the client selection distribution in FL, or the teacher–student embedding distribution in KD) are crucial for the desired application. We also ensure that these distributions are \emph{accessible} or \emph{derivable}, so that they can be explicitly manipulated.

\noindent
\textbf{(3) Disentangle Affected Components from the Original Distribution and Perform Engineering on These Components.}
We carefully observe whether a distribution $P$ is \emph{already influenced} by the newly introduced constraints or not. Two broad cases typically arise:
\begin{itemize}
    \item If $P$ is not materially impacted by the new requirements, then we might \emph{augment} $P$ with new terms or constraints that \emph{are} impacted. This approach layers an engineering module on top of existing model assumptions.
    \item If $P$ is influenced, we attempt to \emph{disentangle} the portion that is subject to these constraints from the unaffected portion. By isolating the “affected sub-distribution,” we can concentrate our engineering or modifications there, preserving whatever stable aspects of $P$ remain.
\end{itemize}

\noindent
\textbf{(4) Integrate Modified and Unmodified Components.}
Following the disentanglement, we effectively have two portions of our probability distribution: the one that remains unchanged and the one that has been reworked to accommodate the new constraints. These must be \emph{reconciled or fused} into a cohesive distribution. Often, this requires additional data, modules, or computations—for example, storing an external memory for knowledge distillation, or introducing a class-imbalance measure in FL. The final distribution is thus an \emph{engineered blend} that meets real-world demands without discarding the advantages of prior modeling assumptions.

\noindent
\textbf{(5) Deploy the Engineered Distribution in the Original AI Workflow.}
Finally, we \emph{apply} this newly formed or augmented distribution to the AI pipeline. This might be an inference process (e.g., performing retrieval during KD), a training loop (e.g., adjusting sampling strategies for FL), or a generative step. The key is that we have systematically shaped the distribution(s) to better fulfill the original real-world needs—be it privacy, efficiency, or improved accuracy under dynamic data.

\section{Future Outlook and Applications}

Probability engineering shows promise beyond the scenarios presented here. Here, we outline several promising directions where Probability Engineering could substantially advance AI capabilities:

\paragraph{Advanced Multimodal AI Systems}
Multimodal models, integrating vision, language, audio, and other modalities, often encounter complex distributional misalignments, especially when certain modalities are incomplete or noisy. Probability Engineering can explicitly align cross-modal distributions, such as employing distributional constraints or contrastive learning objectives, enabling robust and consistent integration across modalities. This alignment significantly improves model robustness under imperfect or missing data scenarios and facilitates efficient inference, especially important on resource-limited edge devices.

\paragraph{Reasoning $\&$ Inference-time Computing}
Current AI models, particularly large language models, struggle with tasks requiring multi-step reasoning and factual consistency, often due to overly simplistic decoding strategies that do not adequately explore probabilistic solution spaces. Probability Engineering provides principled ways to optimize inference-time decisions. We can leverage probabilistic reasoning paths to reconcile multiple candidate solutions, substantially enhancing logical coherence and factual correctness. Additionally, probability-based constraints can guide decoding toward valid and reliable outputs without the need for retraining or external supervision.

\paragraph{AI Agents}
AI agents deployed in dynamic, uncertain environments must manage partial observability and continual environmental changes. Probability Engineering can significantly enhance the adaptability of these agents by maintaining explicit probabilistic belief states. Techniques from Bayesian Reinforcement Learning, for instance, update the agent's belief distributions continuously based on observations, enabling robust decision-making under uncertainty. This approach ensures agents remain resilient and capable of adaptive behaviors, optimizing their policies even in previously unseen scenarios or shifting environments.

\paragraph{Efficiency of AI Systems}
The increasing scale of modern AI models imposes significant computational costs for training and inference, raising concerns around efficiency and sustainability. Probability Engineering can alleviate these burdens through probabilistically-informed sampling strategies, model compression techniques such as probabilistic knowledge distillation, and adaptive computation methods based on model confidence estimates. These approaches allow models to dynamically allocate computational resources, reducing unnecessary processing without compromising performance. Ultimately, this facilitates more scalable and sustainable deployment of AI technologies.

\paragraph{Safe and Trustworthy AI}
Ensuring fairness, robustness, and interpretability in high-stakes AI applications remains a critical challenge. Probability Engineering offers a structured approach to embedding ethical and domain-specific constraints directly into probabilistic models. For example, fairness can be enforced through independence constraints within probabilistic frameworks, preventing inadvertent bias. Robustness is enhanced by defining permissible distributions to mitigate adversarial inputs. Additionally, probabilistic frameworks inherently support interpretability and calibration, enabling clearer explanations of model decisions and facilitating trustworthiness through auditability.

Ultimately, as AI systems continue to expand in scale, diversity, and sophistication, the systematic design of probability distributions will be central to bridging the gap between idealized theory and pragmatic, real-world solutions. We envision probability engineering becoming an integral part of AI development, complementing existing best practices in feature selection, data collection, model architecture, prompt engineering, and beyond.

\chapter{Conclusions}

In this dissertation, we introduced the concept of \textit{probability engineering}, a practical and flexible methodology designed to align probabilistic modeling closely with the pragmatic constraints and dynamic challenges of modern deep learning applications. Unlike traditional probabilistic modeling, which often emphasizes theoretical rigor and fidelity to true data distributions, probability engineering takes a pragmatic stance—manipulating distributions to directly satisfy specific real-world requirements such as efficiency, robustness, privacy, and adaptability.

We began our exploration in Chapter 2 by addressing the complexity of distributions encountered in Bayesian Deep Learning. Here, we developed the \textit{Stochastic Particle-Optimization Sampling} (SPOS) algorithm, effectively enhancing Bayesian inference by engineering sampling dynamics. By incorporating variance reduction techniques and cyclical mechanisms, SPOS not only achieved superior convergence but also laid the groundwork for further applications of probability engineering in scalable Bayesian inference scenarios.

In Chapter 3, we extended the principles of probability engineering to Edge AI systems, focusing specifically on federated learning (FL) and knowledge distillation (KD). For federated learning, we proposed \textit{Federated Class-balanced Sampling} (Fed-CBS), which ingeniously engineered client sampling distributions to mitigate class imbalance, significantly outperforming standard methods. In the knowledge distillation context, we introduced \textit{Retrieval-Augmented Knowledge Distillation} (ReAugKD), leveraging an external memory to dynamically adapt student model distributions to evolving teacher knowledge, thereby greatly improving generalization and robustness in real-world NLP tasks.

We further explored probability engineering within generative AI scenarios, focusing on large language models (LLMs) and multimodal text-to-image generation tasks. Methods such as \textit{Self Logits Evolution Decoding} (SLED) and \textit{ARTIST} explicitly engineered distributions to ensure factual correctness in text generation and address multimodal complexities, respectively. These approaches demonstrated the versatility and effectiveness of probability engineering in resolving practical challenges faced by generative AI.

Looking forward, probability engineering has substantial potential in diverse applications, such as reinforcement learning, trustworthy and explainable AI, and automated generative systems. Future research could focus on standardized frameworks for probability engineering, scalable theoretical analysis, or the integration of this methodology into comprehensive federated and distillation paradigms, addressing ever-growing data complexity and system heterogeneity.

Ultimately, \textit{probability engineering} stands to become a core methodology in real-world AI design. By systematically restructuring probability distributions to better fit real-world constraints, we can deliver robust performance even in dynamic, data-scarce, or resource-limited environments. We anticipate that this thesis will inspire broader adoption of probability engineering, laying a foundation for more reliable, efficient, and adaptable AI systems in the future.


\begin{appendices}
	\titleformat{\chapter}[block]
	{
		\usefont{T1}{phv}{b}{n}\fontsize{16pt}{\normalbaselines}
		\normalbaselines
		\usefont{T1}{phv}{b}{n}\fontsize{16pt}{\normalbaselineskip}}
	{Appendix \thechapter.}{.2em}{\normalbaselines}
	\titlespacing\chapter{0pt}{-10pt plus 4pt minus 0pt}{6pt plus 2pt minus 0pt}
	
	\appendixtitleon

\section{Density Function of the Multi-Mode Distribution in Figure~\ref{fig:multimode}}\label{app:multimode}
The negative log-density function of the multi-mode distribution in Figure~\ref{fig:multimode} is defined as:
\begin{align*}
U(\bm{\theta}) \triangleq e^{\frac{3}{4}\bm{\theta}^2 - \frac{3}{2}\sum_{i=1}^{10}c_i \sin\left(\frac{1}{4}\pi i (\bm{\theta} + 4)\right)},
\end{align*}
where $c = (-0.47,-0.83,-0.71,-0.02,0.24,0.01,0.27,-0.37,0.87,-0.37)$ is a vector, $c_i$ is the $i$-th element of $c$.

\section{Algorithms for Variance Reduction in SPOS} \label{app:vrspos}
{\small 
\begin{algorithm} [!htbp]
\caption{SAGA-POS}
{\bf Input:} A set of initial particles $\{\theta_0^{(i)}\}_{i=1}^{M}$, each $\theta_0^{(i)}\in{\mathbb{R}^d}$, step size $h_k$, batch size $B$.\\
Initialize $\{g_{0,j}^{(i)}\}_{j=1}^{N}=\{F_j(\theta_0^{(i)})\}_{j=1}^{N}$ for all $i\in{\{1,...,M\}}$;
\begin{algorithmic}[1]\label{algo:algo1}
\FOR{iteration $k$= 0,1,...,T}
    \STATE Uniformly sample $I_k$ from $\{1,2,...,N\}$ randomly with replacement such that $|I_k|=B$; \\
    \STATE Sample $\xi_{k}^{(i)} \sim N($0$,I_{d \times d}),~\forall i~$;\\
    \STATE Update $G_k^{(i)}\leftarrow \sum\limits_{j=1}^{N}g_{k,j}^{(i)}+\frac{N}{B}\sum\limits_{j\in{I_k}}(F_j(\theta_k^{(i)})-g_{k,j}^{(i)}),~\forall i~$;\\
    \STATE Update each $\theta_k^{(i)}$ with Eq.(\ref{particle_num});\\
    \STATE Update $\{g_{k,j}^{(i)}\}_{j=1}^N,~~\forall i~$: if $j \in{I_k}$, set $g_{k+1,j}^{(i)}\leftarrow F_j(\theta_k^{(i)})$; else, set $g_{k+1,j}^{(i)}\leftarrow g_{k,j}^{(i)}$
\ENDFOR
\end{algorithmic}
{\bf Output:}{$\{\theta_T^{(i)}\}_{i=1}^M$}
\end{algorithm}
}

{\small 
\begin{algorithm} [h]
\caption{SVRG-POS}
{\bf Input:} A set of initial particles $\{\theta_0^{(i)}\}_{i=1}^{M}$, each $\theta_0^{(i)}\in{\mathbb{R}^d}$, step size $h$, epoch length $\tau$, batch size $B$.\\
Initialize $\{\widetilde{\theta}^{(i)}\}\leftarrow\{\theta_0^{(i)}\},\widetilde{G}^{(i)}\leftarrow F(\theta_0^{(i)}),~\forall i~$;
\begin{algorithmic}[1]\label{algo:algo2}
\FOR{iteration $k$= 0,1,...,T}
    \IF{k mod $\tau$ =0}
    \STATE{\textbf{Option \uppercase\expandafter{\romannumeral1}} $\RN{1}) $Sample $l\sim unif(0,1,..,\tau-1)$\\
    $\RN{2}) $Update $\widetilde{\theta}^{(i)} \leftarrow \theta_{k-l}^{(i)}$\\
    Update $\theta_{k}^{(i)} \leftarrow \widetilde{\theta}^{(i)},~\forall i~$\\
      $\RN{3}) $Update $\widetilde{G}^{(i)}\leftarrow F(\theta_k^{(i)}),~\forall i~$
      \\
      \STATE\textbf{Option \uppercase\expandafter{\romannumeral2}} $\RN{1})$ Update $\widetilde{\theta}^{(i)} \leftarrow \theta_k^{(i)}$\\
           $\RN{2})$Update $\widetilde{G}^{(i)}\leftarrow F(\theta_k^{(i)}),~\forall i~$\\
    }\ENDIF
    \STATE Uniformly sample $I_k$ from $\{1,2,...,N\}$ randomly with replacement such that $|I_k|=B$; \\
    \STATE Sample $\xi_{k}^{(i)} \sim N($0$,I_{d \times d}),~\forall i~$;\\
    \STATE Update $G_k^{(i)}\leftarrow \widetilde{G}^{(i)}+\frac{N}{B}\sum\limits_{j\in{I_k}}[F_j(\theta_k^{(i)})-F_j(\widetilde{\theta}^{(i)})],~\forall i~$;\\
    \STATE Update each $\theta_k^{(i)}$ with  Eq.(\ref{particle_num})
\ENDFOR
\end{algorithmic}
{\bf Output:}{$\{\theta_T^{(i)}\}_{i=1}^M$}
\end{algorithm}
}
{\small
\begin{algorithm}[!htbp]
\caption{SVRG-POS$^+$}
{\bf Input : } A set of initial particles $\{\theta_0^{(i)}\}_{i=1}^{M}$, each $\theta_0^{(i)}\in{\mathbb{R}^d}$, step size $h$, epoch length $\tau$, batch size $B$.\\
Initialize $\{\widetilde{\theta}^{(i)}\}\leftarrow\{\theta_0^{(i)}\},\widetilde{G}^{(i)}\leftarrow F(\theta_0^{(i)}),~\forall i~$;
\begin{algorithmic}[1]\label{algo:algo3}
\FOR{iteration $k$= 0,1,...,T}
    \IF{k mod $\tau$ =0}
    \STATE$\RN{1})$ Uniformly sample $J_k$ from $\{1,2,...,N\}$ with replacement such that $|J_k|=b$;\\
    $\RN{2})$ Update $\widetilde{\theta}^{(i)} \leftarrow \theta_k^{(i)}$\;
             $\widetilde{G}^{(i)}\leftarrow \frac{N}{b}\sum_{j\in J_k}F_j(\theta_k^{(i)}),~\forall i~$;
    \ENDIF
    \STATE Uniformly sample $I_k$ from $\{1,2,...,N\}$ with replacement such that $|I_k|=B$; \\
    \STATE Sample $\xi_{k}^{(i)} \sim N($0$,I_{d \times d}),~\forall i~$;\\
    \STATE Update $G_k^{(i)}\leftarrow \widetilde{G}^{(i)}+\frac{N}{B}\sum\limits_{j\in{I_k}}[F_j(\theta_k^{(i)})-F_j(\widetilde{\theta}^{(i)})],~\forall i~$;\\
    \STATE Update each $\theta_k^{(i)}$ with  Eq.(\ref{particle_num})
\ENDFOR
\end{algorithmic}
{\bf Output:}{$\{\theta_T^{(i)}\}_{i=1}^M$}
\end{algorithm}
}

\section{FHE framework}\label{FHEframework}

\begin{figure}[ht]
\begin{center}
\includegraphics[width=1.0\linewidth]{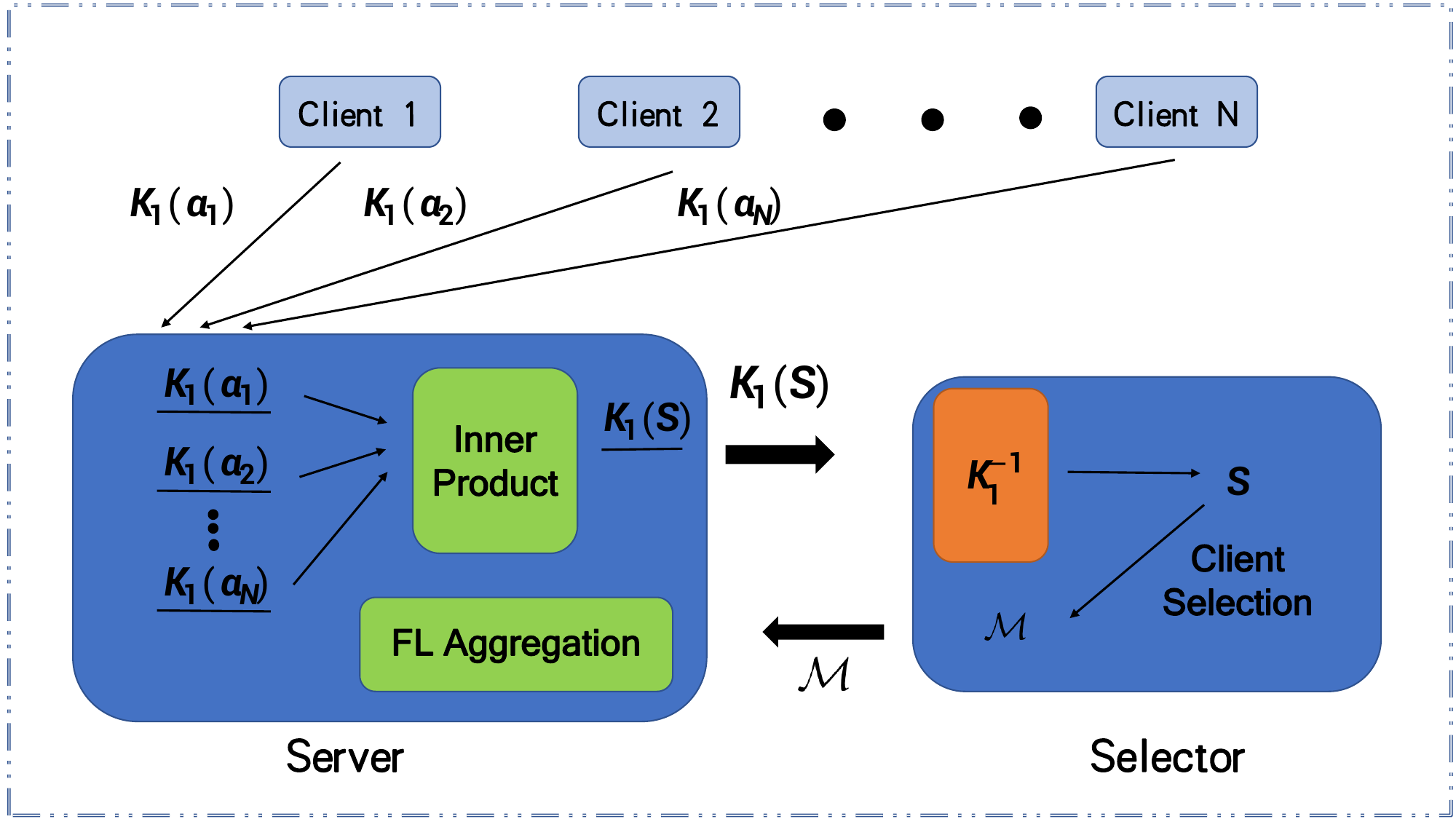}
\end{center}
\caption{An example of FHE to securely transmit ${S}$. }
\label{FHE}
\end{figure}

In Figure \ref{FHE}, we provide a framework as an example to show it is possible to deriving ${S}$ without knowing the values of local label distributions $\{\bm{\alpha}_i\}$ using FHE. Our framework can be realized using off-the-shelf FHE libraries such as~\cite{seal,helib}. 

There is a selector in our example. It is usually from third party and keeps a unique private key, denoted by $K_1^{-1}$. The corresponding public key is denoted by $K_1$. In the confidential transmission between the server and clients, each client first uses $K_1$ to encrypt their label distribution vector $\bm{\alpha_k}$ as $K_1(\bm{\alpha_k})$, and transmits it to the server. Since only the server has access to $K_1(\bm{\alpha_k})$, no one else including the selector can decrypt it and get $\bm{\alpha_k}$. When the server gets all $K_1(\bm{\alpha_k})$, it will conduct FHE computation to get the matrix $K_1(S)=K_1(\{\bm{\alpha}_i^T\bm{\alpha}_j\}_{ij})=\{K_1(\bm{\alpha}_i)^TK_1(\bm{\alpha}_j)\}_{ij}$. Then the server transmits the $K_1(S)$ to selector, and selector uses $K_1^{-1}$ to access the final result $S$. Since only the selector has $K_1^{-1}$, only it knows $S$. After that, the selector will conduct client selection following some strategy to derive the result $\mathcal{M}$ and transmit it back to the server. At last, the server will collect the model parameters of the clients in $\mathcal{M}$ and conduct FL aggregation. In the whole process, the server, selector or any other clients except client $n$ can not get $\alpha_n$. Furthermore, all clients and server have no access to the inner product results $S$, which prevents malicious clients or server inferring the label distributions of the other clients.

\section{Details of the Experiment Setup for Fed-CBS}\label{Sec:settingsec5}
For the MNIST dataset, we adopt an MLP model with one hidden layer of 64 units and FedAvg \cite{mcmahan2016communication} as the FL optimizer. 
 In Figure \ref{fig:fig1-sub-first}, we allocate the MNIST data to $N = 100$ clients with each client only accessing to the same amount of data from one class. In Figure \ref{fig:fig1-sub-second}, each client is associated with the same amount of data from two classes. In Figure \ref{fig:fig1-sub-third} and \ref{fig:fig1-sub-fourth}, we first allocate the whole MNIST dataset to $N = 200$ clients and pick 100 to construct a class-imbalanced global dataset. The global dataset with the 100 clients has the same amount of $n_1$ data samples for five classes and has the same amount of $n_2$ data samples for the other five classes. The ration r between $n_1$ and $n_2$ is set to $3:1$.
 
  In each training round (communication round), all the clients conduct 5 local training epochs. The batch size is $50$ for each client. The local optimizer is SGD with a weight decay of 0.0005. The learning rate is  0.01 initially, and the decay factor is 0.9992. We terminate the FL training after 200 training rounds (communication rounds) and then evaluate the model's performance on the test dataset of MNIST.

For the Cifar10 dataset, we adopt a model with two convolutional layers followed by three fully-connected layers and FedAvg \cite{mcmahan2016communication} as the FL optimizer. The model we adopt has two convolutional layers with the number of the kernels being 6 and 16, respectively. And all convolution kernels are of size 5 × 5. The outputs of convolutional layers are fed into two hidden layers  with 120 and 84 units. The batch size is $50$ for each client. In each communication round, all of them conduct the same number of local updates, which allows the client with the largest local dataset to conduct 5 local training epochs. In our method, we set the $\beta_m=m$, $\gamma=10$ and $L_{b}=10^{-20}$. The local optimizer is SGD with a weight decay of 0.0005. The learning rate is  0.01 initially and the decay factor is 0.9992. We terminate the FL training after 3000 communication rounds and then evaluate the model's performance on the test dataset of CIFAR-10. 
In our implementation of the Power-of-choice selection strategy (pow-d)\cite{Cho2020ClientSI}, we first sample a candidate set $\mathcal{A}$ of 20 clients without replacement such that client $n$ is chosen with probability proportional to the size of their local dataset $q_n$. Then the server sends the current global model to the clients in set $\mathcal{A}$, and these clients compute and send back to the server their local loss. To derive $\mathcal{M}$, we select M clients which have the highest loss from  $\mathcal{A}$.

In our implementation of the method in \cite{yang2020federated} (Fed-cucb), the exploration factor to balance the trade-off between exploitation and exploration is set as  0.2 and the forgetting factor  as 0.99, which are the same as the settings in \cite{yang2020federated}.

With help from FHE, we can derive the matrix of inner products $S$ accurately. Hence, in the simulation of our method, Fed-CBS, we ignore the process of deriving $S$ and focus on our sampling strategy.

\section{ARTIST-Eval benchmark}\label{app:artist_benchmark}
\begin{figure}[htp]
    \centering
    \includegraphics[width=.95\linewidth]{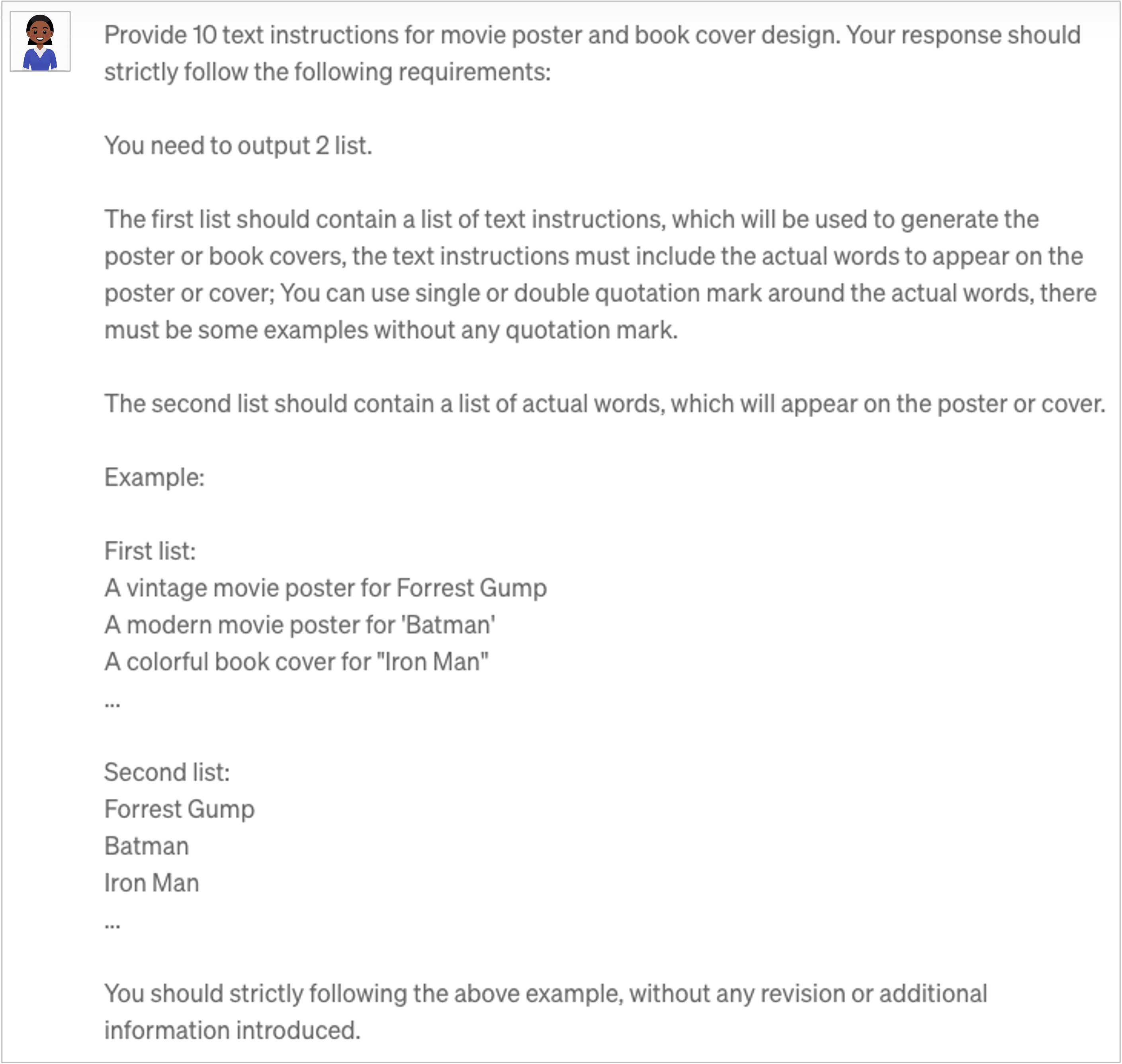}
    \vspace{5mm}
    \caption{Prompt we used to construct our benchmark with GPT-4.}
    \label{fig:benchmark_prompt}
\end{figure}

Here we provide some examples in our ARTIST-Eval benchmark:
\begin{itemize}
    \item A vintage movie poster for Forrest Gump
    \item A modern movie poster for `Batman'
    \item A colorful book cover for ``Iron Man"
    \item A minimalist movie poster for The Godfather
    \item An abstract movie poster for `Pulp Fiction'
    \item A gothic book cover for ``Dracula"
    \item A romantic movie poster for The Notebook
    \item A futuristic movie poster for `Blade Runner'
    \item A watercolor book cover for ``Pride and Prejudice"
    \item A playful movie poster for Finding Nemo
\end{itemize}
Our benchmark is constructed by prompting GPT-4~\cite{OpenAI2023GPT4TR}, the prompt we used is shown in Figure \ref{fig:benchmark_prompt}.

\section{More Results for ARTIST}

\begin{figure}[h]
    \centering
    \includegraphics[width=1.0\linewidth]{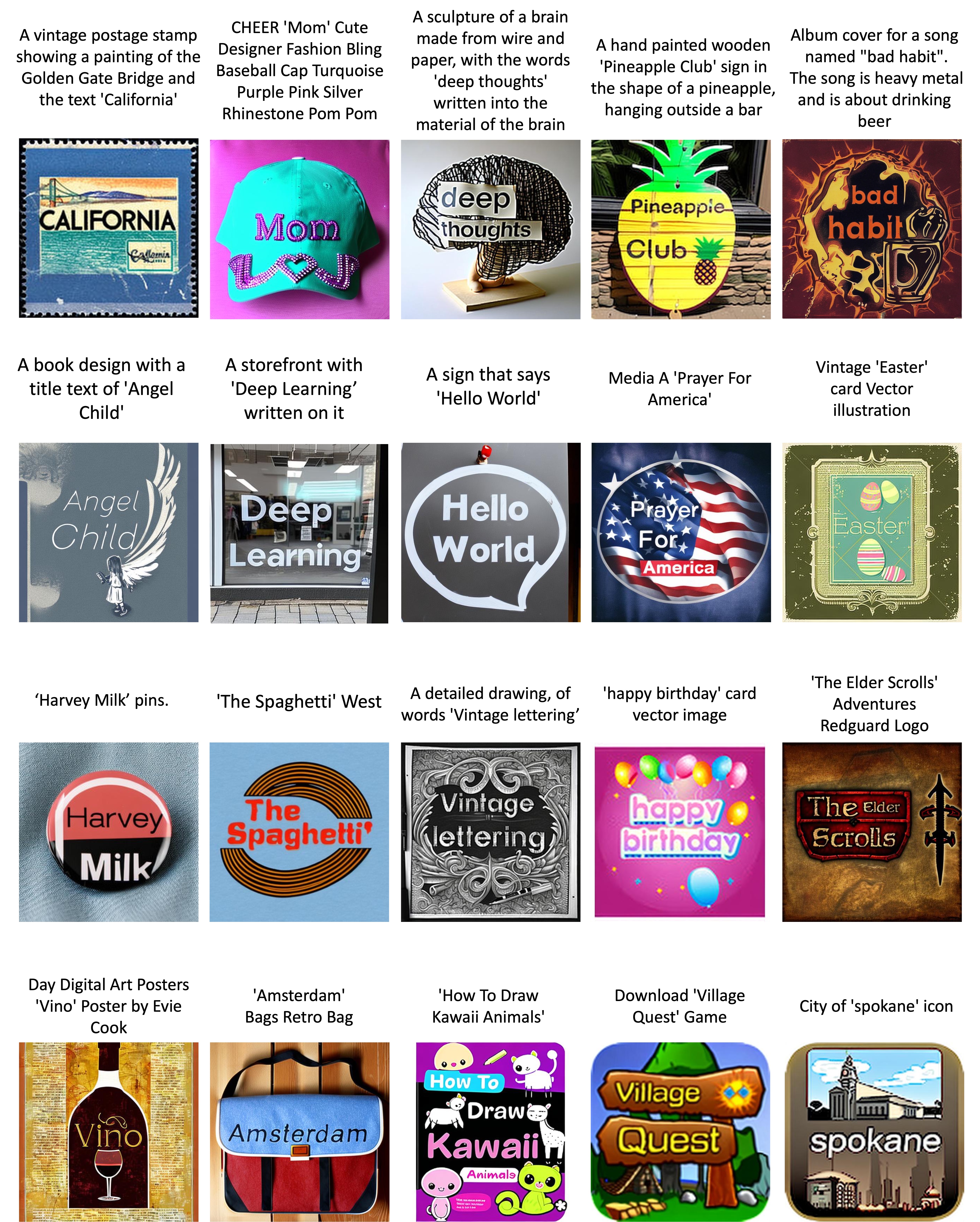}
    \caption{More generated results from the proposed framework.}
    \label{fig:more_results}
\end{figure}

\section{Additional Analysis for SLED} \label{more_results}

\paragraph{Justification on the Gradients Approximation of SLED in Section \ref{method_starting_point}} To further support our method's mechanism, which utilizes $\mathit{logits}_\mathit{n} - \mathit{logits}_\mathit{N}$ to approximate the gradient of \(KL(\mathcal{P}_{\mathit{real}},\mathcal{P}_{\mathit{logits}})\) at \(\mathit{logits}= \mathit{logits}_\mathit{n}\), we manually calculate the $Cosine\_similarity(\mathit{logits}_\mathit{n} - \mathit{logits}_\mathit{N},\nabla_{\mathit{logits}} KL(\mathcal{P}_\mathit{real}, \mathcal{P}_\mathit{logits}) |_{\mathit{logits}=\mathit{logits}_\mathit{n}})$ among thousands of tokens and layers. We plot the density function for different models. We find that the majority of these values are
positive, demonstrating that the directions of these two vectors are very close. Hence, our gradient approximation strategy in Section \ref{method_starting_point} is reasonable.
\begin{figure}[h]
    \centering
        \centering
        \includegraphics[width=1\textwidth]{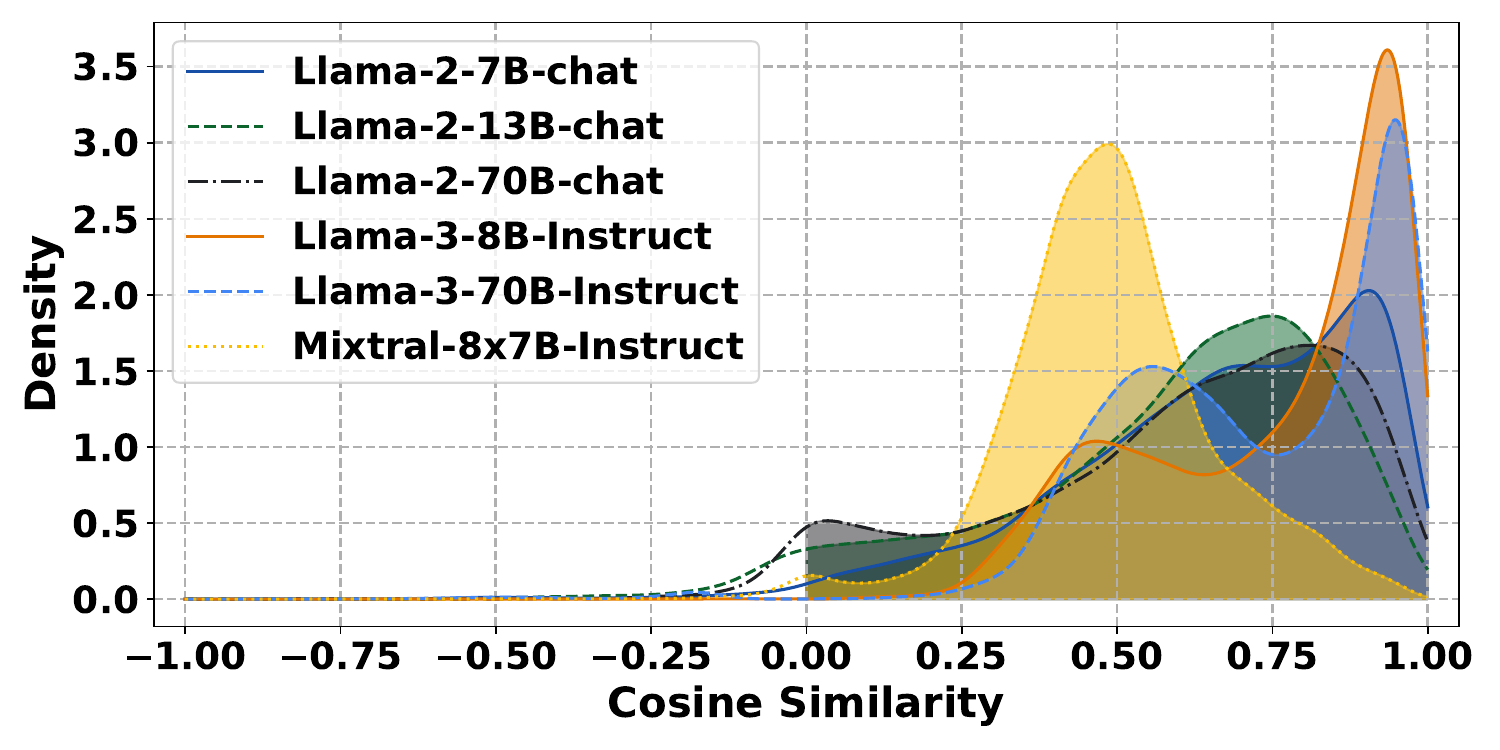}
        \caption{We collect 10k pairs of $(\mathit{logits}_\mathit{n} - \mathit{logits}_\mathit{N},\nabla_{\mathit{logits}_\mathit{n}} KL(\mathcal{P}_\mathit{real}, \mathcal{P}_{\mathit{logits}_\mathit{n}}))$ on different tokens in FACTOR and different early layers. We calculate their cosine similarity values and draw the density function for each LLM. Most of the pairs have positive Cosine similarity values, which verifies that the approximation strategy of SLED is reasonable.}
        \label{fig:model_gradient_approx_cos} 
\end{figure}

\section{Qualitative Studies of SLED}\label{sec:repetition}

We present some examples from the StrategyQA dataset in Table~\ref{tab:repetition} to illustrate that our method addresses the repetition issue of DoLa.

\begin{table}[h]
\renewcommand{\arraystretch}{1.5}
\centering
\begin{tabular}{m{.75cm}m{12cm}}
\toprule
\multicolumn{2}{l}{\textbf{Q:} Can you buy a fair trade laptop?} \\
\midrule
\bf DoLa & \it A laptop is a portable computer. A portable computer is a handheld computer. A handheld computer is a palmtop computer. A palmtop computer is a PDA. A PDA is a personal digital assistant. A personal digital assistant is a PDA. A PDA is a handheld computer. A handheld computer is a palmtop computer. A palmtop computer is a PDA. A PDA is a personal digital assistant. A personal digital assistant is a PDA. Thus, a PDA is a PDA. A PDA is a PDA. A PDA is a PDA. A PDA is a PDA. A PDA is a PDA. A PDA is a PDA ... \\
\midrule
\bf SLED & \it Fair Trade is a certification for coffee. Laptops are not coffee. Thus, you cannot buy a fair trade laptop. So the answer is no.\\
\bottomrule
\toprule
\multicolumn{2}{l}{\textbf{Q:} Does Thiago Moises May 13 2020 submission move hypothetically hurt Achilles?} \\
\midrule
\bf DoLa & \it Thiago Moises is a MMA fighter. He is 1.73 m tall, and Achilles is about \textit{0.000000000000000000000000000000000000000000000000000...}\\
\midrule
\bf SLED & \it 
Thiago Moises's submission was a rear naked choke, which is a blood choke. A blood choke can cause a fighter to pass out. Thus, Thiago Moises's submission could hypothetically hurt Achilles. So the answer is yes.
\\
\bottomrule
\end{tabular}
\vspace{3mm}
\caption{We present three examples of the generated text from LLaMA-2-13B-Base on StrategyQA dataset. SLED method can mitigate the repetition issue.}\label{tab:repetition}
\end{table} 
\end{appendices}



\addcontentsline{toc}{chapter}{Bibliography} 



\clearpage
\bibliographystyle{unsrtnat} 
\bibliography{mybib_dedup} 

\begin{thebibliography}{198}
\providecommand{\natexlab}[1]{#1}
\providecommand{\url}[1]{\texttt{#1}}
\expandafter\ifx\csname urlstyle\endcsname\relax
  \providecommand{\doi}[1]{doi: #1}\else
  \providecommand{\doi}{doi: \begingroup \urlstyle{rm}\Url}\fi

\bibitem[Zhou et~al.(2022)Zhou, Xu, and McAuley]{zhou2022bert}
Wangchunshu Zhou, Canwen Xu, and Julian McAuley.
\newblock Bert learns to teach: Knowledge distillation with meta learning.
\newblock In \emph{Proceedings of the 60th Annual Meeting of the Association for Computational Linguistics (Volume 1: Long Papers)}, pages 7037--7049, 2022.

\bibitem[LeCun et~al.(2015)LeCun, Bengio, and Hinton]{lecun2015deep}
Yann LeCun, Yoshua Bengio, and Geoffrey Hinton.
\newblock Deep learning.
\newblock \emph{nature}, 521\penalty0 (7553):\penalty0 436--444, 2015.

\bibitem[Ho et~al.(2020)Ho, Jain, and Abbeel]{ho2020denoising}
Jonathan Ho, Ajay Jain, and Pieter Abbeel.
\newblock Denoising diffusion probabilistic models.
\newblock \emph{Advances in neural information processing systems}, 33:\penalty0 6840--6851, 2020.

\bibitem[Rombach et~al.(2022)Rombach, Blattmann, Lorenz, Esser, and Ommer]{rombach2022high}
Robin Rombach, Andreas Blattmann, Dominik Lorenz, Patrick Esser, and Bj{\"o}rn Ommer.
\newblock High-resolution image synthesis with latent diffusion models.
\newblock In \emph{Proceedings of the IEEE/CVF conference on computer vision and pattern recognition}, pages 10684--10695, 2022.

\bibitem[Ramesh et~al.(2022)Ramesh, Dhariwal, Nichol, Chu, and Chen]{ramesh2022dalle2}
Aditya Ramesh, Prafulla Dhariwal, Alex Nichol, Casey Chu, and Mark Chen.
\newblock Hierarchical text-conditional image generation with clip latents.
\newblock \emph{arXiv preprint arXiv:2204.06125}, 1\penalty0 (2):\penalty0 3, 2022.

\bibitem[Saharia et~al.(2022{\natexlab{a}})Saharia, Chan, Saxena, Li, Whang, Denton, Ghasemipour, Gontijo~Lopes, Karagol~Ayan, Salimans, et~al.]{saharia2022photorealistic}
Chitwan Saharia, William Chan, Saurabh Saxena, Lala Li, Jay Whang, Emily~L Denton, Kamyar Ghasemipour, Raphael Gontijo~Lopes, Burcu Karagol~Ayan, Tim Salimans, et~al.
\newblock Photorealistic text-to-image diffusion models with deep language understanding.
\newblock \emph{Advances in Neural Information Processing Systems}, 35:\penalty0 36479--36494, 2022{\natexlab{a}}.

\bibitem[AI@Meta(2024)]{llama3modelcard}
AI@Meta.
\newblock Llama 3 model card.
\newblock 2024.
\newblock URL \url{https://github.com/meta-llama/llama3/blob/main/MODEL_CARD.md}.

\bibitem[Anil et~al.(2023)Anil, Dai, Firat, Johnson, Lepikhin, Passos, Shakeri, Taropa, Bailey, Chen, et~al.]{anil2023palm}
Rohan Anil, Andrew~M Dai, Orhan Firat, Melvin Johnson, Dmitry Lepikhin, Alexandre Passos, Siamak Shakeri, Emanuel Taropa, Paige Bailey, Zhifeng Chen, et~al.
\newblock Palm 2 technical report.
\newblock \emph{arXiv preprint arXiv:2305.10403}, 2023.

\bibitem[OpenAI(2022)]{openai-chatgpt}
OpenAI.
\newblock Introducing chatgpt.
\newblock \url{https://openai.com/blog/chatgpt/}, November 2022.

\bibitem[{OpenAI}(2023)]{GPT4report}
{OpenAI}.
\newblock {GPT-4 Technical Report}.
\newblock \emph{arXiv e-prints}, art. arXiv:2303.08774, March 2023.
\newblock \doi{10.48550/arXiv.2303.08774}.

\bibitem[Team et~al.(2023)Team, Anil, Borgeaud, Wu, Alayrac, Yu, Soricut, Schalkwyk, Dai, Hauth, et~al.]{team2023gemini}
Gemini Team, Rohan Anil, Sebastian Borgeaud, Yonghui Wu, Jean-Baptiste Alayrac, Jiahui Yu, Radu Soricut, Johan Schalkwyk, Andrew~M Dai, Anja Hauth, et~al.
\newblock Gemini: a family of highly capable multimodal models.
\newblock \emph{arXiv preprint arXiv:2312.11805}, 2023.

\bibitem[Touvron et~al.(2023{\natexlab{a}})Touvron, Lavril, Izacard, Martinet, Lachaux, Lacroix, Rozi{\`e}re, Goyal, Hambro, Azhar, et~al.]{touvron2023llama}
Hugo Touvron, Thibaut Lavril, Gautier Izacard, Xavier Martinet, Marie-Anne Lachaux, Timoth{\'e}e Lacroix, Baptiste Rozi{\`e}re, Naman Goyal, Eric Hambro, Faisal Azhar, et~al.
\newblock Llama: Open and efficient foundation language models.
\newblock \emph{arXiv preprint arXiv:2302.13971}, 2023{\natexlab{a}}.

\bibitem[Wang et~al.(2024{\natexlab{a}})Wang, Si, Yu, Wiesmann, Hsieh, and Dhillon]{wang2024large}
Ruochen Wang, Si~Si, Felix Yu, Dorothea Wiesmann, Cho-Jui Hsieh, and Inderjit Dhillon.
\newblock Large language models are interpretable learners.
\newblock \emph{arXiv preprint arXiv:2406.17224}, 2024{\natexlab{a}}.

\bibitem[Kingma and Welling(2013)]{kingma2013auto}
Diederik~P Kingma and Max Welling.
\newblock Auto-encoding variational bayes.
\newblock \emph{arXiv preprint arXiv:1312.6114}, 2013.

\bibitem[James et~al.(2013)James, Witten, Hastie, and Tibshirani]{James2013}
Gareth James, Daniela Witten, Trevor Hastie, and Robert Tibshirani.
\newblock \emph{An Introduction to Statistical Learning: with Applications in R}.
\newblock Springer, 2013.
\newblock URL \url{https://faculty.marshall.usc.edu/gareth-james/ISL/}.

\bibitem[Zhang et~al.(2020{\natexlab{a}})Zhang, Zhang, Carin, and Chen]{zhang2020stochastic}
Jianyi Zhang, Ruiyi Zhang, Lawrence Carin, and Changyou Chen.
\newblock Stochastic particle-optimization sampling and the non-asymptotic convergence theory.
\newblock In \emph{International Conference on Artificial Intelligence and Statistics}, pages 1877--1887. PMLR, 2020{\natexlab{a}}.

\bibitem[Zhang et~al.(2024{\natexlab{a}})Zhang, Zhou, Gu, Wigington, Yu, Chen, Sun, and Zhang]{zhang2024artistimprovinggenerationtextrich}
Jianyi Zhang, Yufan Zhou, Jiuxiang Gu, Curtis Wigington, Tong Yu, Yiran Chen, Tong Sun, and Ruiyi Zhang.
\newblock Artist: Improving the generation of text-rich images with disentangled diffusion models and large language models, 2024{\natexlab{a}}.
\newblock URL \url{https://arxiv.org/abs/2406.12044}.

\bibitem[Joren et~al.(2024)Joren, Zhang, Ferng, Juan, Taly, and Rashtchian]{joren2024sufficient}
Hailey Joren, Jianyi Zhang, Chun-Sung Ferng, Da-Cheng Juan, Ankur Taly, and Cyrus Rashtchian.
\newblock Sufficient context: A new lens on retrieval augmented generation systems.
\newblock \emph{arXiv preprint arXiv:2411.06037}, 2024.

\bibitem[Zhang et~al.(2023{\natexlab{a}})Zhang, Muhamed, Anantharaman, Wang, Chen, Zhong, Cui, Xu, Zeng, Chilimbi, et~al.]{zhang2023reaugkd}
Jianyi Zhang, Aashiq Muhamed, Aditya Anantharaman, Guoyin Wang, Changyou Chen, Kai Zhong, Qingjun Cui, Yi~Xu, Belinda Zeng, Trishul Chilimbi, et~al.
\newblock Reaugkd: Retrieval-augmented knowledge distillation for pre-trained language models.
\newblock In \emph{Proceedings of the 61st Annual Meeting of the Association for Computational Linguistics (Volume 2: Short Papers)}, pages 1128--1136, 2023{\natexlab{a}}.

\bibitem[Zhang et~al.(2024{\natexlab{b}})Zhang, Vahidian, Kuo, Li, Zhang, Yu, Wang, and Chen]{zhang2024towards}
Jianyi Zhang, Saeed Vahidian, Martin Kuo, Chunyuan Li, Ruiyi Zhang, Tong Yu, Guoyin Wang, and Yiran Chen.
\newblock Towards building the federatedgpt: Federated instruction tuning.
\newblock In \emph{ICASSP 2024-2024 IEEE International Conference on Acoustics, Speech and Signal Processing (ICASSP)}, pages 6915--6919. IEEE, 2024{\natexlab{b}}.

\bibitem[Zhang et~al.(2024{\natexlab{c}})Zhang, Juan, Rashtchian, Ferng, Jiang, and Chen]{zhang2024sled}
Jianyi Zhang, Da-Cheng Juan, Cyrus Rashtchian, Chun-Sung Ferng, Heinrich Jiang, and Yiran Chen.
\newblock Sled: Self logits evolution decoding for improving factuality in large language models.
\newblock \emph{arXiv preprint arXiv:2411.02433}, 2024{\natexlab{c}}.

\bibitem[Yang et~al.(2021)Yang, Zhang, Hao, Spell, and Carin]{flop_qian2021}
Qian Yang, Jianyi Zhang, Weituo Hao, Gregory~P. Spell, and Lawrence Carin.
\newblock Flop: Federated learning on medical datasets using partial networks.
\newblock In \emph{Proceedings of the 27th ACM SIGKDD Conference on Knowledge Discovery \& Data Mining}, KDD '21, page 3845–3853, New York, NY, USA, 2021. Association for Computing Machinery.
\newblock ISBN 9781450383325.
\newblock \doi{10.1145/3447548.3467185}.
\newblock URL \url{https://doi.org/10.1145/3447548.3467185}.

\bibitem[Zhang et~al.(2023{\natexlab{b}})Zhang, Li, Tang, Sun, Chen, Zhang, Chen, Chen, and Li]{fedcbs}
Jianyi Zhang, Ang Li, Minxue Tang, Jingwei Sun, Xiang Chen, Fan Zhang, Changyou Chen, Yiran Chen, and Hai Li.
\newblock Fed-cbs: A heterogeneity-aware client sampling mechanism for federated learning via class-imbalance reduction.
\newblock In \emph{Proceedings of the 40th International Conference on Machine Learning}, 2023{\natexlab{b}}.

\bibitem[Liu et~al.(2021)Liu, Yuan, Fu, Jiang, Hayashi, and Neubig]{liu2021pretrainpromptpredictsystematic}
Pengfei Liu, Weizhe Yuan, Jinlan Fu, Zhengbao Jiang, Hiroaki Hayashi, and Graham Neubig.
\newblock Pre-train, prompt, and predict: A systematic survey of prompting methods in natural language processing, 2021.
\newblock URL \url{https://arxiv.org/abs/2107.13586}.

\bibitem[Chen et~al.(2015)Chen, Ding, and Carin]{ChenDC:NIPS15}
C.~Chen, N.~Ding, and L.~Carin.
\newblock On the convergence of stochastic gradient {MCMC} algorithms with high-order integrators.
\newblock In \emph{Neural Information Processing Systems (NIPS)}, 2015.

\bibitem[Wilson and Izmailov(2022)]{wilson2022bayesiandeeplearningprobabilistic}
Andrew~Gordon Wilson and Pavel Izmailov.
\newblock Bayesian deep learning and a probabilistic perspective of generalization, 2022.
\newblock URL \url{https://arxiv.org/abs/2002.08791}.

\bibitem[Welling and Teh(2011)]{WellingT:ICML11}
M.~Welling and Y.~W. Teh.
\newblock Bayesian learning via stochastic gradient {L}angevin dynamics.
\newblock In \emph{ICML}, 2011.

\bibitem[Chen et~al.(2014)Chen, Fox, and Guestrin]{ChenFG:ICML14}
T.~Chen, E.~B. Fox, and C.~Guestrin.
\newblock Stochastic gradient {H}amiltonian {M}onte {C}arlo.
\newblock In \emph{International Conference on Machine Learning (ICML)}, 2014.

\bibitem[Ding et~al.(2014)Ding, Fang, Babbush, Chen, Skeel, and Neven]{DingFBCSN:NIPS14}
N.~Ding, Y.~Fang, R.~Babbush, C.~Chen, R.~D. Skeel, and H.~Neven.
\newblock Bayesian sampling using stochastic gradient thermostats.
\newblock In \emph{Neural Information Processing Systems (NIPS)}, 2014.

\bibitem[Liu et~al.(2017)Liu, Ramachandran, Liu, and Peng]{liu2017stein}
Y.~Liu, P.~Ramachandran, Q.~Liu, and J.~Peng.
\newblock Stein variational policy gradient.
\newblock In \emph{UAI}, 2017.

\bibitem[Gill et~al.(2024)Gill, Golec, Hu, Xu, Du, Wu, Walia, Murugesan, Ali, Kumar, Ye, Verma, Kumar, Cuadrado, and Uhlig]{Gill_2024}
Sukhpal~Singh Gill, Muhammed Golec, Jianmin Hu, Minxian Xu, Junhui Du, Huaming Wu, Guneet~Kaur Walia, Subramaniam~Subramanian Murugesan, Babar Ali, Mohit Kumar, Kejiang Ye, Prabal Verma, Surendra Kumar, Felix Cuadrado, and Steve Uhlig.
\newblock Edge ai: A taxonomy, systematic review and future directions.
\newblock \emph{Cluster Computing}, 28\penalty0 (1), October 2024.
\newblock ISSN 1573-7543.
\newblock \doi{10.1007/s10586-024-04686-y}.
\newblock URL \url{http://dx.doi.org/10.1007/s10586-024-04686-y}.

\bibitem[Singh and Gill(2023)]{SINGH202371}
Raghubir Singh and Sukhpal~Singh Gill.
\newblock Edge ai: A survey.
\newblock \emph{Internet of Things and Cyber-Physical Systems}, 3:\penalty0 71--92, 2023.
\newblock ISSN 2667-3452.
\newblock \doi{https://doi.org/10.1016/j.iotcps.2023.02.004}.
\newblock URL \url{https://www.sciencedirect.com/science/article/pii/S2667345223000196}.

\bibitem[Hinton et~al.(2015)Hinton, Vinyals, Dean, et~al.]{hinton2015distilling}
Geoffrey Hinton, Oriol Vinyals, Jeff Dean, et~al.
\newblock Distilling the knowledge in a neural network.
\newblock \emph{arXiv preprint arXiv:1503.02531}, 2\penalty0 (7), 2015.

\bibitem[Liu and Wang(2016)]{liu2016stein}
Qiang Liu and Dilin Wang.
\newblock Stein variational gradient descent: A general purpose bayesian inference algorithm.
\newblock In \emph{NIPS}, 2016.

\bibitem[Kolmogoroff(1931)]{Kolmogoroff:MA31}
A.~Kolmogoroff.
\newblock Some studies in machine learning using the game of checkers.
\newblock \emph{Mathematische Annalen}, 104\penalty0 (1):\penalty0 415--458, 1931.

\bibitem[Risken(1989)]{Risken:FPE89}
H.~Risken.
\newblock \emph{The {F}okker-{P}lanck equation}.
\newblock Springer-Verlag, New York, 1989.

\bibitem[Gan et~al.(2015)Gan, Chen, Henao, Carlson, and Carin]{GanCHCC:icml15}
Z.~Gan, C.~Chen, R.~Henao, D.~Carlson, and L.~Carin.
\newblock Scalable deep {P}oisson factor analysis for topic modeling.
\newblock In \emph{ICML}, 2015.

\bibitem[Liu et~al.(2016)Liu, Zhu, and Song]{LiuZS:NIPS16}
C.~Liu, J.~Zhu, and Y.~Song.
\newblock Stochastic gradient geodesic {MCMC} methods.
\newblock In \emph{NIPS}, 2016.

\bibitem[\c{S}im\c{s}ekli et~al.(2016)\c{S}im\c{s}ekli, Badeau, Cemgil, and Richard]{SBCG:ICML16}
U.~\c{S}im\c{s}ekli, R.~Badeau, A.~T. Cemgil, and G.~Richard.
\newblock Stochastic {Q}uasi-{N}ewton {L}angevin {M}onte {C}arlo.
\newblock In \emph{ICML}, 2016.

\bibitem[Wang et~al.(2015)Wang, Fienberg, and Smola]{WangFS:icml15}
Y.~X. Wang, S.~E. Fienberg, and A.~Smola.
\newblock Privacy for free: Posterior sampling and stochastic gradient {M}onte {C}arlo.
\newblock In \emph{ICML}, 2015.

\bibitem[Li et~al.(2017)Li, Chen, Liu, and Carin]{LiCLC:arxiv17}
B.~Li, C.~Chen, H.~Liu, and L.~Carin.
\newblock On connecting stochastic gradient {MCMC} and differential privacy.
\newblock Technical Report arXiv:1712.09097, 2017.
\newblock URL \url{http://arxiv.org/abs/1712.09097}.

\bibitem[Springenberg et~al.(2016)Springenberg, Klein, Falkner, and Hutter]{SpringenbergKFH:NIPS16}
J.~T. Springenberg, A.~Klein, S.~Falkner, and F.~Hutter.
\newblock Bayesian optimization with robust {B}ayesian neural networks.
\newblock In \emph{NIPS}, 2016.

\bibitem[Haarnoja et~al.(2018)Haarnoja, Tang, Abbeel, and Levine]{haarnoja2017reinforcement}
Tuomas Haarnoja, Haoran Tang, Pieter Abbeel, and Sergey Levine.
\newblock Reinforcement learning with deep energy-based policies.
\newblock In \emph{ICML}, 2018.

\bibitem[Zhang et~al.(2018{\natexlab{a}})Zhang, Chen, Li, and Duke]{zhang2018policy}
Ruiyi Zhang, Changyou Chen, Chunyuan Li, and Lawrence~Carin Duke.
\newblock Policy optimization as wasserstein gradient flows.
\newblock In \emph{International Conference on Machine Learning}, 2018{\natexlab{a}}.

\bibitem[Zhang et~al.(2019)Zhang, Wen, Chen, and Carin]{zhang2018scalable}
Ruiyi Zhang, Zheng Wen, Changyou Chen, and Lawrence Carin.
\newblock Scalable thompson sampling via optimal transport.
\newblock In \emph{AISTATS}, 2019.

\bibitem[Li et~al.(2016)Li, Chen, Carlson, and Carin]{PSGLD:AAAI16}
C.~Li, C.~Chen, D.~Carlson, and L.~Carin.
\newblock Preconditioned stochastic gradient {L}angevin dynamics for deep neural networks.
\newblock In \emph{AAAI}, 2016.

\bibitem[Chen et~al.(2018)Chen, Zhang, Wang, Li, and Chen]{ChenZWLC:UAI18}
C.~Chen, R.~Zhang, W.~Wang, B.~Li, and L.~Chen.
\newblock A unified particle-optimization framework for scalable {B}ayesian sampling.
\newblock In \emph{UAI}, 2018.

\bibitem[Li et~al.(2015)Li, Hern{\'a}ndez-Lobato, and Turner]{li2015stochastic}
Y.~Li, J.~Hern{\'a}ndez-Lobato, and R.~E. Turner.
\newblock Stochastic expectation propagation.
\newblock In \emph{NIPS}, 2015.

\bibitem[Zhang et~al.(2018{\natexlab{b}})Zhang, Li, Chen, and Carin]{ZhangLCC:AISTATS18}
R.~Zhang, C.~Li, C.~Chen, and L.~Carin.
\newblock Learning structural weight uncertainty for sequential decision-making.
\newblock In \emph{AISTATS}, 2018{\natexlab{b}}.

\bibitem[Houthooft et~al.(2016)Houthooft, Chen, Duan, Schulman, De~Turck, and Abbeel]{houthooft2016vime}
Rein Houthooft, Xi~Chen, Yan Duan, John Schulman, Filip De~Turck, and Pieter Abbeel.
\newblock {VIME}: Variational information maximizing exploration.
\newblock In \emph{NIPS}, 2016.

\bibitem[Zhang et~al.(2020{\natexlab{b}})Zhang, Zhao, and Chen]{zhang2020variance}
Jianyi Zhang, Yang Zhao, and Changyou Chen.
\newblock Variance reduction in stochastic particle-optimization sampling.
\newblock In \emph{International Conference on Machine Learning}, pages 11307--11316. PMLR, 2020{\natexlab{b}}.

\bibitem[Zhang et~al.(2020{\natexlab{c}})Zhang, Li, Zhang, Chen, and Wilson]{Zhang2020Cyclical}
Ruqi Zhang, Chunyuan Li, Jianyi Zhang, Changyou Chen, and Andrew~Gordon Wilson.
\newblock Cyclical stochastic gradient mcmc for bayesian deep learning.
\newblock In \emph{International Conference on Learning Representations}, 2020{\natexlab{c}}.
\newblock URL \url{https://openreview.net/forum?id=rkeS1RVtPS}.

\bibitem[Zhao et~al.(2019)Zhao, Zhang, and Chen]{zhao2019self}
Yang Zhao, Jianyi Zhang, and Changyou Chen.
\newblock Self-adversarially learned bayesian sampling.
\newblock In \emph{Proceedings of the AAAI Conference on Artificial Intelligence}, volume~33, pages 5893--5900, 2019.

\bibitem[Chen et~al.(2022)Chen, Zhang, Xu, Chen, Duan, Chen, Tran, Zeng, and Chilimbi]{NEURIPS2022_db174d37}
Changyou Chen, Jianyi Zhang, Yi~Xu, Liqun Chen, Jiali Duan, Yiran Chen, Son Tran, Belinda Zeng, and Trishul Chilimbi.
\newblock Why do we need large batchsizes in contrastive learning? a gradient-bias perspective.
\newblock In S.~Koyejo, S.~Mohamed, A.~Agarwal, D.~Belgrave, K.~Cho, and A.~Oh, editors, \emph{Advances in Neural Information Processing Systems}, volume~35, pages 33860--33875. Curran Associates, Inc., 2022.
\newblock URL \url{https://proceedings.neurips.cc/paper_files/paper/2022/file/db174d373133dcc6bf83bc98e4b681f8-Paper-Conference.pdf}.

\bibitem[Dubey et~al.(2016)Dubey, Reddi, P\'{o}czos, Smola, and Xing]{DubeyRPSX:nips16}
A.~Dubey, S.~J. Reddi, B.~P\'{o}czos, A.~J. Smola, and E.~P. Xing.
\newblock Variance reduction in stochastic gradient {L}angevin dynamics.
\newblock In \emph{NIPS}, 2016.

\bibitem[Chatterji et~al.(2018)Chatterji, Flammarion, Ma, Bartlett, and Jordan]{Nilardri:2018}
Niladri Chatterji, Nicolas Flammarion, Yi-An Ma, Peter Bartlett, and Michael Jordan.
\newblock On the theory of variance reduction for stochastic gradient monte carlo.
\newblock \emph{ICML}, 2018.

\bibitem[Zou et~al.(2018)Zou, Xu, and Gu]{DIFAN:2018}
Difan Zou, Pan Xu, and Quanquan Gu.
\newblock Subsampled stochastic variance-reduced gradient langevin dynamics.
\newblock \emph{UAI}, 2018.

\bibitem[Defazio et~al.(2014)Defazio, Bach, and Lacoste-Julien]{SAGA:DA2014}
Aaron Defazio, Francis Bach, and Simon Lacoste-Julien.
\newblock Saga: A fast incremental gradient method with support for non-strongly convex composite objectives.
\newblock \emph{Nips}, 2014.

\bibitem[Johnson and Zhang(2013)]{RTASGD:AG2013}
Rie Johnson and Tong Zhang.
\newblock Accelerating stochastic gradient descent using predictive variance reduction.
\newblock \emph{NIPS}, 2013.

\bibitem[McMahan et~al.(2017{\natexlab{a}})McMahan, Moore, Ramage, Hampson, and Arcas]{pmlr-v54-mcmahan17a}
Brendan McMahan, Eider Moore, Daniel Ramage, Seth Hampson, and Blaise Aguera~y Arcas.
\newblock {Communication-Efficient Learning of Deep Networks from Decentralized Data}.
\newblock In Aarti Singh and Jerry Zhu, editors, \emph{Proceedings of the 20th International Conference on Artificial Intelligence and Statistics}, volume~54 of \emph{Proceedings of Machine Learning Research}, pages 1273--1282. PMLR, 20--22 Apr 2017{\natexlab{a}}.
\newblock URL \url{https://proceedings.mlr.press/v54/mcmahan17a.html}.

\bibitem[Goetz et~al.(2019)Goetz, Malik, Bui, Moon, Liu, and Kumar]{goetz2019active}
Jack Goetz, Kshitiz Malik, Duc Bui, Seungwhan Moon, Honglei Liu, and Anuj Kumar.
\newblock Active federated learning.
\newblock \emph{arXiv preprint arXiv:1909.12641}, 2019.

\bibitem[Cho et~al.(2020)Cho, Wang, and Joshi]{Cho2020ClientSI}
Yae~Jee Cho, Jianyu Wang, and Gauri Joshi.
\newblock Client selection in federated learning: Convergence analysis and power-of-choice selection strategies.
\newblock \emph{ArXiv}, abs/2010.01243, 2020.

\bibitem[Nishio and Yonetani(2019)]{Nishio2019ClientSF}
Takayuki Nishio and Ryo Yonetani.
\newblock Client selection for federated learning with heterogeneous resources in mobile edge.
\newblock \emph{ICC 2019 - 2019 IEEE International Conference on Communications (ICC)}, pages 1--7, 2019.

\bibitem[Yang et~al.(2020)Yang, Wong, Zhu, Wang, and Qian]{yang2020federated}
Miao Yang, Akitanoshou Wong, Hongbin Zhu, Haifeng Wang, and Hua Qian.
\newblock Federated learning with class imbalance reduction, 2020.

\bibitem[Balakrishnan et~al.(2021)Balakrishnan, Li, Zhou, Himayat, Smith, and Bilmes]{balakrishnan2021}
Ravikumar Balakrishnan, Tian Li, Tianyi Zhou, Nageen Himayat, Virginia Smith, and Jeff Bilmes.
\newblock Diverse client selection for federated learning: Submodularity and convergence analysis.
\newblock In \emph{ICML 2021 International Workshop on Federated Learning for User Privacy and Data Confidentiality}, Virtual, July 2021.

\bibitem[Ribero and Vikalo(2020)]{ribero2020communication}
Monica Ribero and Haris Vikalo.
\newblock Communication-efficient federated learning via optimal client sampling.
\newblock \emph{arXiv preprint arXiv:2007.15197}, 2020.

\bibitem[Duan et~al.(2019)Duan, Liu, Chen, Tan, Ren, Qiao, and Liang]{Duan2019AstraeaSF}
Moming Duan, Duo Liu, Xianzhang Chen, Yujuan Tan, Jinting Ren, Lei Qiao, and Liang Liang.
\newblock Astraea: Self-balancing federated learning for improving classification accuracy of mobile deep learning applications.
\newblock In \emph{2019 IEEE 37th international conference on computer design (ICCD)}, pages 246--254. IEEE, 2019.

\bibitem[Anand et~al.(1993)Anand, Mehrotra, Mohan, and Ranka]{anand1993improved}
Rangachari Anand, Kishan~G Mehrotra, Chilukuri~K Mohan, and Sanjay Ranka.
\newblock An improved algorithm for neural network classification of imbalanced training sets.
\newblock \emph{IEEE Transactions on Neural Networks}, 4\penalty0 (6):\penalty0 962--969, 1993.

\bibitem[Brakerski et~al.(2014)Brakerski, Gentry, and Vaikuntanathan]{bgvfhe}
Zvika Brakerski, Craig Gentry, and Vinod Vaikuntanathan.
\newblock (leveled) fully homomorphic encryption without bootstrapping.
\newblock \emph{ACM Transactions on Computation Theory (TOCT)}, 6\penalty0 (3):\penalty0 1--36, 2014.

\bibitem[Fan and Vercauteren(2012)]{fvfhe}
Junfeng Fan and Frederik Vercauteren.
\newblock Somewhat practical fully homomorphic encryption.
\newblock \emph{IACR Cryptol. ePrint Arch.}, 2012:\penalty0 144, 2012.

\bibitem[Halevi and Shoup(2015)]{halevi2015bootstrapping}
Shai Halevi and Victor Shoup.
\newblock Bootstrapping for {HElib}.
\newblock In \emph{Annual International conference on the theory and applications of cryptographic techniques}, pages 641--670. Springer, 2015.

\bibitem[Hsu et~al.(2019)Hsu, Qi, and Brown]{Hsu2019MeasuringTE}
Tzu-Ming~Harry Hsu, Hang Qi, and Matthew Brown.
\newblock Measuring the effects of non-identical data distribution for federated visual classification.
\newblock \emph{ArXiv}, abs/1909.06335, 2019.

\bibitem[Zhang et~al.(2024{\natexlab{d}})Zhang, Vahidian, Kuo, Li, Zhang, Yu, Zhou, Wang, and Chen]{zhang2024buildingfederatedgptfederated}
Jianyi Zhang, Saeed Vahidian, Martin Kuo, Chunyuan Li, Ruiyi Zhang, Tong Yu, Yufan Zhou, Guoyin Wang, and Yiran Chen.
\newblock Towards building the federated gpt: Federated instruction tuning, 2024{\natexlab{d}}.
\newblock URL \url{https://arxiv.org/abs/2305.05644}.

\bibitem[Hao et~al.(2021)Hao, El-Khamy, Lee, Zhang, Liang, Chen, and Duke]{hao2021towards}
Weituo Hao, Mostafa El-Khamy, Jungwon Lee, Jianyi Zhang, Kevin~J Liang, Changyou Chen, and Lawrence~Carin Duke.
\newblock Towards fair federated learning with zero-shot data augmentation.
\newblock In \emph{Proceedings of the IEEE/CVF Conference on Computer Vision and Pattern Recognition}, pages 3310--3319, 2021.

\bibitem[Du et~al.(2022)Du, Sun, Li, Chen, Zhang, Li, and Chen]{du2022rethinking}
Zhixu Du, Jingwei Sun, Ang Li, Pin-Yu Chen, Jianyi Zhang, Hai"~Helen" Li, and Yiran Chen.
\newblock Rethinking normalization methods in federated learning.
\newblock In \emph{Proceedings of the 3rd International Workshop on Distributed Machine Learning}, pages 16--22, 2022.

\bibitem[Zhang et~al.(2022{\natexlab{a}})Zhang, Du, Sun, Li, Tang, Wu, Gao, Kuo, Li, and Chen]{zhang2022next}
Jianyi Zhang, Zhixu Du, Jingwei Sun, Ang Li, Minxue Tang, Yuhao Wu, Zhihui Gao, Martin Kuo, Hai-Helen Li, and Yiran Chen.
\newblock Next generation federated learning for edge devices: An overview.
\newblock In \emph{2022 IEEE 8th International Conference on Collaboration and Internet Computing (CIC)}, pages 10--15. IEEE, 2022{\natexlab{a}}.

\bibitem[Tang et~al.(2023)Tang, Zhang, Ma, DiValentin, Ding, Hassanzadeh, Li, and Chen]{tang2023fade}
Minxue Tang, Jianyi Zhang, Mingyuan Ma, Louis DiValentin, Aolin Ding, Amin Hassanzadeh, Hai Li, and Yiran Chen.
\newblock {FADE}: Enabling large-scale federated adversarial training on resource-constrained edge devices, 2023.
\newblock URL \url{https://openreview.net/forum?id=NzrpxT5hTY_}.

\bibitem[Jia et~al.(2025)Jia, Vahidian, Sun, Zhang, Kungurtsev, Gong, and Chen]{jia2025unlocking}
Yuqi Jia, Saeed Vahidian, Jingwei Sun, Jianyi Zhang, Vyacheslav Kungurtsev, Neil~Zhenqiang Gong, and Yiran Chen.
\newblock Unlocking the potential of federated learning: The symphony of dataset distillation via deep generative latents.
\newblock In Ale{\v{s}} Leonardis, Elisa Ricci, Stefan Roth, Olga Russakovsky, Torsten Sattler, and G{\"u}l Varol, editors, \emph{Computer Vision -- ECCV 2024}, pages 18--33, Cham, 2025. Springer Nature Switzerland.
\newblock ISBN 978-3-031-73229-4.

\bibitem[Yao et~al.(2024)Yao, Zhang, Wu, Huang, Xia, Yu, Zhang, Kim, Rossi, Li, et~al.]{yao2024federated}
Yuhang Yao, Jianyi Zhang, Junda Wu, Chengkai Huang, Yu~Xia, Tong Yu, Ruiyi Zhang, Sungchul Kim, Ryan Rossi, Ang Li, et~al.
\newblock Federated large language models: Current progress and future directions.
\newblock \emph{arXiv preprint arXiv:2409.15723}, 2024.

\bibitem[Zhang et~al.(2024{\natexlab{e}})Zhang, Yang, Li, Guo, Wang, Wang, Chen, and Li]{zhang2024mllm}
Jianyi Zhang, Hao~Frank Yang, Ang Li, Xin Guo, Pu~Wang, Haiming Wang, Yiran Chen, and Hai Li.
\newblock Mllm-fl: Multimodal large language model assisted federated learning on heterogeneous and long-tailed data.
\newblock \emph{arXiv preprint arXiv:2409.06067}, 2024{\natexlab{e}}.

\bibitem[Devlin et~al.(2018)Devlin, Chang, Lee, and Toutanova]{devlin2018bert}
Jacob Devlin, Ming-Wei Chang, Kenton Lee, and Kristina Toutanova.
\newblock Bert: Pre-training of deep bidirectional transformers for language understanding.
\newblock \emph{arXiv preprint arXiv:1810.04805}, 2018.

\bibitem[Liu et~al.(2019)Liu, Ott, Goyal, Du, Joshi, Chen, Levy, Lewis, Zettlemoyer, and Stoyanov]{liu2019roberta}
Yinhan Liu, Myle Ott, Naman Goyal, Jingfei Du, Mandar Joshi, Danqi Chen, Omer Levy, Mike Lewis, Luke Zettlemoyer, and Veselin Stoyanov.
\newblock Roberta: A robustly optimized bert pretraining approach.
\newblock \emph{arXiv preprint arXiv:1907.11692}, 2019.

\bibitem[Radford et~al.(2018)Radford, Narasimhan, Salimans, Sutskever, et~al.]{radford2018improving}
Alec Radford, Karthik Narasimhan, Tim Salimans, Ilya Sutskever, et~al.
\newblock Improving language understanding by generative pre-training.
\newblock 2018.

\bibitem[Clark et~al.(2020)Clark, Luong, Le, and Manning]{clark2020electra}
Kevin Clark, Minh-Thang Luong, Quoc~V Le, and Christopher~D Manning.
\newblock Electra: Pre-training text encoders as discriminators rather than generators.
\newblock \emph{arXiv preprint arXiv:2003.10555}, 2020.

\bibitem[Ding et~al.(2019)Ding, Zhou, Chen, Yang, and Tang]{ding2019cognitive}
Ming Ding, Chang Zhou, Qibin Chen, Hongxia Yang, and Jie Tang.
\newblock Cognitive graph for multi-hop reading comprehension at scale.
\newblock \emph{arXiv preprint arXiv:1905.05460}, 2019.

\bibitem[Wang et~al.(2018)Wang, Singh, Michael, Hill, Levy, and Bowman]{wang2018glue}
Alex Wang, Amanpreet Singh, Julian Michael, Felix Hill, Omer Levy, and Samuel~R Bowman.
\newblock Glue: A multi-task benchmark and analysis platform for natural language understanding.
\newblock \emph{arXiv preprint arXiv:1804.07461}, 2018.

\bibitem[Park et~al.(2019)Park, Kim, Lu, and Cho]{park2019relational}
Wonpyo Park, Dongju Kim, Yan Lu, and Minsu Cho.
\newblock Relational knowledge distillation.
\newblock In \emph{Proceedings of the IEEE/CVF Conference on Computer Vision and Pattern Recognition}, pages 3967--3976, 2019.

\bibitem[Zhang et~al.(2018{\natexlab{c}})Zhang, Xiang, Hospedales, and Lu]{zhang2018deep}
Ying Zhang, Tao Xiang, Timothy~M Hospedales, and Huchuan Lu.
\newblock Deep mutual learning.
\newblock In \emph{Proceedings of the IEEE conference on computer vision and pattern recognition}, pages 4320--4328, 2018{\natexlab{c}}.

\bibitem[Mirzadeh et~al.(2020)Mirzadeh, Farajtabar, Li, Levine, Matsukawa, and Ghasemzadeh]{mirzadeh2020improved}
Seyed~Iman Mirzadeh, Mehrdad Farajtabar, Ang Li, Nir Levine, Akihiro Matsukawa, and Hassan Ghasemzadeh.
\newblock Improved knowledge distillation via teacher assistant.
\newblock In \emph{Proceedings of the AAAI conference on artificial intelligence}, volume~34, pages 5191--5198, 2020.

\bibitem[Jin et~al.(2019)Jin, Peng, Wu, Liu, Liu, Liang, Yan, and Hu]{jin2019knowledge}
Xiao Jin, Baoyun Peng, Yichao Wu, Yu~Liu, Jiaheng Liu, Ding Liang, Junjie Yan, and Xiaolin Hu.
\newblock Knowledge distillation via route constrained optimization.
\newblock In \emph{Proceedings of the IEEE/CVF International Conference on Computer Vision}, pages 1345--1354, 2019.

\bibitem[Sun et~al.(2019)Sun, Cheng, Gan, and Liu]{sun2019patient}
Siqi Sun, Yu~Cheng, Zhe Gan, and Jingjing Liu.
\newblock Patient knowledge distillation for bert model compression.
\newblock \emph{arXiv preprint arXiv:1908.09355}, 2019.

\bibitem[Jiao et~al.(2019)Jiao, Yin, Shang, Jiang, Chen, Li, Wang, and Liu]{jiao2019tinybert}
Xiaoqi Jiao, Yichun Yin, Lifeng Shang, Xin Jiang, Xiao Chen, Linlin Li, Fang Wang, and Qun Liu.
\newblock Tinybert: Distilling bert for natural language understanding.
\newblock \emph{arXiv preprint arXiv:1909.10351}, 2019.

\bibitem[Sun et~al.(2020)Sun, Yu, Song, Liu, Yang, and Zhou]{sun2020mobilebert}
Zhiqing Sun, Hongkun Yu, Xiaodan Song, Renjie Liu, Yiming Yang, and Denny Zhou.
\newblock Mobilebert: a compact task-agnostic bert for resource-limited devices.
\newblock \emph{arXiv preprint arXiv:2004.02984}, 2020.

\bibitem[Sanh et~al.(2019)Sanh, Debut, Chaumond, and Wolf]{sanh2019distilbert}
Victor Sanh, Lysandre Debut, Julien Chaumond, and Thomas Wolf.
\newblock Distilbert, a distilled version of bert: smaller, faster, cheaper and lighter.
\newblock \emph{arXiv preprint arXiv:1910.01108}, 2019.

\bibitem[Wang et~al.(2020{\natexlab{a}})Wang, Bao, Huang, Dong, and Wei]{wang2020minilmv2}
Wenhui Wang, Hangbo Bao, Shaohan Huang, Li~Dong, and Furu Wei.
\newblock Minilmv2: Multi-head self-attention relation distillation for compressing pretrained transformers.
\newblock \emph{arXiv preprint arXiv:2012.15828}, 2020{\natexlab{a}}.

\bibitem[Xu et~al.(2019)Xu, Li, Zhang, Wen, Wang, Dai, Qi, Chen, Lin, and Xiong]{TRPEDNN}
Yuhui Xu, Yuxi Li, Shuai Zhang, Wei Wen, Botao Wang, Wenrui Dai, Yingyong Qi, Yiran Chen, Weiyao Lin, and Hongkai Xiong.
\newblock Trained rank pruning for efficient deep neural networks.
\newblock In \emph{2019 Fifth Workshop on Energy Efficient Machine Learning and Cognitive Computing - NeurIPS Edition (EMC2-NIPS)}, pages 14--17, 2019.
\newblock \doi{10.1109/EMC2-NIPS53020.2019.00011}.

\bibitem[Weston et~al.(2018)Weston, Dinan, and Miller]{weston2018retrieve}
Jason Weston, Emily Dinan, and Alexander~H Miller.
\newblock Retrieve and refine: Improved sequence generation models for dialogue.
\newblock \emph{arXiv preprint arXiv:1808.04776}, 2018.

\bibitem[Lewis et~al.(2020)Lewis, Perez, Piktus, Petroni, Karpukhin, Goyal, K{\"u}ttler, Lewis, Yih, Rockt{\"a}schel, et~al.]{lewis2020retrieval}
Patrick Lewis, Ethan Perez, Aleksandra Piktus, Fabio Petroni, Vladimir Karpukhin, Naman Goyal, Heinrich K{\"u}ttler, Mike Lewis, Wen-tau Yih, Tim Rockt{\"a}schel, et~al.
\newblock Retrieval-augmented generation for knowledge-intensive nlp tasks.
\newblock \emph{Advances in Neural Information Processing Systems}, 33:\penalty0 9459--9474, 2020.

\bibitem[Guu et~al.(2020)Guu, Lee, Tung, Pasupat, and Chang]{guu2020retrieval}
Kelvin Guu, Kenton Lee, Zora Tung, Panupong Pasupat, and Mingwei Chang.
\newblock Retrieval augmented language model pre-training.
\newblock In \emph{International Conference on Machine Learning}, pages 3929--3938. PMLR, 2020.

\bibitem[Lin et~al.(2022{\natexlab{a}})Lin, Tan, Miller, Tian, and Ren]{lin2022unsupervised}
Bill~Yuchen Lin, Kangmin Tan, Chris Miller, Beiwen Tian, and Xiang Ren.
\newblock Unsupervised cross-task generalization via retrieval augmentation.
\newblock \emph{arXiv preprint arXiv:2204.07937}, 2022{\natexlab{a}}.

\bibitem[Kassner and Sch{\"u}tze(2020)]{kassner2020bert}
Nora Kassner and Hinrich Sch{\"u}tze.
\newblock Bert-knn: Adding a knn search component to pretrained language models for better qa.
\newblock \emph{arXiv preprint arXiv:2005.00766}, 2020.

\bibitem[Khosla et~al.(2020)Khosla, Teterwak, Wang, Sarna, Tian, Isola, Maschinot, Liu, and Krishnan]{khosla2020supervised}
Prannay Khosla, Piotr Teterwak, Chen Wang, Aaron Sarna, Yonglong Tian, Phillip Isola, Aaron Maschinot, Ce~Liu, and Dilip Krishnan.
\newblock Supervised contrastive learning.
\newblock \emph{Advances in Neural Information Processing Systems}, 33:\penalty0 18661--18673, 2020.

\bibitem[Li et~al.(2021)Li, Song, Ma, Qiu, and Huang]{li2021knn}
Linyang Li, Demin Song, Ruotian Ma, Xipeng Qiu, and Xuanjing Huang.
\newblock Knn-bert: fine-tuning pre-trained models with knn classifier.
\newblock \emph{arXiv preprint arXiv:2110.02523}, 2021.

\bibitem[Malkov and Yashunin(2018)]{malkov2018efficient}
Yu~A Malkov and Dmitry~A Yashunin.
\newblock Efficient and robust approximate nearest neighbor search using hierarchical navigable small world graphs.
\newblock \emph{IEEE transactions on pattern analysis and machine intelligence}, 42\penalty0 (4):\penalty0 824--836, 2018.

\bibitem[Shi et~al.(2020)Shi, Song, Zhou, Li, and Li]{shi2020learning}
Wenxian Shi, Yuxuan Song, Hao Zhou, Bohan Li, and Lei Li.
\newblock Learning from deep model via exploring local targets.
\newblock 2020.

\bibitem[Park et~al.(2021)Park, Cha, Kim, Han, et~al.]{park2021learning}
Dae~Young Park, Moon-Hyun Cha, Daesin Kim, Bohyung Han, et~al.
\newblock Learning student-friendly teacher networks for knowledge distillation.
\newblock \emph{Advances in Neural Information Processing Systems}, 34:\penalty0 13292--13303, 2021.

\bibitem[Turc et~al.(2019)Turc, Chang, Lee, and Toutanova]{turc2019well}
Iulia Turc, Ming-Wei Chang, Kenton Lee, and Kristina Toutanova.
\newblock Well-read students learn better: On the importance of pre-training compact models.
\newblock \emph{arXiv preprint arXiv:1908.08962}, 2019.

\bibitem[Touvron et~al.(2023{\natexlab{b}})Touvron, Martin, Stone, Albert, Almahairi, Babaei, Bashlykov, Batra, Bhargava, Bhosale, et~al.]{touvron2023llama2}
Hugo Touvron, Louis Martin, Kevin Stone, Peter Albert, Amjad Almahairi, Yasmine Babaei, Nikolay Bashlykov, Soumya Batra, Prajjwal Bhargava, Shruti Bhosale, et~al.
\newblock Llama 2: Open foundation and fine-tuned chat models.
\newblock \emph{arXiv preprint arXiv:2307.09288}, 2023{\natexlab{b}}.

\bibitem[Huang et~al.(2023{\natexlab{a}})Huang, Yu, Ma, Zhong, Feng, Wang, Chen, Peng, Feng, Qin, et~al.]{huang2023survey}
Lei Huang, Weijiang Yu, Weitao Ma, Weihong Zhong, Zhangyin Feng, Haotian Wang, Qianglong Chen, Weihua Peng, Xiaocheng Feng, Bing Qin, et~al.
\newblock A survey on hallucination in large language models: Principles, taxonomy, challenges, and open questions.
\newblock \emph{arXiv preprint arXiv:2311.05232}, 2023{\natexlab{a}}.

\bibitem[Ji et~al.(2023)Ji, Lee, Frieske, Yu, Su, Xu, Ishii, Bang, Madotto, and Fung]{ji2023survey}
Ziwei Ji, Nayeon Lee, Rita Frieske, Tiezheng Yu, Dan Su, Yan Xu, Etsuko Ishii, Ye~Jin Bang, Andrea Madotto, and Pascale Fung.
\newblock Survey of hallucination in natural language generation.
\newblock \emph{ACM Computing Surveys}, 55\penalty0 (12):\penalty0 1--38, 2023.

\bibitem[Zhang et~al.(2023{\natexlab{c}})Zhang, Li, Cui, Cai, Liu, Fu, Huang, Zhao, Zhang, Chen, et~al.]{zhang2023siren}
Yue Zhang, Yafu Li, Leyang Cui, Deng Cai, Lemao Liu, Tingchen Fu, Xinting Huang, Enbo Zhao, Yu~Zhang, Yulong Chen, et~al.
\newblock Siren's song in the ai ocean: a survey on hallucination in large language models.
\newblock \emph{arXiv preprint arXiv:2309.01219}, 2023{\natexlab{c}}.

\bibitem[Shi et~al.(2024)Shi, Yang, Cai, Zhang, Wang, Yang, and Lam]{shi2024thorough}
Chufan Shi, Haoran Yang, Deng Cai, Zhisong Zhang, Yifan Wang, Yujiu Yang, and Wai Lam.
\newblock A thorough examination of decoding methods in the era of llms.
\newblock \emph{arXiv preprint arXiv:2402.06925}, 2024.

\bibitem[Welleck et~al.(2024)Welleck, Bertsch, Finlayson, Schoelkopf, Xie, Neubig, Kulikov, and Harchaoui]{welleck2024decoding}
Sean Welleck, Amanda Bertsch, Matthew Finlayson, Hailey Schoelkopf, Alex Xie, Graham Neubig, Ilia Kulikov, and Zaid Harchaoui.
\newblock From decoding to meta-generation: Inference-time algorithms for large language models.
\newblock \emph{arXiv preprint arXiv:2406.16838}, 2024.

\bibitem[Lei et~al.(2023)Lei, Li, Hu, Wang, Yun, Ching, Kamal, et~al.]{lei2023chain}
Deren Lei, Yaxi Li, Mengya Hu, Mingyu Wang, Vincent Yun, Emily Ching, Eslam Kamal, et~al.
\newblock Chain of natural language inference for reducing large language model ungrounded hallucinations.
\newblock \emph{arXiv preprint arXiv:2310.03951}, 2023.

\bibitem[Tian et~al.(2024)Tian, Mitchell, Yao, Manning, and Finn]{tian2024finetuning}
Katherine Tian, Eric Mitchell, Huaxiu Yao, Christopher~D Manning, and Chelsea Finn.
\newblock Fine-tuning language models for factuality.
\newblock In \emph{The Twelfth International Conference on Learning Representations}, 2024.
\newblock URL \url{https://openreview.net/forum?id=WPZ2yPag4K}.

\bibitem[Du et~al.(2024)Du, Li, Torralba, Tenenbaum, and Mordatch]{du2024improving}
Yilun Du, Shuang Li, Antonio Torralba, Joshua~B. Tenenbaum, and Igor Mordatch.
\newblock Improving factuality and reasoning in language models through multiagent debate, 2024.
\newblock URL \url{https://openreview.net/forum?id=QAwaaLJNCk}.

\bibitem[Kadavath et~al.(2022)Kadavath, Conerly, Askell, Henighan, Drain, Perez, Schiefer, Hatfield-Dodds, DasSarma, Tran-Johnson, et~al.]{kadavath2022language}
Saurav Kadavath, Tom Conerly, Amanda Askell, Tom Henighan, Dawn Drain, Ethan Perez, Nicholas Schiefer, Zac Hatfield-Dodds, Nova DasSarma, Eli Tran-Johnson, et~al.
\newblock Language models (mostly) know what they know.
\newblock \emph{arXiv preprint arXiv:2207.05221}, 2022.

\bibitem[Li et~al.(2023{\natexlab{a}})Li, Patel, Vi\'{e}gas, Pfister, and Wattenberg]{NEURIPS2023_ITI}
Kenneth Li, Oam Patel, Fernanda Vi\'{e}gas, Hanspeter Pfister, and Martin Wattenberg.
\newblock Inference-time intervention: Eliciting truthful answers from a language model.
\newblock In A.~Oh, T.~Naumann, A.~Globerson, K.~Saenko, M.~Hardt, and S.~Levine, editors, \emph{Advances in Neural Information Processing Systems}, volume~36, pages 41451--41530. Curran Associates, Inc., 2023{\natexlab{a}}.
\newblock URL \url{https://proceedings.neurips.cc/paper_files/paper/2023/file/81b8390039b7302c909cb769f8b6cd93-Paper-Conference.pdf}.

\bibitem[Saunders et~al.(2022)Saunders, Yeh, Wu, Bills, Ouyang, Ward, and Leike]{saunders2022self}
William Saunders, Catherine Yeh, Jeff Wu, Steven Bills, Long Ouyang, Jonathan Ward, and Jan Leike.
\newblock Self-critiquing models for assisting human evaluators.
\newblock \emph{arXiv preprint arXiv:2206.05802}, 2022.

\bibitem[Wang et~al.(2020{\natexlab{b}})Wang, Liu, and Song]{wang2020language}
Chenguang Wang, Xiao Liu, and Dawn Song.
\newblock Language models are open knowledge graphs.
\newblock \emph{arXiv preprint arXiv:2010.11967}, 2020{\natexlab{b}}.

\bibitem[Chuang et~al.(2024)Chuang, Xie, Luo, Kim, Glass, and He]{chuang2024dola}
Yung-Sung Chuang, Yujia Xie, Hongyin Luo, Yoon Kim, James~R. Glass, and Pengcheng He.
\newblock Dola: Decoding by contrasting layers improves factuality in large language models.
\newblock In \emph{The Twelfth International Conference on Learning Representations}, 2024.
\newblock URL \url{https://openreview.net/forum?id=Th6NyL07na}.

\bibitem[Li et~al.(2022)Li, Holtzman, Fried, Liang, Eisner, Hashimoto, Zettlemoyer, and Lewis]{li2022contrastive}
Xiang~Lisa Li, Ari Holtzman, Daniel Fried, Percy Liang, Jason Eisner, Tatsunori Hashimoto, Luke Zettlemoyer, and Mike Lewis.
\newblock Contrastive decoding: Open-ended text generation as optimization.
\newblock \emph{arXiv preprint arXiv:2210.15097}, 2022.

\bibitem[Zhang et~al.(2023{\natexlab{d}})Zhang, Cui, Bi, and Shi]{zhang2023alleviating}
Yue Zhang, Leyang Cui, Wei Bi, and Shuming Shi.
\newblock Alleviating hallucinations of large language models through induced hallucinations.
\newblock \emph{arXiv preprint arXiv:2312.15710}, 2023{\natexlab{d}}.

\bibitem[Mesnard et~al.(2024)Mesnard, Hardin, Dadashi, Bhupatiraju, Pathak, Sifre, Riviere, Kale, Love, Tafti, Hussenot, Chowdhery, Roberts, Barua, Botev, Castro-Ros, Slone, H'eliou, Tacchetti, Bulanova, Paterson, Tsai, Shahriari, Lan, Choquette-Choo, Crepy, Cer, Ippolito, Reid, Buchatskaya, Ni, Noland, Yan, Tucker, Muraru, Rozhdestvenskiy, Michalewski, Tenney, Grishchenko, Austin, Keeling, Labanowski, Lespiau, Stanway, Brennan, Chen, Ferret, Chiu, Mao-Jones, Lee, Yu, Millican, Sjoesund, Lee, Dixon, Reid, Mikula, Wirth, Sharman, Chinaev, Thain, Bachem, Chang, Wahltinez, Bailey, Michel, Yotov, Sessa, Chaabouni, Comanescu, Jana, Anil, McIlroy, Liu, Mullins, Smith, Borgeaud, Girgin, Douglas, Pandya, Shakeri, De, Klimenko, Hennigan, Feinberg, Stokowiec, hui Chen, Ahmed, Gong, Warkentin, Peran, Giang, Farabet, Vinyals, Dean, Kavukcuoglu, Hassabis, Ghahramani, Eck, Barral, Pereira, Collins, Joulin, Fiedel, Senter, Andreev, and Kenealy]{Mesnard2024GemmaOM}
Gemma Team~Thomas Mesnard, Cassidy Hardin, Robert Dadashi, Surya Bhupatiraju, Shreya Pathak, L.~Sifre, Morgane Riviere, Mihir Kale, J~Christopher Love, Pouya~Dehghani Tafti, L'eonard Hussenot, Aakanksha Chowdhery, Adam Roberts, Aditya Barua, Alex Botev, Alex Castro-Ros, Ambrose Slone, Am'elie H'eliou, Andrea Tacchetti, Anna Bulanova, Antonia Paterson, Beth Tsai, Bobak Shahriari, Charline~Le Lan, Christopher~A. Choquette-Choo, Cl'ement Crepy, Daniel Cer, Daphne Ippolito, David Reid, Elena Buchatskaya, Eric Ni, Eric Noland, Geng Yan, George Tucker, George-Christian Muraru, Grigory Rozhdestvenskiy, Henryk Michalewski, Ian Tenney, Ivan Grishchenko, Jacob Austin, James Keeling, Jane Labanowski, Jean-Baptiste Lespiau, Jeff Stanway, Jenny Brennan, Jeremy Chen, Johan Ferret, Justin Chiu, Justin Mao-Jones, Katherine Lee, Kathy Yu, Katie Millican, Lars~Lowe Sjoesund, Lisa Lee, Lucas Dixon, Machel Reid, Maciej Mikula, Mateo Wirth, Michael Sharman, Nikolai Chinaev, Nithum Thain, Olivier Bachem, Oscar Chang, Oscar
  Wahltinez, Paige Bailey, Paul Michel, Petko Yotov, Pier~Giuseppe Sessa, Rahma Chaabouni, Ramona Comanescu, Reena Jana, Rohan Anil, Ross McIlroy, Ruibo Liu, Ryan Mullins, Samuel~L Smith, Sebastian Borgeaud, Sertan Girgin, Sholto Douglas, Shree Pandya, Siamak Shakeri, Soham De, Ted Klimenko, Tom Hennigan, Vladimir Feinberg, Wojciech Stokowiec, Yu~hui Chen, Zafarali Ahmed, Zhitao Gong, Tris~Brian Warkentin, Ludovic Peran, Minh Giang, Cl'ement Farabet, Oriol Vinyals, Jeffrey Dean, Koray Kavukcuoglu, Demis Hassabis, Zoubin Ghahramani, Douglas Eck, Joelle Barral, Fernando Pereira, Eli Collins, Armand Joulin, Noah Fiedel, Evan Senter, Alek Andreev, and Kathleen Kenealy.
\newblock Gemma: Open models based on gemini research and technology.
\newblock \emph{ArXiv}, abs/2403.08295, 2024.
\newblock URL \url{https://api.semanticscholar.org/CorpusID:268379206}.

\bibitem[Li et~al.(2023{\natexlab{b}})Li, Bubeck, Eldan, Del~Giorno, Gunasekar, and Lee]{li2023textbooks}
Yuanzhi Li, S{\'e}bastien Bubeck, Ronen Eldan, Allie Del~Giorno, Suriya Gunasekar, and Yin~Tat Lee.
\newblock Textbooks are all you need ii: phi-1.5 technical report.
\newblock \emph{arXiv preprint arXiv:2309.05463}, 2023{\natexlab{b}}.

\bibitem[Wang et~al.(2024{\natexlab{b}})Wang, Vahidian, Ye, Gu, Zhang, and Chen]{wang2024coreinfer}
Qinsi Wang, Saeed Vahidian, Hancheng Ye, Jianyang Gu, Jianyi Zhang, and Yiran Chen.
\newblock Coreinfer: Accelerating large language model inference with semantics-inspired adaptive sparse activation.
\newblock \emph{arXiv preprint arXiv:2410.18311}, 2024{\natexlab{b}}.

\bibitem[Chen et~al.(2024{\natexlab{a}})Chen, Lin, Han, and Sun]{chen2024benchmarking}
Jiawei Chen, Hongyu Lin, Xianpei Han, and Le~Sun.
\newblock Benchmarking large language models in retrieval-augmented generation.
\newblock In \emph{Proceedings of the AAAI Conference on Artificial Intelligence}, volume~38, pages 17754--17762, 2024{\natexlab{a}}.

\bibitem[Cheng et~al.(2024)Cheng, Luo, Chen, Liu, Zhao, and Yan]{cheng2024lift}
Xin Cheng, Di~Luo, Xiuying Chen, Lemao Liu, Dongyan Zhao, and Rui Yan.
\newblock Lift yourself up: Retrieval-augmented text generation with self-memory.
\newblock \emph{Advances in Neural Information Processing Systems}, 36, 2024.

\bibitem[Ding et~al.(2024)Ding, Fan, Ning, Wang, Li, Yin, Chua, and Li]{ding2024survey}
Yujuan Ding, Wenqi Fan, Liangbo Ning, Shijie Wang, Hengyun Li, Dawei Yin, Tat-Seng Chua, and Qing Li.
\newblock A survey on rag meets llms: Towards retrieval-augmented large language models.
\newblock \emph{arXiv preprint arXiv:2405.06211}, 2024.

\bibitem[Gao et~al.(2023)Gao, Dai, Pasupat, Chen, Chaganty, Fan, Zhao, Lao, Lee, Juan, and Guu]{gao2023rarr}
Luyu Gao, Zhuyun Dai, Panupong Pasupat, Anthony Chen, Arun~Tejasvi Chaganty, Yicheng Fan, Vincent Zhao, Ni~Lao, Hongrae Lee, Da-Cheng Juan, and Kelvin Guu.
\newblock {RARR}: Researching and revising what language models say, using language models.
\newblock In Anna Rogers, Jordan Boyd-Graber, and Naoaki Okazaki, editors, \emph{Proceedings of the 61st Annual Meeting of the Association for Computational Linguistics (Volume 1: Long Papers)}, pages 16477--16508, Toronto, Canada, July 2023. Association for Computational Linguistics.
\newblock \doi{10.18653/v1/2023.acl-long.910}.
\newblock URL \url{https://aclanthology.org/2023.acl-long.910}.

\bibitem[Ovadia et~al.(2023)Ovadia, Brief, Mishaeli, and Elisha]{ovadia2023fine}
Oded Ovadia, Menachem Brief, Moshik Mishaeli, and Oren Elisha.
\newblock Fine-tuning or retrieval? comparing knowledge injection in llms.
\newblock \emph{arXiv preprint arXiv:2312.05934}, 2023.

\bibitem[Ouyang et~al.(2022)Ouyang, Wu, Jiang, Almeida, Wainwright, Mishkin, Zhang, Agarwal, Slama, Ray, et~al.]{ouyang2022training}
Long Ouyang, Jeffrey Wu, Xu~Jiang, Diogo Almeida, Carroll Wainwright, Pamela Mishkin, Chong Zhang, Sandhini Agarwal, Katarina Slama, Alex Ray, et~al.
\newblock Training language models to follow instructions with human feedback.
\newblock \emph{Advances in Neural Information Processing Systems}, 35:\penalty0 27730--27744, 2022.

\bibitem[Rafailov et~al.(2024)Rafailov, Sharma, Mitchell, Manning, Ermon, and Finn]{rafailov2024direct}
Rafael Rafailov, Archit Sharma, Eric Mitchell, Christopher~D Manning, Stefano Ermon, and Chelsea Finn.
\newblock Direct preference optimization: Your language model is secretly a reward model.
\newblock \emph{Advances in Neural Information Processing Systems}, 36, 2024.

\bibitem[Yuan et~al.(2024)Yuan, Pang, Cho, Sukhbaatar, Xu, and Weston]{yuan2024self}
Weizhe Yuan, Richard~Yuanzhe Pang, Kyunghyun Cho, Sainbayar Sukhbaatar, Jing Xu, and Jason Weston.
\newblock Self-rewarding language models.
\newblock \emph{arXiv preprint arXiv:2401.10020}, 2024.

\bibitem[Vijayakumar et~al.(2018)Vijayakumar, Cogswell, Selvaraju, Sun, Lee, Crandall, and Batra]{Vijayakumar_Cogswell_Selvaraju_Sun_Lee_Crandall_Batra_2018}
Ashwin Vijayakumar, Michael Cogswell, Ramprasaath Selvaraju, Qing Sun, Stefan Lee, David Crandall, and Dhruv Batra.
\newblock Diverse beam search for improved description of complex scenes.
\newblock \emph{Proceedings of the AAAI Conference on Artificial Intelligence}, 32\penalty0 (1), Apr. 2018.
\newblock \doi{10.1609/aaai.v32i1.12340}.
\newblock URL \url{https://ojs.aaai.org/index.php/AAAI/article/view/12340}.

\bibitem[Holtzman et~al.(2019)Holtzman, Buys, Du, Forbes, and Choi]{holtzman2019curious}
Ari Holtzman, Jan Buys, Li~Du, Maxwell Forbes, and Yejin Choi.
\newblock The curious case of neural text degeneration.
\newblock \emph{arXiv preprint arXiv:1904.09751}, 2019.

\bibitem[Yang et~al.(2023{\natexlab{a}})Yang, Cai, Li, Bi, Lam, and Shi]{yang2023frustratingly}
Haoran Yang, Deng Cai, Huayang Li, Wei Bi, Wai Lam, and Shuming Shi.
\newblock A frustratingly simple decoding method for neural text generation.
\newblock \emph{arXiv preprint arXiv:2305.12675}, 2023{\natexlab{a}}.

\bibitem[M{\"u}ller et~al.(2019)M{\"u}ller, Kornblith, and Hinton]{muller2019does}
Rafael M{\"u}ller, Simon Kornblith, and Geoffrey~E Hinton.
\newblock When does label smoothing help?
\newblock \emph{Advances in neural information processing systems}, 32, 2019.

\bibitem[Thiel(2008)]{ClassificationSoftRobust}
Christian Thiel.
\newblock Classification on soft labels is robust against label noise.
\newblock In Ignac Lovrek, Robert~J. Howlett, and Lakhmi~C. Jain, editors, \emph{Knowledge-Based Intelligent Information and Engineering Systems}, pages 65--73, Berlin, Heidelberg, 2008. Springer Berlin Heidelberg.
\newblock ISBN 978-3-540-85563-7.

\bibitem[Zhang et~al.(2021)Zhang, Jiang, Hou, Wei, Han, Li, and Cheng]{zhang2021delving}
Chang-Bin Zhang, Peng-Tao Jiang, Qibin Hou, Yunchao Wei, Qi~Han, Zhen Li, and Ming-Ming Cheng.
\newblock Delving deep into label smoothing.
\newblock \emph{IEEE Transactions on Image Processing}, 30:\penalty0 5984--5996, 2021.

\bibitem[Lin et~al.(2022{\natexlab{b}})Lin, Hilton, and Evans]{lin-etal-2022-truthfulqa}
Stephanie Lin, Jacob Hilton, and Owain Evans.
\newblock {T}ruthful{QA}: Measuring how models mimic human falsehoods.
\newblock In Smaranda Muresan, Preslav Nakov, and Aline Villavicencio, editors, \emph{Proceedings of the 60th Annual Meeting of the Association for Computational Linguistics (Volume 1: Long Papers)}, pages 3214--3252, Dublin, Ireland, May 2022{\natexlab{b}}. Association for Computational Linguistics.
\newblock \doi{10.18653/v1/2022.acl-long.229}.
\newblock URL \url{https://aclanthology.org/2022.acl-long.229}.

\bibitem[Muhlgay et~al.(2023)Muhlgay, Ram, Magar, Levine, Ratner, Belinkov, Abend, Leyton-Brown, Shashua, and Shoham]{muhlgay2023generating}
Dor Muhlgay, Ori Ram, Inbal Magar, Yoav Levine, Nir Ratner, Yonatan Belinkov, Omri Abend, Kevin Leyton-Brown, Amnon Shashua, and Yoav Shoham.
\newblock Generating benchmarks for factuality evaluation of language models.
\newblock \emph{arXiv preprint arXiv:2307.06908}, 2023.

\bibitem[Wei et~al.(2022)Wei, Wang, Schuurmans, Bosma, brian ichter, Xia, Chi, Le, and Zhou]{wei2022chain}
Jason Wei, Xuezhi Wang, Dale Schuurmans, Maarten Bosma, brian ichter, Fei Xia, Ed~H. Chi, Quoc~V Le, and Denny Zhou.
\newblock Chain of thought prompting elicits reasoning in large language models.
\newblock In Alice~H. Oh, Alekh Agarwal, Danielle Belgrave, and Kyunghyun Cho, editors, \emph{Advances in Neural Information Processing Systems}, 2022.
\newblock URL \url{https://openreview.net/forum?id=_VjQlMeSB_J}.

\bibitem[Geva et~al.(2021)Geva, Khashabi, Segal, Khot, Roth, and Berant]{geva2021did}
Mor Geva, Daniel Khashabi, Elad Segal, Tushar Khot, Dan Roth, and Jonathan Berant.
\newblock Did aristotle use a laptop? a question answering benchmark with implicit reasoning strategies.
\newblock \emph{Transactions of the Association for Computational Linguistics}, 9:\penalty0 346--361, 2021.

\bibitem[Cobbe et~al.(2021)Cobbe, Kosaraju, Bavarian, Chen, Jun, Kaiser, Plappert, Tworek, Hilton, Nakano, et~al.]{cobbe2021training}
Karl Cobbe, Vineet Kosaraju, Mohammad Bavarian, Mark Chen, Heewoo Jun, Lukasz Kaiser, Matthias Plappert, Jerry Tworek, Jacob Hilton, Reiichiro Nakano, et~al.
\newblock Training verifiers to solve math word problems.
\newblock \emph{arXiv preprint arXiv:2110.14168}, 2021.

\bibitem[Jiang et~al.(2024)Jiang, Sablayrolles, Roux, Mensch, Savary, Bamford, Chaplot, de~las Casas, Hanna, Bressand, Lengyel, Bour, Lample, Lavaud, Saulnier, Lachaux, Stock, Subramanian, Yang, Antoniak, Scao, Gervet, Lavril, Wang, Lacroix, and Sayed]{jiang2024mixtralexperts}
Albert~Q. Jiang, Alexandre Sablayrolles, Antoine Roux, Arthur Mensch, Blanche Savary, Chris Bamford, Devendra~Singh Chaplot, Diego de~las Casas, Emma~Bou Hanna, Florian Bressand, Gianna Lengyel, Guillaume Bour, Guillaume Lample, Lélio~Renard Lavaud, Lucile Saulnier, Marie-Anne Lachaux, Pierre Stock, Sandeep Subramanian, Sophia Yang, Szymon Antoniak, Teven~Le Scao, Théophile Gervet, Thibaut Lavril, Thomas Wang, Timothée Lacroix, and William~El Sayed.
\newblock Mixtral of experts, 2024.
\newblock URL \url{https://arxiv.org/abs/2401.04088}.

\bibitem[Chen et~al.(2024{\natexlab{b}})Chen, Xiong, Liu, Wu, Xiao, Gao, and He]{chen2024context}
Shiqi Chen, Miao Xiong, Junteng Liu, Zhengxuan Wu, Teng Xiao, Siyang Gao, and Junxian He.
\newblock In-context sharpness as alerts: An inner representation perspective for hallucination mitigation.
\newblock \emph{arXiv preprint arXiv:2403.01548}, 2024{\natexlab{b}}.

\bibitem[Xu et~al.(2022)Xu, Liu, Yan, Cai, Li, and Li]{xu2022learning}
Jin Xu, Xiaojiang Liu, Jianhao Yan, Deng Cai, Huayang Li, and Jian Li.
\newblock Learning to break the loop: Analyzing and mitigating repetitions for neural text generation.
\newblock \emph{Advances in Neural Information Processing Systems}, 35:\penalty0 3082--3095, 2022.

\bibitem[Kuo et~al.(2025{\natexlab{a}})Kuo, Zhang, Zhang, Tang, DiValentin, Ding, Sun, Chen, Hass, Chen, Chen, and Li]{kuo2025proactive}
Martin Kuo, Jingyang Zhang, Jianyi Zhang, Minxue Tang, Louis DiValentin, Aolin Ding, Jingwei Sun, William Chen, Amin Hass, Tianlong Chen, Yiran Chen, and Hai Li.
\newblock Proactive privacy amnesia for large language models: Safeguarding {PII} with negligible impact on model utility.
\newblock In \emph{The Thirteenth International Conference on Learning Representations}, 2025{\natexlab{a}}.
\newblock URL \url{https://openreview.net/forum?id=io8uRPYktn}.

\bibitem[Zhang et~al.(2024{\natexlab{f}})Zhang, Sun, Yeats, Ouyang, Kuo, Zhang, Yang, and Li]{zhang2024min}
Jingyang Zhang, Jingwei Sun, Eric Yeats, Yang Ouyang, Martin Kuo, Jianyi Zhang, Hao~Frank Yang, and Hai Li.
\newblock Min-k\%++: Improved baseline for detecting pre-training data from large language models.
\newblock \emph{arXiv preprint arXiv:2404.02936}, 2024{\natexlab{f}}.

\bibitem[Kuo et~al.(2025{\natexlab{b}})Kuo, Zhang, Ding, Wang, DiValentin, Bao, Wei, Li, and Chen]{kuo2025hcothijackingchainofthoughtsafety}
Martin Kuo, Jianyi Zhang, Aolin Ding, Qinsi Wang, Louis DiValentin, Yujia Bao, Wei Wei, Hai Li, and Yiran Chen.
\newblock H-cot: Hijacking the chain-of-thought safety reasoning mechanism to jailbreak large reasoning models, including openai o1/o3, deepseek-r1, and gemini 2.0 flash thinking, 2025{\natexlab{b}}.
\newblock URL \url{https://arxiv.org/abs/2502.12893}.

\bibitem[Lin et~al.(2024)Lin, Fu, Zhang, Liu, Zhang, Sun, Li, Chen, et~al.]{lin2024speechprune}
Yueqian Lin, Yuzhe Fu, Jingyang Zhang, Yudong Liu, Jianyi Zhang, Jingwei Sun, Hai Li, Yiran Chen, et~al.
\newblock Speechprune: Context-aware token pruning for speech information retrieval.
\newblock \emph{arXiv preprint arXiv:2412.12009}, 2024.

\bibitem[Zhang et~al.(2022{\natexlab{b}})Zhang, Chen, and Chen]{zhang2022join}
Jianyi Zhang, Yiran Chen, and Jianshu Chen.
\newblock Join-chain network: A logical reasoning view of the multi-head attention in transformer.
\newblock In \emph{2022 IEEE International Conference on Data Mining Workshops (ICDMW)}, pages 1--11. IEEE, 2022{\natexlab{b}}.

\bibitem[Kuo et~al.(2023)Kuo, Zhang, and Chen]{kuo2023dacbertleveragingdependencyagreement}
Martin Kuo, Jianyi Zhang, and Yiran Chen.
\newblock Dacbert: Leveraging dependency agreement for cost-efficient bert pretraining, 2023.
\newblock URL \url{https://arxiv.org/abs/2311.04799}.

\bibitem[Yang et~al.(2022)Yang, Zhang, Song, Hong, Xu, Zhao, Shao, Zhang, Cui, and Yang]{yang2022diffusion}
Ling Yang, Zhilong Zhang, Yang Song, Shenda Hong, Runsheng Xu, Yue Zhao, Yingxia Shao, Wentao Zhang, Bin Cui, and Ming-Hsuan Yang.
\newblock Diffusion models: A comprehensive survey of methods and applications.
\newblock \emph{arXiv preprint arXiv:2209.00796}, 2022.

\bibitem[Zhou et~al.(2023)Zhou, Liu, Zhu, Yang, Chen, and Xu]{zhou2023shifted}
Yufan Zhou, Bingchen Liu, Yizhe Zhu, Xiao Yang, Changyou Chen, and Jinhui Xu.
\newblock Shifted diffusion for text-to-image generation.
\newblock In \emph{Proceedings of the IEEE/CVF Conference on Computer Vision and Pattern Recognition}, pages 10157--10166, 2023.

\bibitem[Ruiz et~al.(2023)Ruiz, Li, Jampani, Pritch, Rubinstein, and Aberman]{ruiz2023dreambooth}
Nataniel Ruiz, Yuanzhen Li, Varun Jampani, Yael Pritch, Michael Rubinstein, and Kfir Aberman.
\newblock Dreambooth: Fine tuning text-to-image diffusion models for subject-driven generation.
\newblock In \emph{Proceedings of the IEEE/CVF Conference on Computer Vision and Pattern Recognition}, pages 22500--22510, 2023.

\bibitem[Lugmayr et~al.(2022)Lugmayr, Danelljan, Romero, Yu, Timofte, and Van~Gool]{lugmayr2022repaint}
Andreas Lugmayr, Martin Danelljan, Andres Romero, Fisher Yu, Radu Timofte, and Luc Van~Gool.
\newblock Repaint: Inpainting using denoising diffusion probabilistic models.
\newblock In \emph{Proceedings of the IEEE/CVF Conference on Computer Vision and Pattern Recognition}, pages 11461--11471, 2022.

\bibitem[Esser et~al.(2021)Esser, Rombach, Blattmann, and Ommer]{esser2021imagebart}
Patrick Esser, Robin Rombach, Andreas Blattmann, and Bjorn Ommer.
\newblock Imagebart: Bidirectional context with multinomial diffusion for autoregressive image synthesis.
\newblock \emph{Advances in neural information processing systems}, 34:\penalty0 3518--3532, 2021.

\bibitem[Jing et~al.(2022)Jing, Corso, Berlinghieri, and Jaakkola]{jing2022subspace}
Bowen Jing, Gabriele Corso, Renato Berlinghieri, and Tommi Jaakkola.
\newblock Subspace diffusion generative models.
\newblock In \emph{European Conference on Computer Vision}, pages 274--289. Springer, 2022.

\bibitem[Xie et~al.(2023)Xie, Yuan, Dong, and Li]{xie2023diffusion}
Yutong Xie, Minne Yuan, Bin Dong, and Quanzheng Li.
\newblock Diffusion model for generative image denoising.
\newblock \emph{arXiv preprint arXiv:2302.02398}, 2023.

\bibitem[Yang et~al.(2023{\natexlab{b}})Yang, Liang, and Su]{Yang2023RealWorldDV}
Chenghan Yang, Lijing Liang, and Zhixun Su.
\newblock Real-world denoising via diffusion model.
\newblock \emph{ArXiv}, abs/2305.04457, 2023{\natexlab{b}}.
\newblock URL \url{https://api.semanticscholar.org/CorpusID:258557313}.

\bibitem[Han et~al.(2022)Han, Zheng, and Zhou]{han2022card}
Xizewen Han, Huangjie Zheng, and Mingyuan Zhou.
\newblock Card: Classification and regression diffusion models.
\newblock \emph{Advances in Neural Information Processing Systems}, 35:\penalty0 18100--18115, 2022.

\bibitem[Blattmann et~al.(2023)Blattmann, Rombach, Ling, Dockhorn, Kim, Fidler, and Kreis]{blattmann2023videoldm}
Andreas Blattmann, Robin Rombach, Huan Ling, Tim Dockhorn, Seung~Wook Kim, Sanja Fidler, and Karsten Kreis.
\newblock Align your latents: High-resolution video synthesis with latent diffusion models.
\newblock In \emph{IEEE Conference on Computer Vision and Pattern Recognition ({CVPR})}, 2023.

\bibitem[Wang et~al.(2023)Wang, Zhao, and Xing]{wang2023stylediffusion}
Zhizhong Wang, Lei Zhao, and Wei Xing.
\newblock Stylediffusion: Controllable disentangled style transfer via diffusion models.
\newblock \emph{arXiv preprint arXiv:2308.07863}, 2023.

\bibitem[Zhang et~al.(2022{\natexlab{c}})Zhang, Huang, Tang, Huang, Ma, Dong, and Xu]{zhang2022inversion}
Yuxin Zhang, Nisha Huang, Fan Tang, Haibin Huang, Chongyang Ma, Weiming Dong, and Changsheng Xu.
\newblock Inversion-based creativity transfer with diffusion models.
\newblock \emph{arXiv preprint arXiv:2211.13203}, 2022{\natexlab{c}}.

\bibitem[Balaji et~al.(2022)Balaji, Nah, Huang, Vahdat, Song, Kreis, Aittala, Aila, Laine, Catanzaro, et~al.]{balaji2022ediffi}
Yogesh Balaji, Seungjun Nah, Xun Huang, Arash Vahdat, Jiaming Song, Karsten Kreis, Miika Aittala, Timo Aila, Samuli Laine, Bryan Catanzaro, et~al.
\newblock ediffi: Text-to-image diffusion models with an ensemble of expert denoisers.
\newblock \emph{arXiv preprint arXiv:2211.01324}, 2022.

\bibitem[Shonenkov et~al.(2023)Shonenkov, Konstantinov, Bakshandaeva, Schuhmann, Ivanova, and Klokova]{DeepFloyd}
Alex Shonenkov, Misha Konstantinov, Daria Bakshandaeva, Christoph Schuhmann, Ksenia Ivanova, and Nadiia Klokova.
\newblock Deepfloyd.
\newblock \url{https://github.com/deep-floyd/if}, 2023.

\bibitem[Raffel et~al.(2020)Raffel, Shazeer, Roberts, Lee, Narang, Matena, Zhou, Li, Liu, et~al.]{raffel2020exploring}
Colin Raffel, Noam Shazeer, Adam Roberts, Katherine Lee, Sharan Narang, Michael Matena, Yanqi Zhou, Wei Li, Peter~J Liu, et~al.
\newblock Exploring the limits of transfer learning with a unified text-to-text transformer.
\newblock \emph{J. Mach. Learn. Res.}, 21\penalty0 (140):\penalty0 1--67, 2020.

\bibitem[Radford et~al.(2021)Radford, Kim, Hallacy, Ramesh, Goh, Agarwal, Sastry, Askell, Mishkin, Clark, et~al.]{radford2021learning}
Alec Radford, Jong~Wook Kim, Chris Hallacy, Aditya Ramesh, Gabriel Goh, Sandhini Agarwal, Girish Sastry, Amanda Askell, Pamela Mishkin, Jack Clark, et~al.
\newblock Learning transferable visual models from natural language supervision.
\newblock In \emph{International conference on machine learning}, pages 8748--8763. PMLR, 2021.

\bibitem[Liu et~al.(2022)Liu, Garrette, Saharia, Chan, Roberts, Narang, Blok, Mical, Norouzi, and Constant]{liu2022character}
Rosanne Liu, Dan Garrette, Chitwan Saharia, William Chan, Adam Roberts, Sharan Narang, Irina Blok, RJ~Mical, Mohammad Norouzi, and Noah Constant.
\newblock Character-aware models improve visual text rendering.
\newblock \emph{arXiv preprint arXiv:2212.10562}, 2022.

\bibitem[Ma et~al.(2023)Ma, Zhao, Chen, Wang, Niu, Lu, and Lin]{ma2023glyphdraw}
Jian Ma, Mingjun Zhao, Chen Chen, Ruichen Wang, Di~Niu, Haonan Lu, and Xiaodong Lin.
\newblock Glyphdraw: Learning to draw chinese characters in image synthesis models coherently, 2023.

\bibitem[Chen et~al.(2023{\natexlab{a}})Chen, Huang, Lv, Cui, Chen, and Wei]{chen2023textdiffuser}
Jingye Chen, Yupan Huang, Tengchao Lv, Lei Cui, Qifeng Chen, and Furu Wei.
\newblock Textdiffuser: Diffusion models as text painters.
\newblock \emph{arXiv preprint arXiv:2305.10855}, 2023{\natexlab{a}}.

\bibitem[OpenAI(2023)]{OpenAI2023GPT4TR}
OpenAI.
\newblock Gpt-4 technical report.
\newblock \emph{ArXiv}, abs/2303.08774, 2023.
\newblock URL \url{https://api.semanticscholar.org/CorpusID:257532815}.

\bibitem[Ramesh et~al.(2021)Ramesh, Pavlov, Goh, Gray, Voss, Radford, Chen, and Sutskever]{ramesh2021zero}
Aditya Ramesh, Mikhail Pavlov, Gabriel Goh, Scott Gray, Chelsea Voss, Alec Radford, Mark Chen, and Ilya Sutskever.
\newblock Zero-shot text-to-image generation.
\newblock In \emph{International Conference on Machine Learning}, pages 8821--8831. PMLR, 2021.

\bibitem[Yu et~al.(2022)Yu, Xu, Koh, Luong, Baid, Wang, Vasudevan, Ku, Yang, Ayan, et~al.]{yu2022scaling}
Jiahui Yu, Yuanzhong Xu, Jing~Yu Koh, Thang Luong, Gunjan Baid, Zirui Wang, Vijay Vasudevan, Alexander Ku, Yinfei Yang, Burcu~Karagol Ayan, et~al.
\newblock Scaling autoregressive models for content-rich text-to-image generation.
\newblock \emph{arXiv preprint arXiv:2206.10789}, 2\penalty0 (3):\penalty0 5, 2022.

\bibitem[Zhang and Agrawala(2023)]{zhang2023adding}
Lvmin Zhang and Maneesh Agrawala.
\newblock Adding conditional control to text-to-image diffusion models.
\newblock \emph{arXiv preprint arXiv:2302.05543}, 2023.

\bibitem[Li et~al.(2023{\natexlab{c}})Li, Liu, Wu, Mu, Yang, Gao, Li, and Lee]{li2023gligen}
Yuheng Li, Haotian Liu, Qingyang Wu, Fangzhou Mu, Jianwei Yang, Jianfeng Gao, Chunyuan Li, and Yong~Jae Lee.
\newblock Gligen: Open-set grounded text-to-image generation.
\newblock In \emph{Proceedings of the IEEE/CVF Conference on Computer Vision and Pattern Recognition}, pages 22511--22521, 2023{\natexlab{c}}.

\bibitem[Yang et~al.(2024)Yang, Gui, Yuan, Liang, Ding, Hu, and Chen]{yang2024glyphcontrol}
Yukang Yang, Dongnan Gui, Yuhui Yuan, Weicong Liang, Haisong Ding, Han Hu, and Kai Chen.
\newblock Glyphcontrol: Glyph conditional control for visual text generation.
\newblock \emph{Advances in Neural Information Processing Systems}, 36, 2024.

\bibitem[Tuo et~al.(2023)Tuo, Xiang, He, Geng, and Xie]{tuo2023anytext}
Yuxiang Tuo, Wangmeng Xiang, Jun-Yan He, Yifeng Geng, and Xuansong Xie.
\newblock Anytext: Multilingual visual text generation and editing.
\newblock \emph{arXiv preprint arXiv:2311.03054}, 2023.

\bibitem[Chen et~al.(2023{\natexlab{b}})Chen, Huang, Lv, Cui, Chen, and Wei]{chen2023textdiffuser-2}
Jingye Chen, Yupan Huang, Tengchao Lv, Lei Cui, Qifeng Chen, and Furu Wei.
\newblock Textdiffuser-2: Unleashing the power of language models for text rendering.
\newblock \emph{arXiv preprint arXiv:2311.16465}, 2023{\natexlab{b}}.

\bibitem[White and Rohrer(1983)]{white1983image}
James~M White and Gene~D Rohrer.
\newblock Image thresholding for optical character recognition and other applications requiring character image extraction.
\newblock \emph{IBM Journal of research and development}, 27\penalty0 (4):\penalty0 400--411, 1983.

\bibitem[Cash and Hatamian(1987)]{cash1987optical}
Glenn~L Cash and Mehdi Hatamian.
\newblock Optical character recognition by the method of moments.
\newblock \emph{Computer vision, graphics, and image processing}, 39\penalty0 (3):\penalty0 291--310, 1987.

\bibitem[Omran and Jarallah(2017)]{Iraqi_car_license}
Safaa~S. Omran and Jumana~A. Jarallah.
\newblock Iraqi car license plate recognition using ocr.
\newblock In \emph{2017 Annual Conference on New Trends in Information \& Communications Technology Applications (NTICT)}, pages 298--303, 2017.
\newblock \doi{10.1109/NTICT.2017.7976127}.

\bibitem[Schreiber et~al.(2014)Schreiber, Poggenhans, and Stiller]{Detecting_symbols6957755}
Markus Schreiber, Fabian Poggenhans, and Christoph Stiller.
\newblock Detecting symbols on road surface for mapping and localization using ocr.
\newblock In \emph{17th International IEEE Conference on Intelligent Transportation Systems (ITSC)}, pages 597--602, 2014.
\newblock \doi{10.1109/ITSC.2014.6957755}.

\bibitem[Wu et~al.(2023)Wu, Chen, Wang, Wang, Gan, Fang, and Xu]{wu2023ocr}
Zizhang Wu, Xinyuan Chen, Jizheng Wang, Xiaoquan Wang, Yuanzhu Gan, Muqing Fang, and Tianhao Xu.
\newblock Ocr-rtps: an ocr-based real-time positioning system for the valet parking.
\newblock \emph{Applied Intelligence}, pages 1--15, 2023.

\bibitem[Huang et~al.(2023{\natexlab{b}})Huang, Dong, Wang, Hao, Singhal, Ma, Lv, Cui, Mohammed, Liu, et~al.]{huang2023language}
Shaohan Huang, Li~Dong, Wenhui Wang, Yaru Hao, Saksham Singhal, Shuming Ma, Tengchao Lv, Lei Cui, Owais~Khan Mohammed, Qiang Liu, et~al.
\newblock Language is not all you need: Aligning perception with language models.
\newblock \emph{arXiv preprint arXiv:2302.14045}, 2023{\natexlab{b}}.

\bibitem[Shen et~al.(2023)Shen, Song, Tan, Li, Lu, and Zhuang]{shen2023hugginggpt}
Yongliang Shen, Kaitao Song, Xu~Tan, Dongsheng Li, Weiming Lu, and Yueting Zhuang.
\newblock Hugginggpt: Solving ai tasks with chatgpt and its friends in huggingface.
\newblock \emph{arXiv preprint arXiv:2303.17580}, 2023.

\bibitem[Saharia et~al.(2022{\natexlab{b}})Saharia, Chan, Saxena, Li, Whang, Denton, Ghasemipour, Gontijo~Lopes, Karagol~Ayan, Salimans, et~al.]{saharia2022imagen}
Chitwan Saharia, William Chan, Saurabh Saxena, Lala Li, Jay Whang, Emily~L Denton, Kamyar Ghasemipour, Raphael Gontijo~Lopes, Burcu Karagol~Ayan, Tim Salimans, et~al.
\newblock Photorealistic text-to-image diffusion models with deep language understanding.
\newblock \emph{Advances in Neural Information Processing Systems}, 35:\penalty0 36479--36494, 2022{\natexlab{b}}.

\bibitem[Brown et~al.(2020)Brown, Mann, Ryder, Subbiah, Kaplan, Dhariwal, Neelakantan, Shyam, Sastry, Askell, et~al.]{brown2020GPT3}
Tom Brown, Benjamin Mann, Nick Ryder, Melanie Subbiah, Jared~D Kaplan, Prafulla Dhariwal, Arvind Neelakantan, Pranav Shyam, Girish Sastry, Amanda Askell, et~al.
\newblock Language models are few-shot learners.
\newblock \emph{Advances in neural information processing systems}, 33:\penalty0 1877--1901, 2020.

\bibitem[von Platen et~al.(2022)von Platen, Patil, Lozhkov, Cuenca, Lambert, Rasul, Davaadorj, and Wolf]{von-platen-etal-2022-diffusers}
Patrick von Platen, Suraj Patil, Anton Lozhkov, Pedro Cuenca, Nathan Lambert, Kashif Rasul, Mishig Davaadorj, and Thomas Wolf.
\newblock Diffusers: State-of-the-art diffusion models.
\newblock \url{https://github.com/huggingface/diffusers}, 2022.

\bibitem[Rombach et~al.(2021)Rombach, Blattmann, Lorenz, Esser, and Ommer]{rombach2021LDM}
Robin Rombach, Andreas Blattmann, Dominik Lorenz, Patrick Esser, and Björn Ommer.
\newblock High-resolution image synthesis with latent diffusion models, 2021.

\bibitem[Loshchilov and Hutter(2017)]{loshchilov2017AdamW}
Ilya Loshchilov and Frank Hutter.
\newblock Decoupled weight decay regularization.
\newblock \emph{arXiv preprint arXiv:1711.05101}, 2017.

\bibitem[Heusel et~al.(2017)Heusel, Ramsauer, Unterthiner, Nessler, and Hochreiter]{heusel2017FID}
Martin Heusel, Hubert Ramsauer, Thomas Unterthiner, Bernhard Nessler, and Sepp Hochreiter.
\newblock Gans trained by a two time-scale update rule converge to a local nash equilibrium.
\newblock \emph{Advances in neural information processing systems}, 30, 2017.

\bibitem[Liao et~al.(2020)Liao, Pang, Huang, Hassner, and Bai]{liao2020maskspotter}
Minghui Liao, Guan Pang, Jing Huang, Tal Hassner, and Xiang Bai.
\newblock Mask textspotter v3: Segmentation proposal network for robust scene text spotting.
\newblock In \emph{Computer Vision--ECCV 2020: 16th European Conference, Glasgow, UK, August 23--28, 2020, Proceedings, Part XI 16}, pages 706--722. Springer, 2020.

\bibitem[Chen et~al.(2017)Chen, Laine, and Player]{seal}
Hao Chen, Kim Laine, and Rachel Player.
\newblock Simple encrypted arithmetic library-seal v2.1.
\newblock In \emph{International Conference on Financial Cryptography and Data Security}, pages 3--18. Springer, 2017.

\bibitem[Halevi and Shoup(2014)]{helib}
Shai Halevi and Victor Shoup.
\newblock Algorithms in {HElib}.
\newblock In \emph{Annual Cryptology Conference}, pages 554--571. Springer, 2014.

\bibitem[McMahan et~al.(2017{\natexlab{b}})McMahan, Moore, Ramage, Hampson, et~al.]{mcmahan2016communication}
H~Brendan McMahan, Eider Moore, Daniel Ramage, Seth Hampson, et~al.
\newblock {Communication-efficient Learning of Deep Networks from Decentralized Data}.
\newblock \emph{Artificial Intelligence and Statistics}, 2017{\natexlab{b}}.

\end{thebibliography}

\biography
Jianyi Zhang completes his Ph.D at CEI Center, Duke University. He previously received his B.S. in Mathematics from Fudan University. His research focuses on probabilistic modeling for generative AI and trustworthy AI, with a particular focus on Large (Vision-)Language Models and Diffusion Models. His research projects include LLM-controlled decoding, LLM factuality, privacy, and security, retrieval-augmented generation for LLMs, text-rich image generation with Diffusion Models, LLM (federated) alignment, and efficient inference for LLMs and Diffusion Models. He conducted research internships at Adobe, Amazon, and Google, enriching his experience across academia and industry. In 2024, he is the Principal Investigator at Duke University $\&$ National Artificial Intelligence Research Resource (NAIRR) on diffusion model acceleration, NAIRR240270. In 2023, he served as the Principal Investigator for a project on LLM for Hardware between Duke University and the Pittsburgh Supercomputing Center.

\end{document}